\documentclass[twoside,11pt]{article}

\usepackage{blindtext}

\usepackage{dmlr2e}

\usepackage{lastpage}
\dmlrheading{26}{2026}{1-\pageref{LastPage}}{3/26; Revised 7/26}{8/26}{21-0000}{Ancarani, Tores et al.} 
\def\openreview{\url{https://openreview.net/forum?id=ai2PAWRIVA}} 

\ShortHeadings{MObyGaze}{Ancarani, Tores, et al.}

\usepackage[utf8]{inputenc} 
\usepackage[T1]{fontenc}    
\usepackage{hyperref}       
\usepackage{url}            
\usepackage{booktabs}       
\usepackage{amsfonts}       
\usepackage{nicefrac}       
\usepackage{microtype}      
\usepackage[x11names]{xcolor}         
\usepackage{bbold}          
\usepackage{amsmath}        
\usepackage{amssymb}        
\usepackage{multirow, graphicx}
\usepackage{pgfplots}
\pgfplotsset{compat=1.18}
\usepackage{pgfplotstable}
\usepgfplotslibrary{groupplots}
\usetikzlibrary{patterns}
\usetikzlibrary{positioning, shapes.geometric, arrows.meta, calc, fit}
\usepackage{multicol}
\usepackage{caption}
\usepackage{subcaption}
\usepackage{comment}
\usepackage{makecell}

\usepackage{listings}
\newcommand{\revised}[1]{\textcolor{black}{{#1}}} 
\newcommand{\darkgreen}[1]{\textcolor{darkgreen}{{#1}}}
\newcommand{\qwen}{\texttt{Qwen2.5-VL}}
\newcommand{\intern}{\texttt{InternVideo2.5}}
\newcommand{\internvl}{\texttt{InternVL3}}
\newcommand{\llama}{\texttt{VideoLLaMA 3}}
\newcommand{\llava}{\texttt{LLaVA-Video}}

\definecolor{darkgreen}{rgb}{0.0, 0.5, 0.0}

\title{MObyGaze: a film dataset of multimodal objectification densely annotated by experts}

\author{\name Elisa Ancarani \email elisa.ancarani@univ-cotedazur.fr \\
       \addr Université Côte d’Azur, CNRS, I3S, France
       \AND
       \name Julie Tores \email julie.tores@univ-cotedazur.fr \\
       \addr Université Côte d’Azur, CNRS, Inria, I3S, France
       \AND
       Lucile Sassatelli \email lucile.sassatelli@univ-cotedazur.fr \\
       \addr Université Côte d’Azur, CNRS, I3S, France
       \AND
       Hui-Yin Wu \email hui-yin.wu@inria.fr \\
       \addr Université Côte d’Azur, Inria, France
       \AND
       Rémy Sun \email remy.sun@inria.fr \\
       \addr Université Côte d’Azur, Inria, CNRS, I3S, France
       \AND
       Paul Mouret \email paul.mouret@univ-cotedazur.fr \\
       \addr Université Côte d’Azur, CNRS, I3S, France
       \AND
       Clément Bergman \email clement.bergman@inria.fr \\
       \addr Université Côte d’Azur, Inria, France
       \AND
       Léa Andolfi \email lea.andolfi@sorbonne-universite.fr \\
       \addr Sorbonne Université, GRIPIC, France
       \AND
       Victor Ecrement \email victor.ecrement@sorbonne-universite.fr \\
       \addr Sorbonne Université, GRIPIC, France
       \AND
       Thierry Devars \email thierry.devars@sorbonne-universite.fr \\
       \addr Sorbonne Université, GRIPIC, France
       \AND
       Magali Guaresi \email magali.guaresi@univ-cotedazur.fr \\
       \addr Université Côte d’Azur, CNRS, BCL, France
       \AND
       Virginie Julliard \email virginie.julliard@sorbonne-universite.fr \\
       \addr Sorbonne Université, GRIPIC, France
       \AND
       Sarah Lécossais \email sarah.lecossais@univ-paris13.fr \\
       \addr Université Sorbonne Paris Nord, LabSIC, France
       \AND
       Frédéric Precioso \email frederic.precioso@univ-cotedazur.fr \\
       \addr Université Côte d’Azur, CNRS, Inria, I3S, France
       }

\editor{Hugo Jair Escalante}

\begin{document}

\maketitle

\begin{abstract}
Characterizing and quantifying gender representation disparities in audiovisual storytelling contents is necessary to grasp how stereotypes may perpetuate on screen.
In this article, we consider the high-level construct of objectification and introduce a new AI task to the ML community: characterize and quantify complex multimodal (visual, speech, audio) temporal patterns producing objectification in films. 
Building on film studies and psychology, we define the construct of objectification in a structured thesaurus involving 5 sub-constructs manifesting through 11 concepts spanning 3 modalities. We introduce the Multimodal Objectifying Gaze (MObyGaze) dataset, made of 20 movies annotated densely by experts for objectification levels and concepts over freely delimited segments: it amounts to 6072 segments over 43 hours of video with fine-grained localization and categorization.
We formulate new video interpretation tasks, show the feasibility of multimodal objectification detection, and analyze data and model bias to propose improvements. We exemplify two applications of MObyGaze, showing how rich concept annotation can improve model reliability and explainability.
We make our code and our dataset available to the community and described in the Croissant format: \url{https://github.com/husky-helen/MObyGaze}
\end{abstract}

\begin{keywords}
  video interpretation, objectification, multimodality, explainability
\end{keywords}

\section{Introduction}

While audiovisual storytelling contents have been shown to strongly shape our perception of sociological constructs (see, e.g., \cite{ward_media_2020}), such as gender, race and others, disparities in on-screen representation persist, particularly 
between genders. 
Grasping subtle patterns of disparities in gender portrayal requires understanding how the content produces different perceptions of the characters, beyond quantifying gender presence. In film studies, this question has been the subject of numerous qualitative analyses, and the concept of \textit{male gaze} was introduced by \cite{mulvey1975} and recently revisited by \cite{brey2020regard}. Male gaze refers to the way the content can be composed to produce objectification, i.e., so that a character is perceived more as an object, often of desire, than a subject of action.
\revised{Male gaze has a strong gendered component. Mulvey's 1975 account in introducing the idea of male gaze, is explicitly a feminist psychoanalytic argument about how Hollywood cinema structures spectatorship around a masculine heterosexual position. The asymmetry Mulvey identifies is not incidental — cinema, in her argument, routinely aligns three positions (the camera, the male protagonist, and the spectator) as the bearer of an active, desiring look, while coding the woman on screen for her "to-be-looked-at-ness": she is seen, he sees and acts.
Brey's work sets out from the observation that the visual apparatus gives women the function of spectacle that interrupts or decorates that plot. These techniques are not gender-neutral formal devices — historically and statistically they are applied differentially to women's bodies, converting the female character from a subject the audience is invited to identify with into a surface the audience (and the male characters, and the camera-as-proxy-viewer) is invited to look at.}

\begin{figure*}[t]
  \centering
  \includegraphics[width=0.9\linewidth]{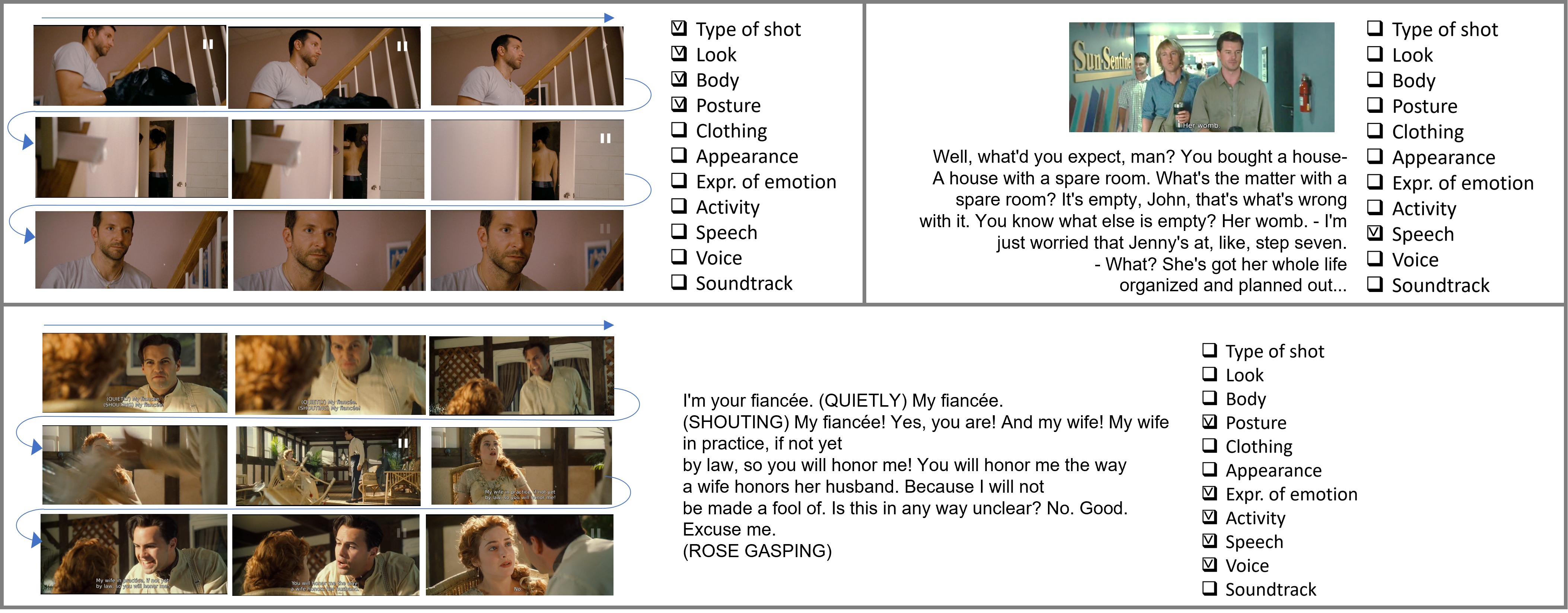}
  \caption{Examples of segments tagged with a \textit{Sure} level of objectification. Top left: vision modality only. Top right: text modality only. Bottom: multimodal concepts producing objectification.}
  \label{fig:example}
\end{figure*}

But \textbf{how is objectification produced by the content?}
This involves deliberate filmmaking choices to compose the audiovisual content that unfolds over time, such as:
What is the camera perspective? Who are the viewers looking at, whom does the camera embody, and how are characters portrayed? What are the dialogue dynamics? Who is talking, to whom, of whom, about what, and how?
Fig. \ref{fig:example} shows an example of how objectification is produced through the combination of camera position, character gaze, posture, speech, and voice.
Recent advances in multimodal media computing provide new approaches that may be wielded to characterize complex temporal and multimodal patterns of objectification, and quantify them.

Towards this goal, we introduce a new AI task: characterizing and quantifying how complex multimodal (visual, speech, audio) discursive patterns produce objectification in film.
So far, interpretive tasks have only been thoroughly studied in the text modality, with approaches for hate speech detection and beyond, incorporating subtle aspects such as sexism (e.g., \cite{samory_call_nodate}).
Approaches for the visual modality are scarce and mostly limited to still images (\cite{fersini_detecting_2019, kiela_hatefulmeme_2020, wang_hateclipseg_2025}). 
We therefore contribute the necessary elements to make this new interpretive multimodal task accessible to the machine learning community. \textbf{Our contributions are}:

\noindent$\bullet$ the Multimodal Objectifying Gaze (MObyGaze) dataset. 
For this, we devise a thesaurus of objectification by building on existing characterization in film studies and cognitive and social psychology. 
The thesaurus articulates visual, speech and audio components, which we denote as \textit{concepts} involved in the production of objectification. 
The annotation process then consists in 2 experts densely annotating 20 movies: they manually delimit all the segments (unitization) they find relevant for objectification, and label each with a level of objectification (categorization). 
To allow fine-grained data and model analysis, they also annotate which objectification concepts are present, and indicate the classification difficulty (with a \textit{hard negative} label). 
We verify the validity of the produced data with annotator agreement measures for both unitization and categorization.
The resulting dataset comprises 6072 
segments over 43 hours of 20 films each annotated by 2 experts.

\noindent$\bullet$ a new video interpretive task to detect objectification in films. We study and show the feasibility of the task for the individual concept modalities and with multimodal models. We analyze disaggregated model performance, unveil the contextuality of the annotation process, and show how to improve model generalization.

\noindent$\bullet$ two applications of MObyGaze where we show how rich concept annotation can improve multimodal models, how the new video interpretation task of objectification detection raises new challenges for explainable AI methods, and the approaches we propose.

We make our dataset available to the community, with Datasheet documentation (\cite{datasheet}) and Croissant metadata for file and recordset descriptions (\cite{croissant}). 
We believe this dataset is valuable for advancing computational approaches to make subtle patterns of bias in audiovisual content visible and more tangible, and quantify their prevalence. 

The article is organized as follows. Sec. \ref{sec:related_works} positions our contributions in the context of related works. Sec. \ref{sec:dataset} presents the MObyGaze dataset, its creation and analysis. 
Sec. \ref{sec:experiments} presents the study of the feasibility of multimodal objectification detection, and data and performance analysis to improve models in terms of accuracy and explainability.
Sec. \ref{sec:lims_appls} discusses the limitations and broader impact. 

\section{Related work}
\label{sec:related_works}

We position our contributions with respect to works on: analyses of biases in film datasets, annotation of audiovisual and multimodal contents, and dataset creation for interpretive tasks. 

\paragraph{Bias analysis in film datasets}
Disparities in representing different groups of characters in films have been computationally quantified with analyses of low-level characteristics of the visual (\cite{guha_gender_2015,mazieres_computational_2021,jang_quantification_2019}) or textual data.
%
%
%
\cite{jang_quantification_2019} considered 
20 films and show that women characters have a lower spatial and temporal occupancy
, corroborating findings on vision datasets by \cite{wang_revise}.
\cite{somandepalli_computational_2021} show in 1000 movie scripts that female characters appear more often as victims. \cite{schofield_gender-distinguishing_2016} characterize differences in linguistic markers in dialogue utterances of female and male characters.
\cite{agarwal_key_2015}
propose a way to automate the Bechdel test from computationally analyzing 457 film scripts with their pre-existing annotations of Bechdel test results made by volunteers and hosted in a public website.
\cite{martinez_boys_2022}
collected 912 movie scripts to investigate differences in how different genders are associated to different types of actions. 
They show that male characters are generally given more agency than female characters, 
and that female characters are more often the object of the gaze of male characters, with verbs reflecting their sexual objectification.
%
These last two relied on non-expert human annotation only in the text modality, and without considering high-level interpretive constructs as we do in our work.
Neither relied on the analysis of visual or audio data.

\paragraph{Annotation of audiovisual and multimodal content}
\revised{We present in Table \ref{table:dataset-comparison} a synthetic view of the comparative discussion below.}
%
Video annotation is considered a heavier task than image annotation, and has therefore been mostly considered for short videos, notably 
for action recognition and video anomaly detection.
%
For example, the ActivityNet benchmark (\cite{heilbron_activitynet_2015}) comprises ca. 27000 videos lasting ca. 2 minutes in average, representing 203 activity classes, 
crowdworkers annotating the temporal boundaries of each action instance. 
%
Video anomaly is a more interpretive construct, with 
classes such as abuse, assault, or robbery. For example, \cite{sultani2019realworld} introduce the UCF-Crime video anomaly dataset of 1900 videos of ca. 4 minutes each, categorized into 13 anomaly classes. 
Temporal delimitation is costly and variable from annotator to annotator. For this reason, a lot of video anomaly detection data (including the training set of UCF-Crime) are only annotated for classes at the video level, hence requiring weakly-supervised learning approaches to localize and categorize anomaly. 
%

Detecting and quantifying hateful multimodal content is crucial to train less harmful foundation models with large-scale image-text datasets, as investigated by \cite{birhane_into_2023}. 
Yet, manual annotations of multimodal content remain scarce.
\cite{kiela_hatefulmeme_2020} created a dataset of hateful memes, while \cite{fersini_detecting_2019} 
specifically considered sexist memes and advertisement imagery. 
More recently, \cite{das_hatemm_2023} and \cite{wang_hateclipseg_2025} have introduced dataset for multimodal hate speech detection in short videos. Specifically \cite{wang_hateclipseg_2025} have created HateClipSeg, a dataset of 435 short videos (3 to 10 mins) segmented and annotated for Normal and Offensive content, with five sub-classes of Offensive including Sexual and Hateful. 
The dataset neither annotates individual modalities nor explanatory concepts, as we do with MObyGaze. Their work shows that existing models have significant room for improvement on such interpretive video tasks, as we do but on a more specific task (detecting objectification) defined structurally with a thesaurus. 
\cite{bose_mm-autowards_2023} introduced MM-AU, a dataset of 8.4K advertisement videos annotated for three tasks including an interpretive one of so-called social message detection, but with a coarse annotation approach: only the presence of a social message is annotated, not its nature, and neither finer-grained concepts nor individual modalities are annotated. We, however, consider similar multimodal baselines in Sec.~\ref{sec:mm} and \ref{sec:ICIPElisa}.

Movie datasets are usually not annotated manually. A prominent exception is MovieGraphs, introduced by \cite{vicol_moviegraphs_2018}, for which freelance workers were recruited to annotate 51 movies with time-grounded graph-based annotations of character relationships and interactions. 
Owing to the richness of MovieGraphs and the possible relevance of leveraging the original graph annotations to detect objectification, we select 20 out of the 51 movies from MovieGraphs reproducing the same distribution of genres, to be densely annotated for the construct of multimodal objectification. 
%

\revised{Our endeavor of film annotation for objectification therefore has distinct objectives and challenges: 1) the nature of the data source is artistic and story-rich, on which we envision tasks beyond human activity recognition, 2) our tasks require the understanding of complex multimodal event cues (language, visual, audio, etc.) over long prediction horizons (a few minutes), 3) we investigate the social construct of objectification in films, that is not intended for other objectives (e.g., security) that require different types of data and annotations.}

\paragraph{Dataset creation for interpretive tasks}
Like \cite{kiela_hatefulmeme_2020}) and \cite{fersini_detecting_2019} who considered hateful and sexist meme annotation, most of the works mentioned above do not provide a detailed definition of the high-level construct to annotate, rather giving annotators freedom to interpret the term.
In contrast, systematic approaches for rigorous definition of high-level constructs are more common in NLP. 
%
\cite{samory_call_nodate} identified how the lack of proper definition of a high-level construct such as sexism 
impedes proper data analysis. They therefore proposed to leverage questionnaires introduced and validated in social psychology to produce a codebook to assess different dimensions of sexism. They then employed crowdworkers, trained on the codebook, to annotate tweets.   
%
In a similar objective, \cite{da_san_martino_fine-grained_2019} approached the difficulties of annotating propaganda in news articles by identifying 18 propaganda techniques from the existing literature. To avoid political views to excessively noise annotation, they had 4 experts localize and classify relevant text-spans. 
%
Dense annotation by a single expert has also been proposed for medical images by \cite{daneshjou2022skincon}, to assign skin lesion images with an end label of \textit{benign} or \textit{malignant}, along with which of the 48 clinical explanatory concepts are present. 

In this article, we take inspiration from these codebook-based annotation of high-level constructs by \cite{samory_call_nodate}, \cite{da_san_martino_fine-grained_2019} and \cite{daneshjou2022skincon} to approach the creation of data for the multimodal construct of objectification in a systematic and multi-disciplinary way. We leverage existing literature in psychology, cinematography and gender media studies to define a thesaurus, identifying concepts to be annotated by experts, who will annotate feature-length films (2h08min of average duration) with time delimitation (unitization) and categorization of the perceived level of objectification. 
To the best of our knowledge, this is the first time a dataset of audiovisual content is annotated for a high-level construct -- objectification -- defined in a thesaurus of multimodal concepts, with freely delimited timespans. 
In line with approaches advocated by, e.g., \cite{paullada_data_2021}, our purpose is to produce a non-large scale but high-quality dataset enabling efficient model training and data analysis to contribute to unveiling how subtle representation disparities in audiovisual contents may persist. 

\paragraph{Positioning with the works on this dataset}
\revised{A pre-print of this article has been on arxiv since early 2025. An early and preliminary version of the dataset was published at CVPR 2024 by \cite{tores_visual_2024} and included only visual concepts and only 12 movies. The article consisted in showing the feasibility of the task from a simpler and somewhat noisier perspective. First, all annotations were aligned onto the MovieGraphs clip spans, aggregating the annotators' labels by max pulling the objectification levels and concatenating the concept annotations. Second, the splits were only across clips, meaning that different clips of the same movie could appear in train, validation and test. Besides, solely models based on MLP and CBMs were tested.
After this preliminary work, we identified the need to augment the dataset to: (1) incorporate multimodal concepts such as speech and sound, and (2) obtain stronger representation of all concepts from which we can more easily sample for training various classification tasks and models. 
As mentioned above, we subsequently started to inverstigate ML challenges raised by the MObyGaze data: how to tackle interpretive video tasks, how to benefit from expert modality-wise annotations to improve multimodal models (\cite{MMelisa}, \cite{elisaArxivICIP}), 
how to improve concept-based models (\cite{julieACMMM}).
While the core contribution of the present article is the complete dataset workflow (with its production, analysis, the principled proof of feasibility of multimodal objectification detection across movies, and corresponding model analyses), we believe the subsequent works above show the usefulness of MObyGaze as a dataset for ML research. That is why we have dedicated Sec. \ref{sec:interest} to two of these works.}

\begin{table}[!h]
\centering
\caption{\revised{Comparison of MObyGaze with other related datasets across domain, task interpretativity, scale, and annotation approach}}
\label{table:dataset-comparison}
{\scriptsize
\begin{tabular}{l c c c l c c}
\toprule
\textbf{Domain} & \makecell[c]{\textbf{Interpretive}\\\textbf{task}} & \makecell[c]{\textbf{Avg.}\\\textbf{duration}} & \textbf{Total hours} & \makecell[c]{\textbf{Annotators}\\\textbf{/content}} & \textbf{Unitizing} & \makecell[c]{\textbf{Explanatory}\\\textbf{concepts}}\\
\cmidrule(rl){1-7}
ActivityNet & $--$ & 18 min & 849 & crowd, unspec. & \checkmark & -- \\
UCF-Crime & $-$ & 4 min & 128 & unspec. & \checkmark & -- \\
HateMM & $+$ & 4 min & 43 & \makecell{4 novices trained\\ by 2 experts} & \checkmark & --\\
HateClipSeg & $+$ & 4 min & 26 & 2 novices & -- & -- \\
\cmidrule(rl){1-7}
\textbf{MObyGaze} (us) & $++$ & 128 min & 43 & 2 experts & \checkmark & \checkmark\\
\bottomrule
\end{tabular}
}
\end{table}
\section{Dataset}
\label{sec:dataset}
We first present our definition of the construct of objectification in a structured thesaurus and describe the annotation process. We then analyze the obtained data by validating its consistency and showing key characteristics.

\subsection{Dataset composition}\label{sec:dataset_composition}
\begin{figure*}[!ht]
  \centering
  \includegraphics[width=1.0\linewidth]{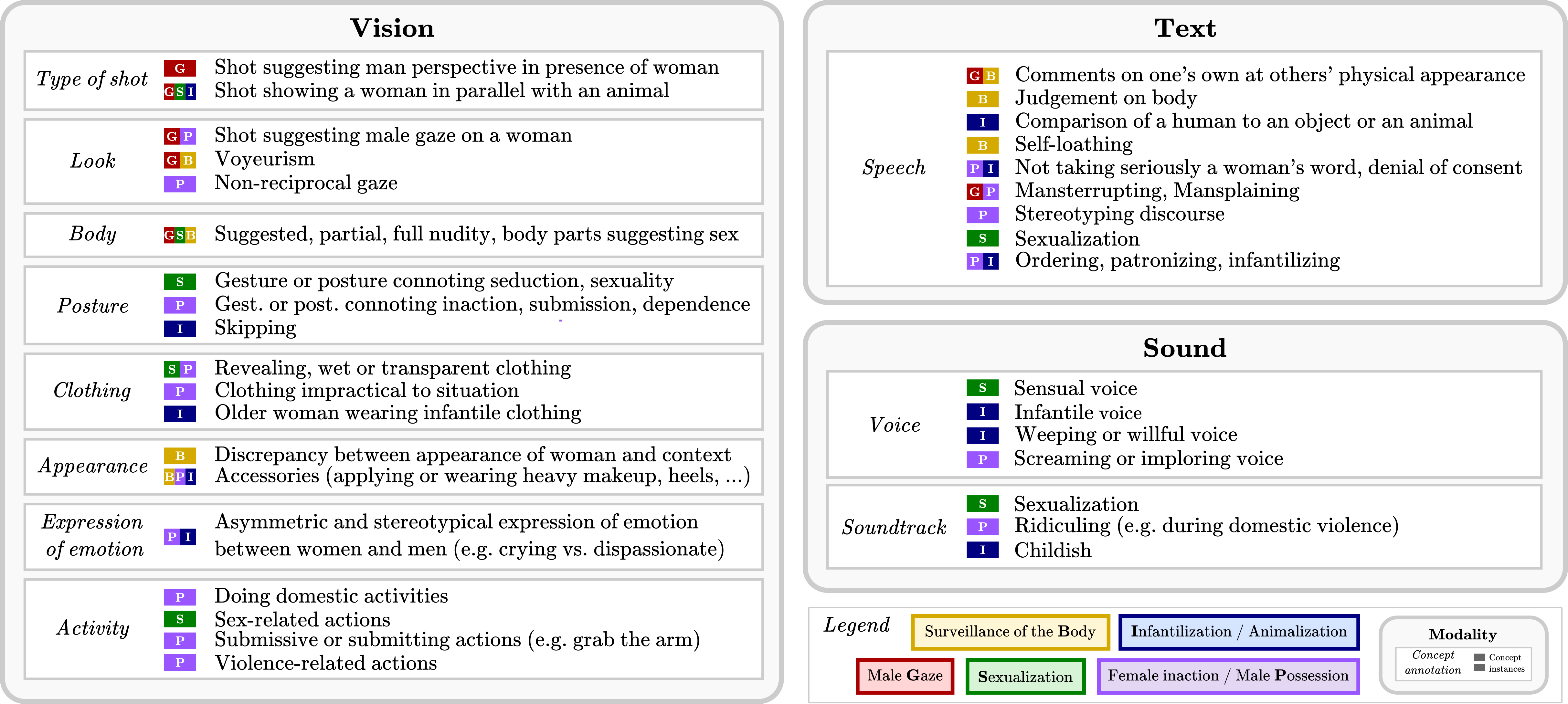}
  \caption{Thesaurus for the construct of objectification: 5 \fcolorbox{Bisque4}{AntiqueWhite1}{Sub-constructs} (bottom right) manifested through 11 \textit{\textbf{Concepts}} spanning 3 \textbf{Modalities}. Best seen in color. A color-blind version is available in App. \ref{app:thes_colorblind}.}
  \label{fig_thes}
\end{figure*}

\paragraph{Thesaurus of multimodal objectification}
We set out from the concept of \textit{male gaze} introduced in film studies to describe the filmmaking choices producing a perception of women characters as objects in male-driven actions, intentions and perspectives. \revised{While some formalizations of gaze stem from the gender of the director \cite{malone_2018}, we privilege the formalizations of \cite{brey2020regard} who defines male gaze only from the film content.} \cite{brey2020regard} carries out a qualitative analysis of over 120 film and series scenes to describe complex temporal patterns involving filmic (framing, camera perspective and motion, etc.) and iconographic (whether the face is shown and how close, what body parts are shown, how the characters are dressed, what are their interactions) aspects, which either allow the audience to understand and engage with the experience of a character, or prevent the audience from doing so, hence de-humanizing, or \textit{objectifying}, the character. 
Objectification has also been investigated in social and cognitive psychology. Results and validated questionnaires study how the perception of objectification depends on various elements such as gaze and appearance (\cite{calogero_test_2004, calogero_operationalizing_2011, mckinley1996objectified}), clothing and posture (\cite{bernard_revealing_2019}), body parts (\cite{bernard_objectifying_2018}), sexualization (\cite{denchik2005development, bernard_sexualizationobjectification_2020}), interactions (\cite{gervais_social_2020}), actions (\cite{sap2017connotation}).
Put together, we identify 5 sub-constructs of objectification, shown in Fig. \ref{fig_thes} (bottom right): \fcolorbox{Firebrick3}{LightPink1}{Male Gaze} (G), i.e., point of view of a man on a woman (\cite{brey2020regard,mulvey1975,bernard_objectifying_2018}), \fcolorbox{Chartreuse4}{PaleGreen1}{Sexualization} (S) (\cite{bernard_sexualizationobjectification_2020,bernard_revealing_2019}), \fcolorbox{Goldenrod2}{LightGoldenrod1}{Surveillance of the feminine body} (B) (\cite{calogero_test_2004,mckinley1996objectified,denchik2005development}), \fcolorbox{Purple1}{Thistle2}{Female inaction and male possession} (P) (\cite{gervais_social_2020,sap2017connotation}), and \fcolorbox{DodgerBlue4}{LightSkyBlue2}{Infantilization and animalization} (I) (\cite{mulvey1975}).
From the questionnaires, experiences and analyses of these above works in film studies and psychology, we enumerate representative instances of each sub-construct, which we group into 11 concepts spanning 3 modalities, vision, text and sound, as depicted in Fig. \ref{fig_thes}.
The 5 sub-constructs manifest through several modalities. 
To align with the literature on explainable AI (\cite{chen_concept_2020, daneshjou2022skincon, zarlenga2022concept}), we use the term \textit{concepts} to denote the annotated factors motivating the rating of objectification. Annotating a segment with a concept means the annotator perceives an objectifying element, in this concept's dimension, that may contribute to objectification. 
\revised{The concept granularity is heterogeneous on purpose. Indeed, Vision Language Models (VLMs) and Multimodal LLMs (MLLMs) have been more brittle at analyzing complex video content (\cite{kesen_vilma_2024,asadi_mirage_2026}). Given that a higher number of concepts entails a higher annotation effort, we decided to obtain a much needed supervision in the visual domain from a finer-grained concept annotation, while saving effort on the textual domain.}

\paragraph{Data selection} As mentioned in Sec. \ref{sec:related_works}, we select movies from the MovieGraphs dataset (\cite{vicol_moviegraphs_2018}) owing to the richness of existing annotations on relationships and interactions. 
\revised{We hoped this choice makes it easier for the community to contribute to the task, as it has already been done by an external team (\cite{serouis2025towards}).}
We select 20 out of 51 movies, maintaining the distribution of genres (see App. \ref{sec:suppl_dataset} in the supplementary material for details). \revised{This was a round number that allowed us to attain the above objectives while keeping the task reasonably feasible by the two annotators within a month.}
\paragraph{Annotation} Each movie is annotated by 2 experts (with background in computer science, film studies and cognitive psychology). \revised{This choice stems from two factors: existing works with interpretive multimodal tasks adopted a similar approach (see our new Table 1 comparing the datasets), and the annotation effort (80+ hours including pilot on the annotation workflow and expertise).} 
The experts watch the movie entirely, and annotate the film by (1) setting temporal boundaries to create segments, and (2) for each such segment, rate the objectification on one of four levels:\\
\noindent$\bullet$ Easy Negative (EN): no objectifying concept is present;\\
\noindent$\bullet$ Hard Negative (HN): one or some concepts are present, are annotated, but are deemed insufficient to produce a perception of objectification;\\
\noindent$\bullet$ Sure (S): objectification is perceived and explained by the annotated concepts from the thesaurus;\\
\noindent$\bullet$ Not Sure (NS): objectification is perceived and concepts are annotated but the annotator considers they do not sufficiently explain the perception of objectification.\\
A custom annotation tool was made (see App. \ref{sec:suppl_dataset}). The annotation steps, including remediation and thesaurus refinement, are detailed in App. \ref{sec:suppl_dataset} and summarized below.
\revised{The annotators first annotated 2 movies. The obtained annotations were then aligned and colored for the annotators to identify their major divergences, exemplified in Fig. \ref{fig:example_alignment} in the appendix. They convened and identified that the agreement was generally high on the rating of objectification. Noticeable differences were in annotation of NS with levels of narratology spanning over more than a single segment. The thesaurus is indeed targeted at annotating salient concepts in segments, and we discuss this limitation in Sec. \ref{sec:lims_appls}.
They specifically identified under-determination of concepts Actitivies and Appearance. They expanded Activities to include all types of actions contributing to objectification, particularly momentary actions by a character onto another (including aspects of domination and violence). They trimmed the concept of Appearance, which initially consisted of instances deploying over the entire film (such as age of character not matching age or appearance of actress) to restrict it to scene-level features. Fig. \ref{fig_thes} shows the resulting thesaurus. They then carried out individually the annotation over the rest of the 20 movies.}
\paragraph{Data format} 
The resulting dataset is made available, described in ML Croissant format with Responsible AI properties, and detailed in a Datasheet document (\cite{datasheet}) in App. \ref{suppl:docu}.
%
\subsection{Dataset analysis}\label{sec:dataset_analysis}

\subsubsection{Validation of the data: \revised{internal consistency}}

\revised{We first compare the annotation data between each annotator. In the next section we compare them against external references.}

\paragraph{\revised{IAA on objectification}}
\revised{We adopt IAA measures recently introduced by \cite{braylan_measuring_2022} for complex multi-object labeling tasks.
To assess how two movie annotation sequences are aligned, we consider two types of distance function.
The first distance function between 2 movie annotations is 1-F1 score when a movie is considered as a set of tokens (equal to its number of frames) with 2 sets of labels. For each annotation considered as the gold/ground truth (GT) and the other as prediction, we obtain 2 F1 scores, then averaged. This distance therefore inherently incorporates agreement on unitizing. It has been introduced for Named Entity Recognition by \cite{braylan_measuring_2022}.
The second distance function we consider, and that may be more intuitive, is dist@IoUthresh, defined as the distance between two movie annotations when we consider the overlap between two segments must be greater than IoUthresh for these segments to be matched and their respective objectification or concept labels to be compared. A formal definition of this distance is provided in App. \ref{app:IAA}.}

\begin{figure}[!h]
    \centering
    \begin{subfigure}[b]{0.40\textwidth}
        \centering
        \includegraphics[width=\linewidth]{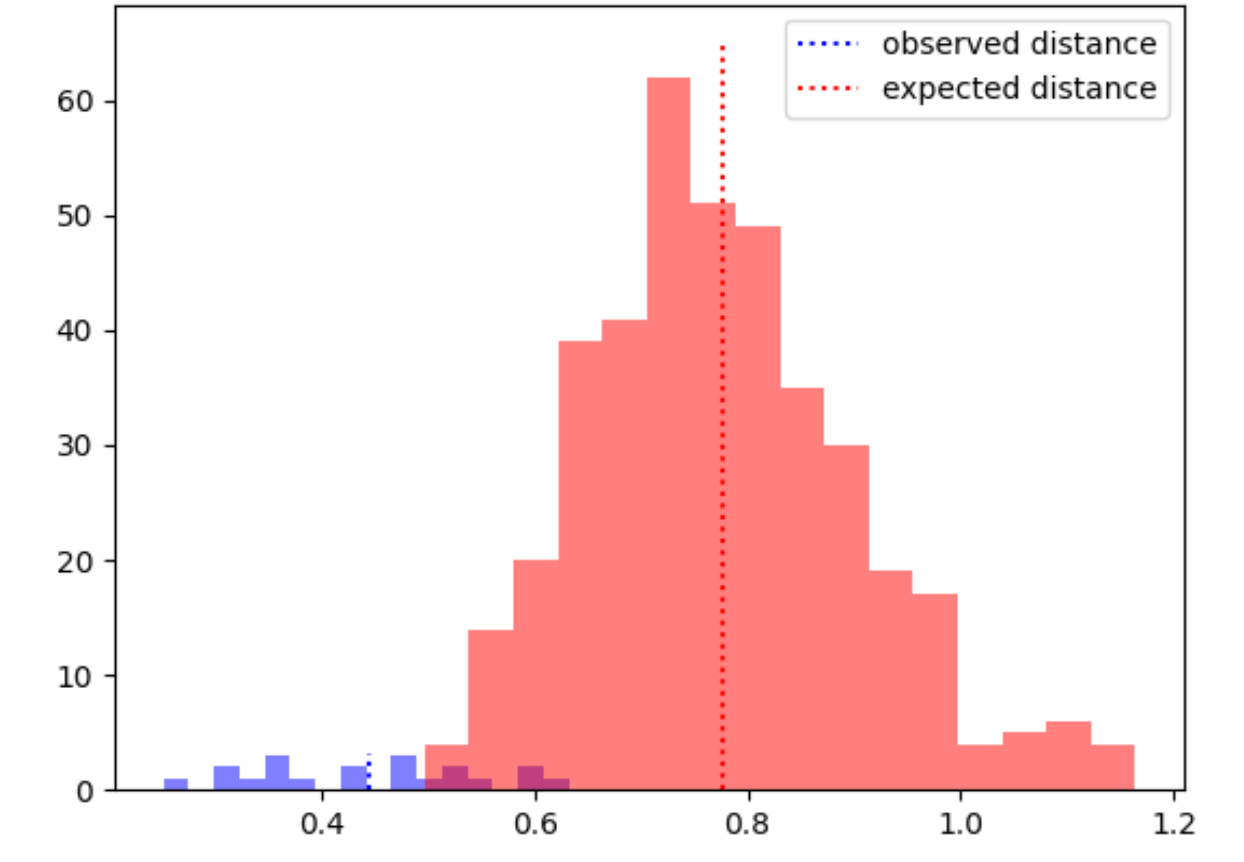}
        \caption{}
        \label{fig:distr_dists}
    \end{subfigure}
    \hfill
    \begin{subfigure}[b]{0.50\textwidth}
        \centering
        \includegraphics[width=1\linewidth]{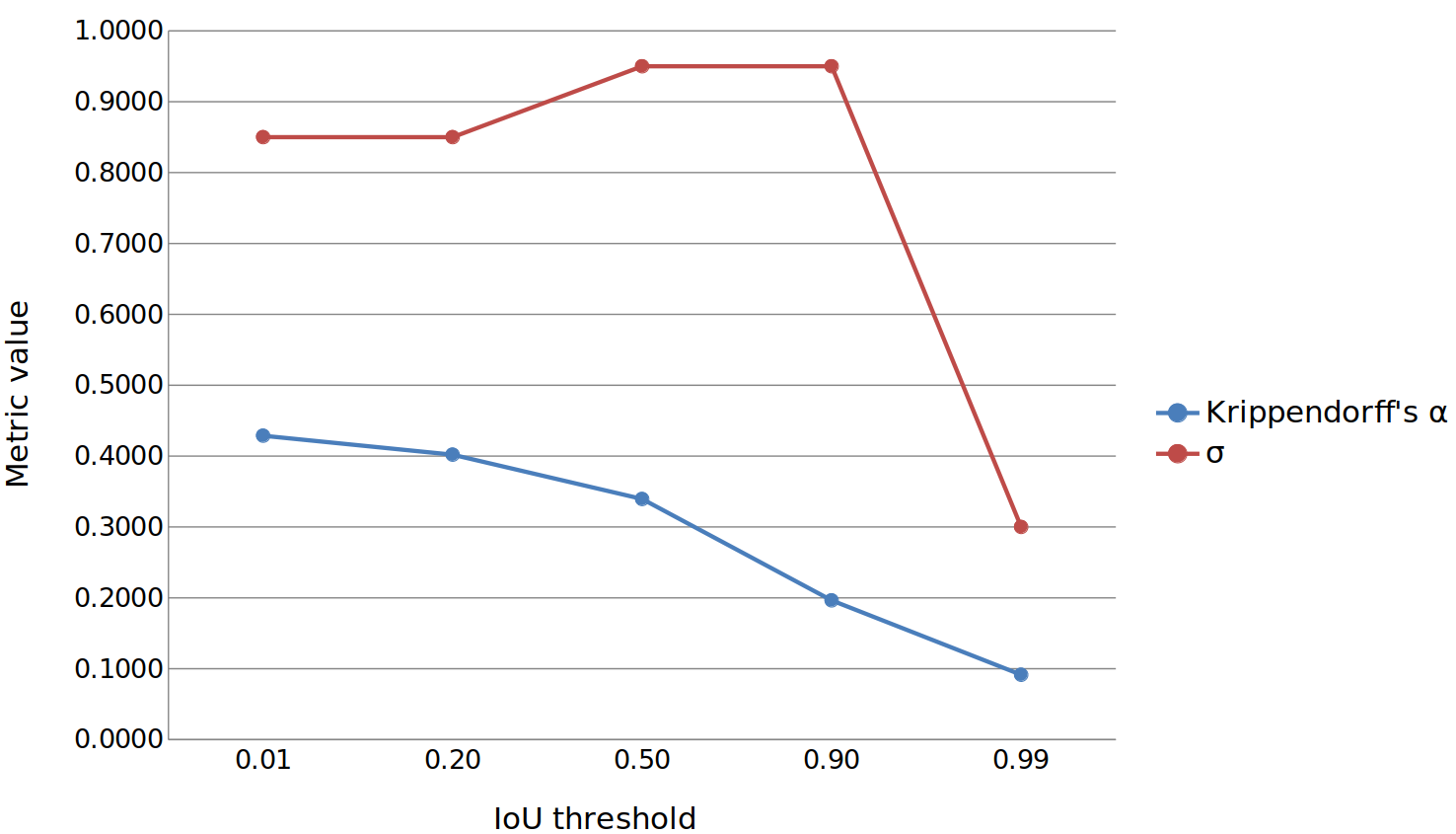}
        \caption{}
        \label{fig:iaa_vs_IoU}
    \end{subfigure}
    \caption{\revised{(a) Distribution of distances dist@IoUthresh between both annotators' annotations within-movies (observed distances) and inter-movies (expected distances), for IoUthresh=0.9. Krippendorff's $\alpha=0.20$, $\sigma=0.95$. (b) Evolution of Krippendorff's $\alpha$ and $\sigma$ as a function of IoUthresh.}}
    \label{fig:iaa_obj}
\end{figure}

\revised{Fig. \ref{fig:distr_dists} shows the distributions of the observed $d_o$ (blue) and expected $d_e$ (red) dist@IoUthresh distances for IoUthresh=0.9. $d_o$ and $d_e$ are the within-item (item = movie) and inter-item distances between 2 annotations, respectively. Chance correction is made by comparing $d_o$ samples with $d_e$ samples.
To perform this comparison, we consider two different IAA metrics: Krippendorff's $\alpha$ and $\sigma$ introduced by \cite{braylan_measuring_2022}. 
Krippendorff's $\alpha$ is  defined as $1-\bar{d_o}/\bar{d_e}$ with the averages of $d_o$ and $d_e$.
Metric $\sigma$ is the fraction of the observed distance samples that have a probability lower than 0.05 to have been drawned from the expected (chance) distance distribution.}

\revised{First we observe in Fig. \ref{fig:distr_dists} that the distributions are significantly apart, which is confirmed by $\sigma=0.95$. This is a strong indicator of the consitency of the annotators' annotations (also qualitatively exemplified in Fig. \ref{fig:example_alignment}). However, corresponding Krippendorff's $\alpha$ is 0.20 owing to the range of average distances (x-axis), and while Fig. \ref{fig:iaa_vs_IoU} also shows that $\alpha$ increases with lower IoU thresholds as expected, its highest value is 0.4. This is expected from the distance definition and the unitizing task itself, as investigated by \cite{braylan_measuring_2022}, and we provide a formal intuition with two toy examples in App. \ref{app:IAA}. They show how unitizing quickly noises the value of $\alpha$ and explain the partly opposite trends of $\alpha$ and $\sigma$ we observe in Fig. \ref{fig:iaa_vs_IoU}: increasing IoUthresh yields a monotonously decreasing $\alpha$ while $\sigma$ stays constant or slightly increases, before decreasing sharply after some very high threshold (above the boundary agreement rate), as expected.
Therefore, as an IAA metric should describe how much the overall data is better than chance, after observing the weakness of Kripendorff's $\alpha$@IoU to describe this phenomenon, we opt for reporting $\sigma$ instead.}
\revised{For the objectification labels (considering labels EN, HN and S and discarding NS amounting to less than 5\% of the data) and both distance definitions:} \\
\revised{$\bullet$ for distance NER based on F1 score of tokenized sequences: $\sigma=0.74$}\\
\revised{$\bullet$ for dist@IoUthresh=0.9, $\sigma=0.95$.}\\
\revised{These results show the consistency of objectification ratings between both expert annotators.}


\paragraph{\revised{IAA on concepts}}
\revised{We now look at the agreement on concept annotations. To ease interpretation, we report in Table \ref{table:concept_iaa} $\sigma$ (and $\alpha$) for dist@IoUthresh with IoUthresh=0.8. We first consider concepts individually with only two labels (either the concept is present in a segment annotation, or it is absent).}

\begin{table*}[!h]
\small
\centering
\caption{\revised{Inter-annotator agreement for individual concepts and concept groups, ordered by $\sigma$ (descending), computed on the dist@IoUthresh distances, with IoUthresh=0.8.}}
\label{table:concept_iaa}
{\scriptsize
\begin{tabular}{lcc}
\toprule
Concept/Axis & Krippendorff's $\alpha$ & $\sigma$ \\
\midrule
\multicolumn{3}{c}{Individual Concepts (ordered by $\sigma$, descending)} \\
\cmidrule(rl){1-3}
Body            & 0.241 & 0.85 \\
Posture         & 0.234 & 0.85 \\
Clothing        & 0.271 & 0.85 \\
Speech          & 0.240 & 0.65 \\
Exp of emotion  & 0.155 & 0.60 \\
\fcolorbox{Bisque4}{AntiqueWhite1}{Type of shot}    & 0.180 & 0.15 \\
Soundtrack      & 0.066 & 0.15 \\
Activities      & 0.165 & 0.10 \\
\fcolorbox{Bisque4}{AntiqueWhite1}{Look}            & 0.214 & 0.05 \\
Appearance      & 0.077 & 0.05 \\
Voice           & 0.155 & 0.00 \\
\midrule
\multicolumn{3}{c}{Concept groups} \\
\cmidrule(rl){1-3}
Body, Clothing         & 0.287 & 0.95 \\
Posture, Activities    & 0.235 & 0.90 \\
\fcolorbox{Bisque4}{AntiqueWhite1}{Type of shot, Look}     & 0.196 & 0.80 \\
Speech                 & 0.240 & 0.65 \\
Voice, Soundtrack      & 0.134 & 0.00 \\
\bottomrule
\end{tabular}
}
\end{table*}

\revised{First, we observe that there is a high ($\sigma\geq 0.8$) or non-trivial ($\sigma\geq 0.6$) agreement for 5 in 11 concepts taken individually (Body, Posture, Clothing, Speech, Expression of an emotion), while the others have a significantly lower $\sigma$. Discussion between annotators after the annotation of two pilot movies made appear that the differences in annotated concepts often do not correspond to disagreement, but rather to some concept samples being sometimes overlooked by either one of the annotators. As a side note, it is also interesting to observe (e.g., for Look) that $\alpha$ can be greater than $\sigma$.}

\revised{Second, we build on the PCA presented below in Sec. \ref{sec:PCA} and analyze IAA not on individual concepts but on groups based on the PCA axes and on the modalities. The concept group is tagged present when at least one concept of the group is present. We consider 5 groups as shown in the lower part in Table \ref{table:concept_iaa}. While grouping some well agreed-on concepts slightly increases IAA wrt to the individual case, it dramatically increases for concepts that have individually low IAA but which are important and well-represented concepts appearing both in the PCA and the decision tree below as driving objectification: Type of shot and Look have $\sigma$ increasing from 0.15 and 0.05 when taken individually, to 0.8 when grouped. This shows that they are connected and the annotators agree to annotate the group consistently. Here for example, Type of shot is defined with camera placement while Look is defined with how one character is made to look at another. Cinematography often embodies the Type of Shot by placing the camera at a character's perspective for the viewer to adopt their Look. This explains the consistency of the annotation despite the non-systematic but frequent occurrence yielding possible ambiguity in what exact concept to annotate.}

\subsubsection{Validation of the data: \revised{external consistency}}\label{sec:ext_cons}

\revised{We further validate the annotations of the multimodal concepts against external measures. For speech and audio, we adopt the same approach as for the HateMM dataset by \cite{das_hatemm_2023}.}

\paragraph{Visual concepts}
\revised{For the visual concepts, we define a concept score for each individual segment, for each of the 8 visual concepts. This score is a similarity between the X-CLIP visual embedding of that segment and the textual descriptions of the concept. Textual descriptions of the concepts are provided in Fig. \ref{fig:vis_concept_prompt} in the appendix. Specifically, we do contrastive scoring precisely defined in App. \ref{app:external_cons}.
For each of the 8 concepts, we analyze whether the concept score significantly differ between segments with that concept annotated present ("annotated") and segments without the concept annotated ("not annotated").
We use a two-sided Mann-Whitney U test to compare both groups and produce a p-value per concept. Since 8 separate tests are run (one per concept), a Benjamini-Hochberg False Discovery Rate (FDR) correction is applied across the 8 p-values, producing a q-value for each. This controls the expected proportion of false positives among the concepts ultimately called "significant".}

\revised{Table \ref{table:vis_cons} reports the mean difference of scores, the rank biserial correlation $r_{rb}$, which can be interpreted as an effect size of how consistently one group's scores exceed the other's, ranging from $-1$ to $+1$. It is independent of the sample size. We also report the level of effect size as \textit{negligible} ($r_{rb}<0.1$), \textit{small} ($0.1\leq r_{rb} <0.3$), \textit{medium} ($0.3\leq r_{rb} <0.5$), and \textit{large}  ($r_{rb}\geq 0.5$).
We observe that for both annotators, and for 7 visual concepts in 8, there is a statistically significant difference in concept score between segments annotated with or without the concept. 
The effect size is negligible for only one concept for both annotators (concept Look).
This is hence another indicator of the consistency of the annotations, for the visual concepts.}

\begin{table}[htbp]
\centering
\caption{\revised{Comparison of the difference in visual concept score for the segments annotated with each concept and those not annotated, for each annotator (quantile 0.9 is used for intra-segment similarity).}}
\label{table:vis_cons}
\resizebox{\textwidth}{!}{%
\begin{tabular}{l ccccccc ccccccc}
\toprule
& \multicolumn{7}{c}{Annotator 1} & \multicolumn{7}{c}{Annotator 2} \\
\cmidrule(lr){2-8} \cmidrule(lr){9-15}
Concept & diff & $\mathbf{r_{rb}}$ & \textbf{effect size} & $n_{\text{ann}}$ & $n_{\lnot\text{ann}}$ & $q$-value & sig. & diff & $\mathbf{r_{rb}}$ & \textbf{effect size} & $n_{\text{ann}}$ & $n_{\lnot\text{ann}}$ & $q$-value & sig. \\
\midrule
Body            & 0.0245 & 0.402 & medium     & 213 & 2056 & $3.38\times10^{-21}$ & *** & 0.0179 & 0.271 & small      & 833 & 2970 & $1.44\times10^{-32}$ & *** \\
Clothing         & 0.0125 & 0.201 & small      & 433 & 1836 & $1.75\times10^{-10}$ & *** & 0.0149 & 0.214 & small      & 674 & 3129 & $4.41\times10^{-18}$ & *** \\
Type of shot     & 0.0131 & 0.197 & small      & 165 & 2104 & $3.33\times10^{-5}$  & *** & 0.0104 & 0.174 & small      & 565 & 3238 & $5.51\times10^{-11}$ & *** \\
Look             & 0.0017 & 0.015 & negligible & 334 & 1935 & 0.654                & --  & 0.0064 & 0.090 & negligible & 337 & 3466 & 0.00604              & *   \\
Posture          & 0.0181 & 0.293 & small      & 248 & 2021 & $2.01\times10^{-13}$ & *** & 0.0203 & 0.314 & medium     & 677 & 3126 & $5.59\times10^{-37}$ & *** \\
Appearance       & 0.0136 & 0.230 & small      & 127 & 2142 & $2.04\times10^{-5}$  & *** & 0.0191 & 0.283 & small      & 186 & 3617 & $7.66\times10^{-11}$ & *** \\
Exp.\ of emotion & 0.0116 & 0.154 & small      & 251 & 2018 & $8.02\times10^{-5}$  & *** & 0.0317 & 0.346 & medium     & 308 & 3495 & $1.28\times10^{-23}$ & *** \\
Activities       & 0.0223 & 0.235 & small      & 178 & 2091 & $3.52\times10^{-7}$  & *** & 0.0413 & 0.403 & medium     & 438 & 3365 & $5.31\times10^{-42}$ & *** \\
\bottomrule
\end{tabular}%
}
\end{table}

\paragraph{Textual speech concept}
\revised{For speech, we use Empath by \cite{fast_empath_2016} to analyze correlation of relevant Empath categories with concept Speech. We run Empath on the subtitle text of each segment, using a curated subset of Empath categories relevant to objectification (body/appearance, power/dominance, emotion, speech/communication), instead of the full 194 categories. The retained categories are shown in Fig. \ref{fig:speech_concept_prompt} in the appendix.
For each Empath category, we compare the per-segment normalized Empath score between the two groups. As earlier, we use the Mann-Whitney U test and a Benjamini-Hochberg False Discovery Rate (FDR) correction since we are testing 71 categories at once.}

\begin{figure}[!h]
    \centering
    \begin{subfigure}[b]{0.48\textwidth}
        \centering
        \includegraphics[width=\linewidth]{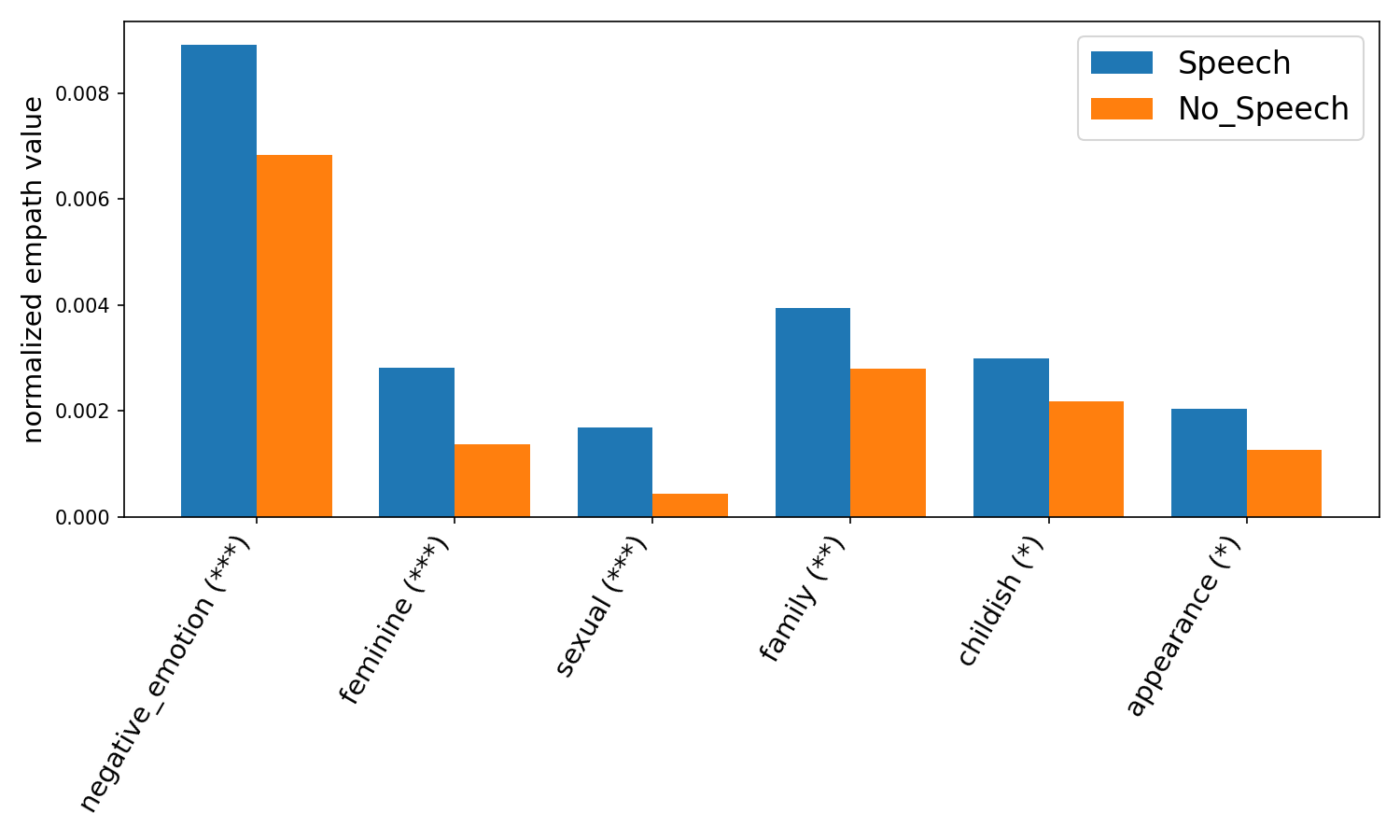}
        \caption{}
        \label{fig:speech_cons_ann1}
    \end{subfigure}
    \hfill
    \begin{subfigure}[b]{0.48\textwidth}
        \centering
        \includegraphics[width=1\linewidth]{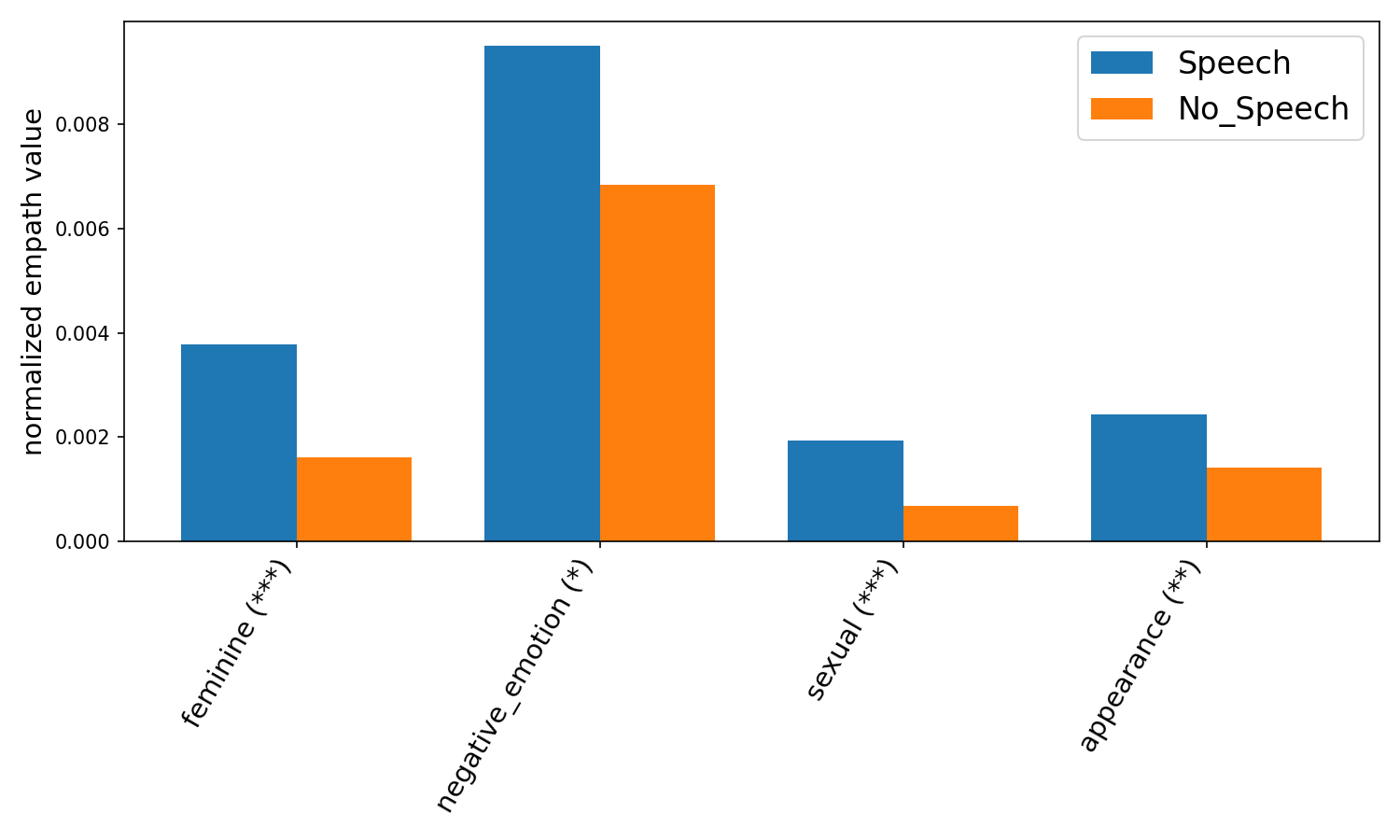}
        \caption{}
        \label{fig:speech_cons_ann2}
    \end{subfigure}
    \caption{\revised{Comparative Empath scores for the curated categories for Speech vs. No-Speech concept (the FDR-corrected q-value stars are indicated. (a) Data of annotator 1. (b) Data of annotator 2.}}
    \label{fig:speech_cons}
\end{figure}

\revised{Fig. \ref{fig:speech_cons} shows, for each annotator, the top N lexical categories with the most significant score differences.
We observe that the categories related to feminine, negative emotion, sexual and appearance are among the ones with most significant scores across both annotator. We notice that annotator 1 has additional significant categories, family and childish, indicating sensible difference in sensitivity of annotation between both expert annotators.}

\paragraph{Audio concepts}

\begin{figure*}[]
  \centering
  \includegraphics[width=0.95\linewidth]{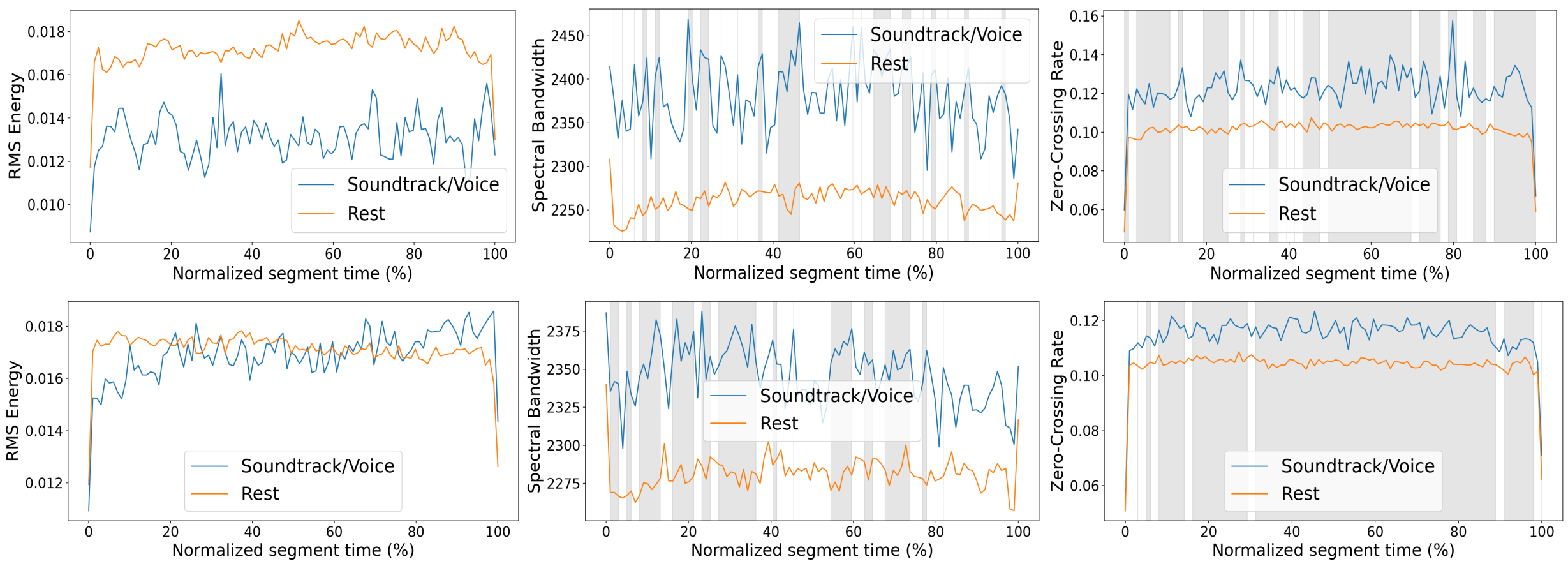}
  \caption{\revised{Audio features RMS (left), Spectral bandwidth (center), ZCR (right), shown over a normalized segment duration for each annotator: annotator 1 in top-row, annotator 2 in bottom-row (shaded = FDR q<0.01).}}
  \label{fig:audio_cons}
\end{figure*}

\revised{We then compare differences between segments annotated with an audio (voice or soundtrack) concept and those without, over three audio features: Zero-Crossing Rate (ZCR), Spectral Bandwidth, and Root Mean Square (RMS) Energy.
The three features are computed with librosa as per-frame time series, then each time series is resampled onto a common normalized time axis (0-100\% of the segment's duration) so that curves from segments of different durations can be averaged together. They are shown for both annotators (row-wise) in Fig. \ref{fig:audio_cons}. A pointwise Mann-Whitney U test at each of the normalized-time bins, FDR-corrected across bins, with significant bins shaded on the time-series plot.}

\revised{First, we observe with spectral bandwidth that annotators have selected audio concepts for significantly higher-pitched track. This may be due to aspect of sensual, feminine or infantile voices specified in the thesaurus. Second, we also observe that ZCR, that can be interpreted as a measure of the noisiness of a signal, is significantly higher in segments annotated with an audio concept. This may be due to "sensual", "weeping" voices in the thesaurus, that are more unvoiced and hence have a higher ZCR. Finally, RMS can help estimate the average loudness of an audio track. For annotator 1, we observe that tracks with the audio concept are less loud than those without, while the difference is not significant for annotator 2.} 

\revised{In conclusion, this audio analysis also indicates significant consistency both between the annotators and with external indicators related to the thesaurus. As for Look and Type of shot, the consistency in features despite the low IAA for the audio concepts may suggest aggregating the annotations may be relevant to lower the false negative rate.}

\subsubsection{Analysis of objectification levels independently from concepts}
\begin{figure}[!htb]
    \centering
    \includegraphics[width=\linewidth]{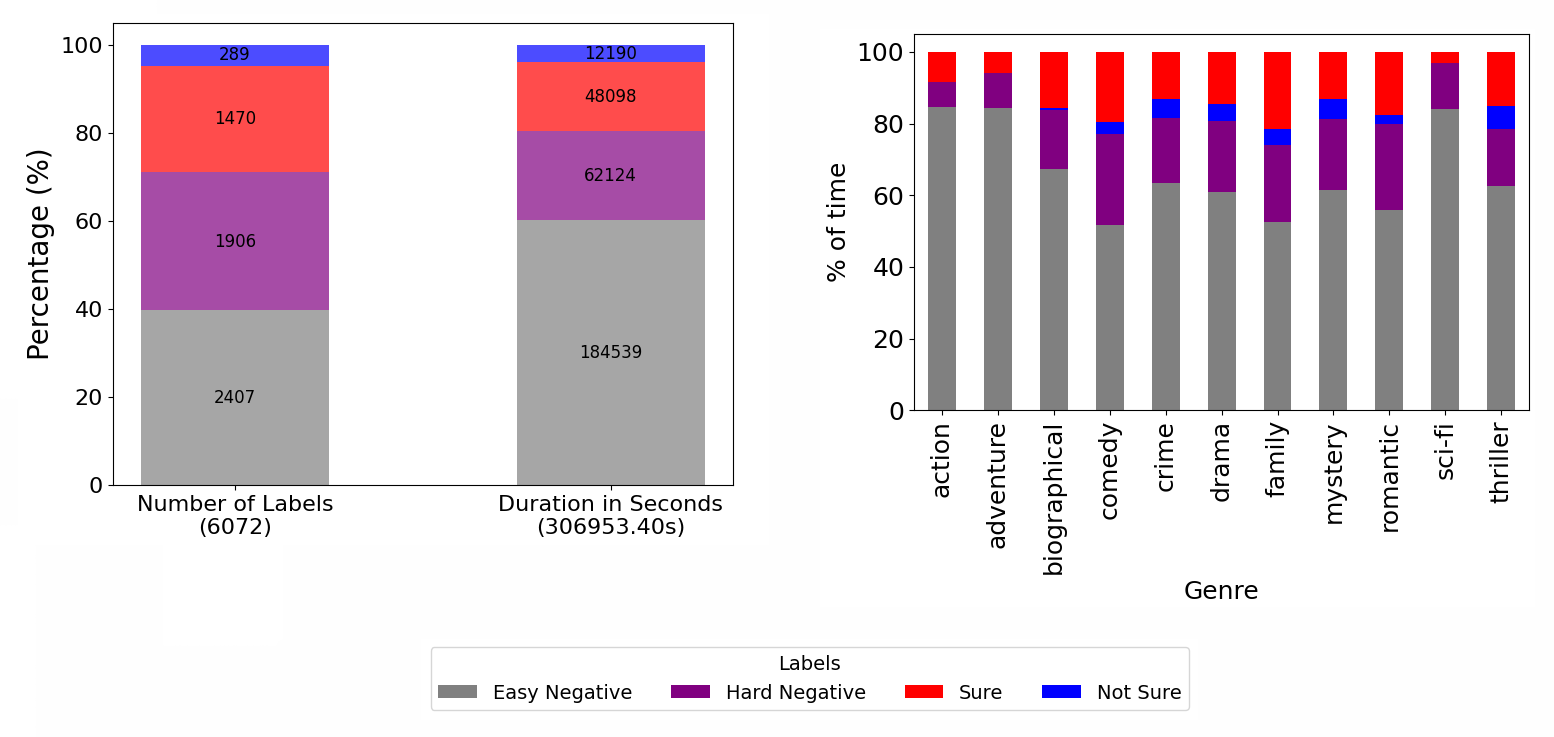}
    \caption{Descriptive analysis of the MObyGaze data. Left: distribution of label frequencies and corresponding durations. Right: Distribution of objectification levels over the film genres.}
    \label{fig:distr_obj}
\end{figure}

The 20 films make a total of 43 hours of footage annotated by two annotators, yielding 6072 delimited and annotated segments.
We first analyze the annotations of objectification levels, independently from concepts.
Fig. \ref{fig:distr_obj} (left) shows the distribution of objectification levels in number of occurrences and time duration. EN segments represent 39.7\% of segments and 60.1\% of total duration, HN 31.4\% and 20.2\%, and S segments 24.2\% and 15.7\%, respectively.
Fig. \ref{fig:distr_obj} (right) depicts the distribution of objectification levels over the film genres in the MObyGaze dataset (genres are shown in Table \ref{table:films} of the supplementary material): comedy, family and romance are the genres with most S and overall non-EN segments; action, adventure and sci-fi are those with the least.

\begin{figure}[!h]
    \centering
    \begin{subfigure}[b]{0.45\textwidth}
        \centering
        \includegraphics[width=\linewidth]{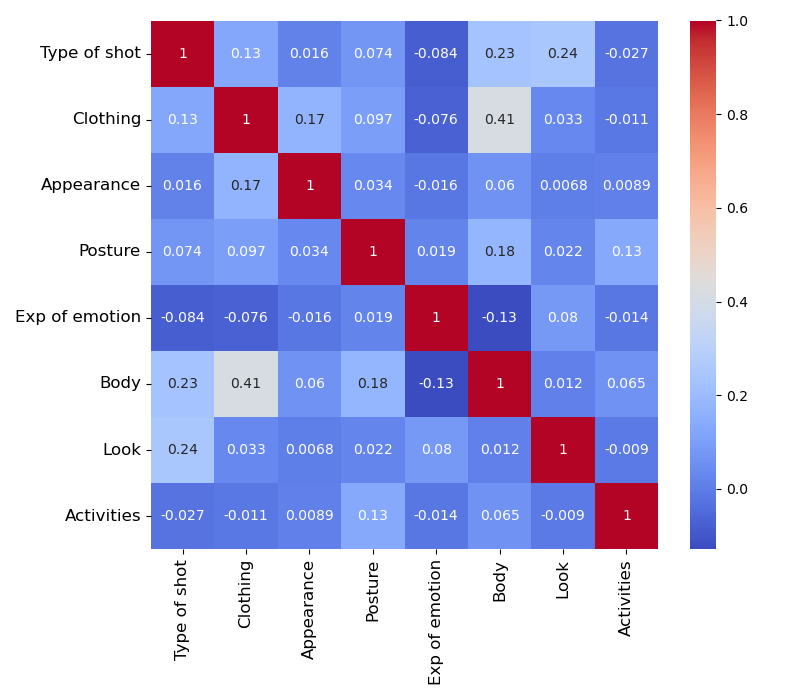}
        \caption{Cross-correlation matrix of concepts in the MObyGaze data.}
        \label{fig:subfig_corr_matrix}
    \end{subfigure}
    \hfill
    \begin{subfigure}[b]{0.50\textwidth}
        \centering
        \includegraphics[width=1\linewidth]{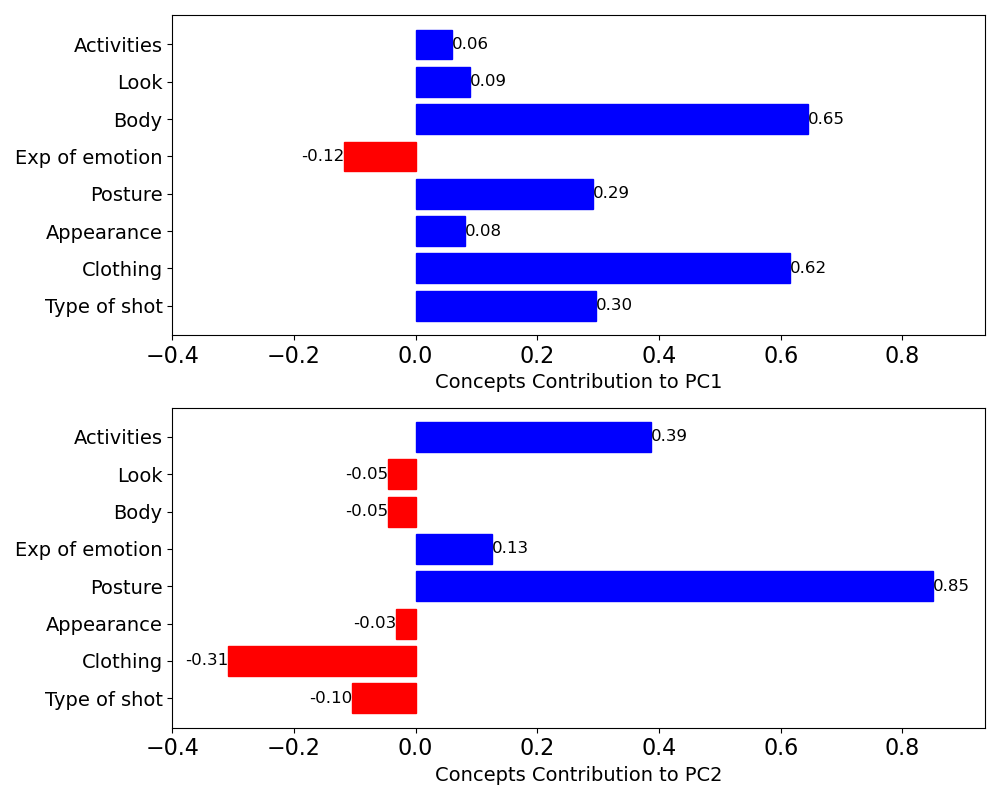}
        \caption{Contribution of each concept to the first two principal components (PCA), explaining most variance.}
        \label{fig:subfig_pca}
    \end{subfigure}
    \caption{(a) and (b): Overview of concept cross-correlation and principal component contributions in MObyGaze data.}
    \label{fig:concept_coocc_PCA}
\end{figure}

\subsubsection{Analysis of concepts independently from objectification levels}\label{sec:PCA}
Figure \ref{fig:subfig_corr_matrix} shows the cross-correlation matrix derived from the annotations of visual concepts and reveals distinct patterns of co-occurrence. For instance, the \textit{Body} and \textit{Clothing} concepts are frequently annotated together, as are \textit{Type of Shot} and \textit{Look}. Fig. \ref{fig:subfig_pca} shows the contribution of each concept to the first two principal components of the principal component analysis (PCA). This analysis reveals which concept associations account for the greatest variance in the data. Notably, the \textit{Body}-\textit{Clothing} pair contributes most significantly to the first principal component, while \textit{Activities} and \textit{Posture} contributes significantly to the second component.

\subsubsection{Analysis of the connection between objectification levels and concepts}

\begin{figure}[!h]
\centering
\begin{minipage}[t]{0.48\textwidth}
    \centering
    \includegraphics[width=\linewidth]{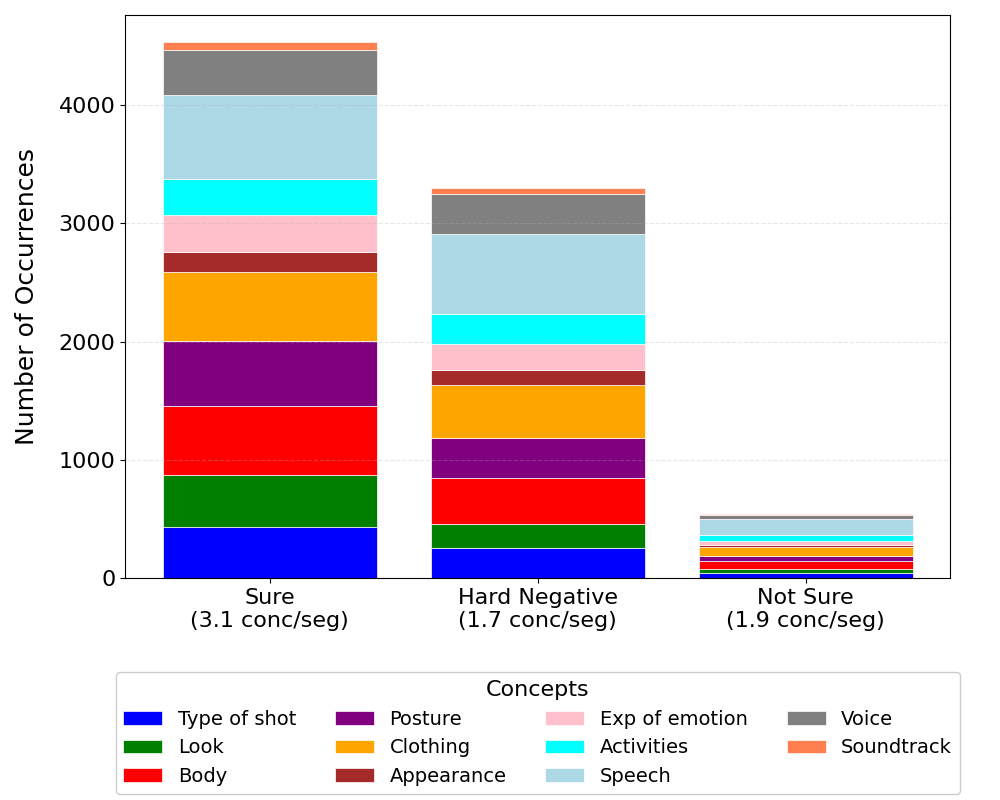}
    \caption{Number of concept occurrences for each objectification level}
    \label{fig:distr_concepts}
    
    \vspace{0.5cm} 
    
    \includegraphics[width=\linewidth]{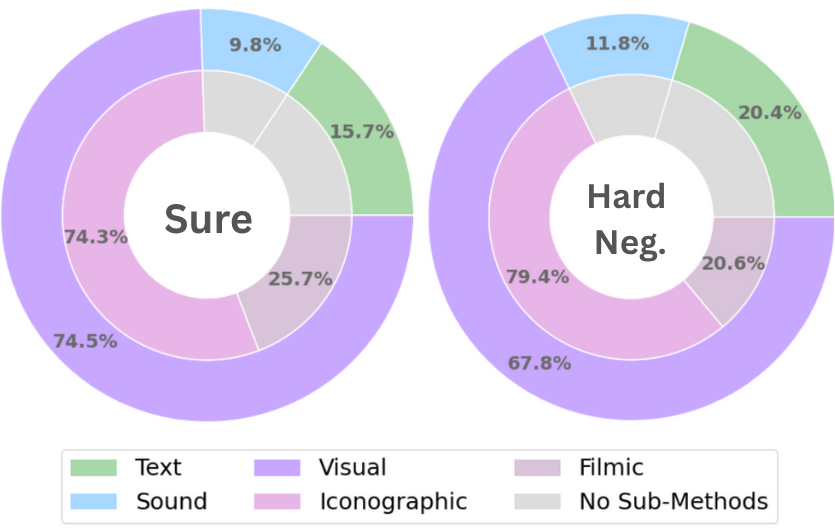}
    \caption{Distribution of concept modalities for S (left) and HN (right) segments, neglecting NS < 5\%.}
    \label{fig:pie_charts_concepts}
\end{minipage}
\hfill
\begin{minipage}[t]{0.48\textwidth}
    \centering
    \includegraphics[width=\linewidth]{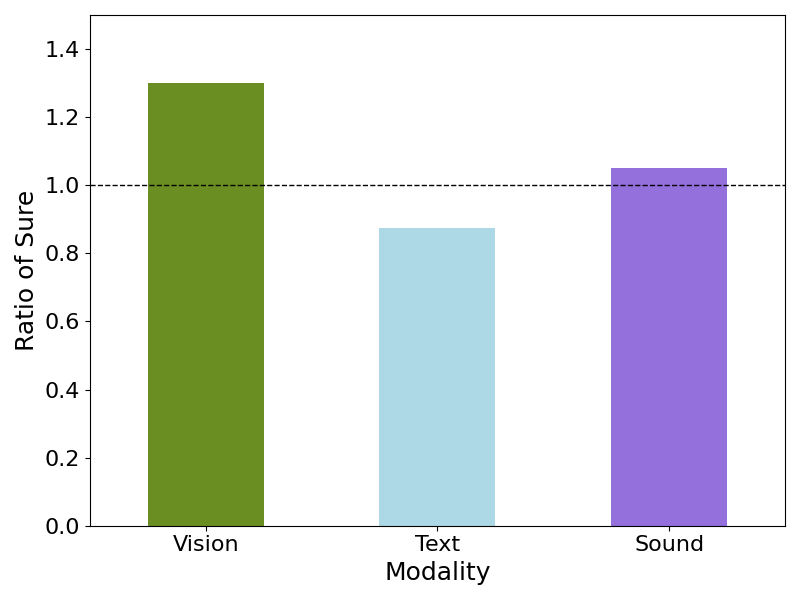} \\
    \vspace{0.3cm}
    \includegraphics[width=\linewidth]{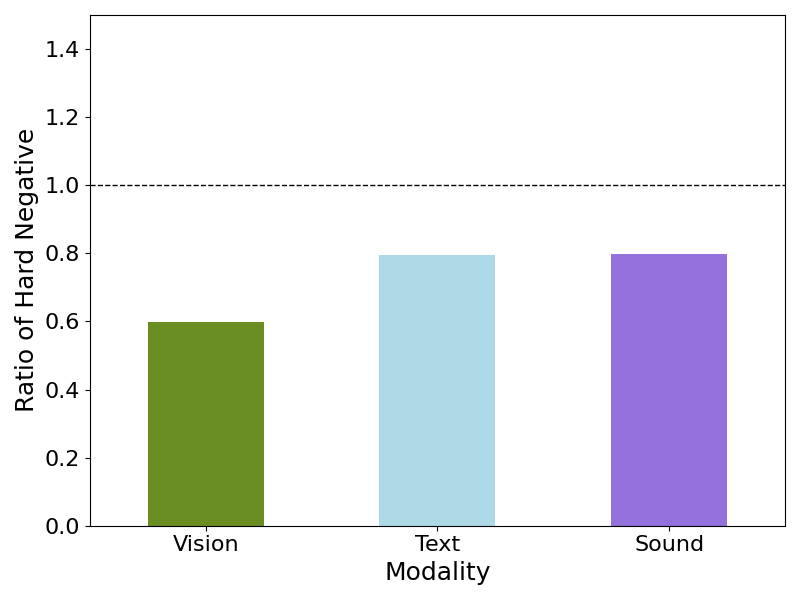}
    \caption{Ratios of concept occurrences. Top: ratio of S to HN and NS segments. Bottom: ratio of HN to S and NS segments for each modality.}
    \label{fig:barplots_concepts}
\end{minipage}
\end{figure}

The distribution of concepts in connection with the objectification levels is presented in Fig.\ref{fig:distr_concepts}, which shows the number of occurrences of each concept, disaggregated over each of the non-EN levels. We can see that the most prominent concepts are \textit{Look}, \textit{Posture}, \textit{Clothing}, \textit{Body} and \textit{Speech}. All concepts have significant representation except for Soundtrack. It is notable that the average number of concepts annotated per segment increases significantly with the level of objectification: the number of concepts for S segments (3.1) is almost twice that of HN segments (1.7). 
The fact that objectification is a multi-factorial phenomenon interestingly corroborates with results in neuro-psychology where \cite{bernard_revealing_2019} showed that clothing alone is not sufficient to produce objectification.
Fig. \ref{fig:pie_charts_concepts} depicts, for each level S and HN, the distribution of concept modalities. We see that the visual modality is most frequent, and relatively more frequent than text and sound in S compared to HN segments. Visual concepts are also
split into filmic (gathering \textit{Type of shot} and \textit{Look}) and iconographic properties (the rest of the visual concepts). Fig. \ref{fig:barplots_concepts} represents, for each concept modality, the ratio of occurrences of S (resp. HN) to HN (resp. S) and NS segments. We observe that visual concepts are significantly more specific to S than to HN segments (in the visual filmic modality, Sure segments are ca. 50\% more numerous than the other levels).

\begin{figure*}[!h]
\centering
  \includegraphics[width=1.0\textwidth]{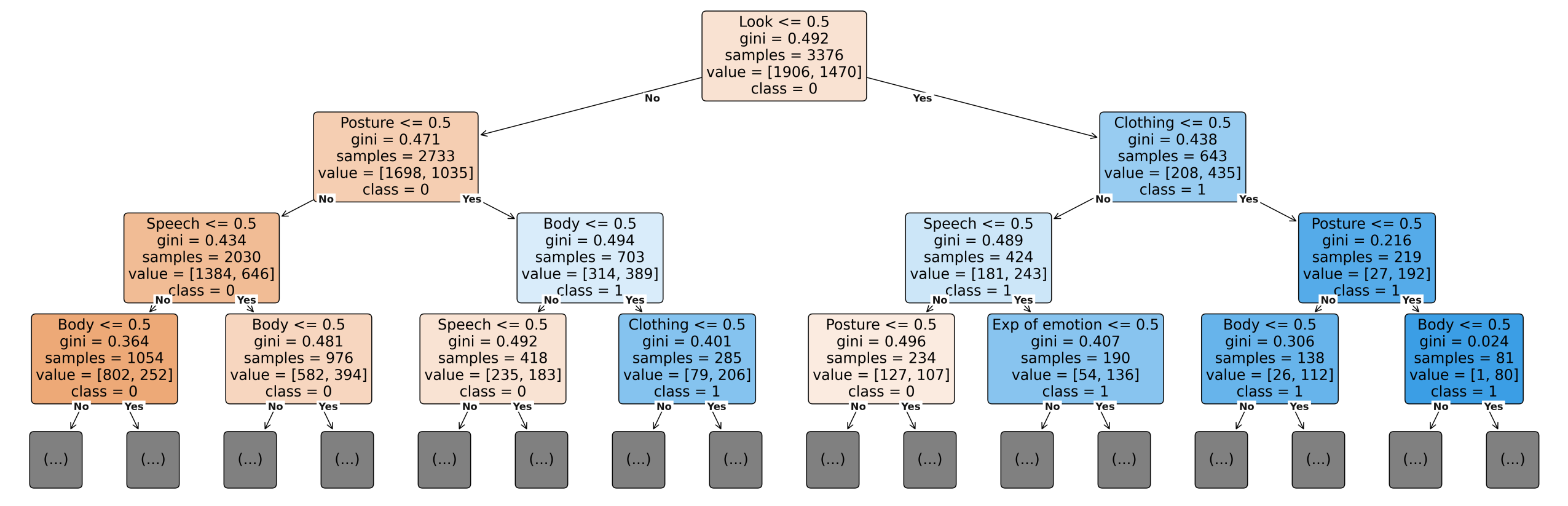}
  \caption{Decision tree to classify between HN and S segment based only on the list of concepts present for each segment. \revised{Produced with scikit-learn. Nodes report the splitting rule (feature $\leq$ 0.5), Gini impurity, sample count, and class distribution. Since all features are binary indicators (1 = concept present, 0 = absent) derived from annotations, splitting at the midpoint 0.5 is equivalent to splitting on presence/absence (not a hyper-parameter). Left branches correspond to the concept being absent, right branches to the concept being present. Node fill color indicates the majority class (orange = class 0, HN; blue = class 1, S), with color intensity proportional to node purity — the proportion of samples belonging to the majority class.} Accuracy: 0.768, F1 Score: 0.684 , Weighted F1 Score: 0.759}
  \label{DT_general}
\end{figure*}

We finally examine whether the annotated concepts alone allow us to classify a video clip as showing objectification (S) or as a hard negative (HN), since easy negatives (EN) contain no concepts by definition. Fig. \ref{DT_general} depicts a decision tree to classify a clip between HN and S. It shows that concept patterns more specific to S segments are: \textit{Look}, \textit{Posture}, \textit{Look}+\textit{Posture}, \textit{Body}+\textit{Posture}, \textit{Speech}+\textit{Posture}, \textit{Speech}+\textit{Body}.

\section{New multimodal interpretive tasks for ML: feasibility, model assessment and analysis, benefits from concept annotations}
\label{sec:experiments}

The objective of this section is threefold.
After formulating the learning tasks in Sec. \ref{sec:tasks} and describing the common experimental settings in Sec. \ref{sec:common_settings}, in Sec. \ref{sec:concept_detec}, we study the feasibility of detecting objectification at the level of each concept modality.
In Sec. \ref{sec:obj_detec}, we analyze more precisely model design: (i) how hard negative samples affect them, (ii) how multimodality can be approached, (iii) what are the model biases in connection with our specific annotation process and how to design more sound models for MObyGaze.
In Sec. \ref{sec:interest}, we exemplify how to leverage the rich concept annotation of MObyGaze to improve multimodal and explainable models.

\subsection{Formulation of the tasks}\label{sec:tasks}

The MObyGaze data consists of temporal segments with annotated boundaries, levels of objectification, and associated concepts, for which we define a formal notation as follows. A set of movies $M=\{M_m\}_{m=1}^{20}$ is annotated by labellers $L=\{L_l\}_{l=1}^2$, each producing a sequence of annotations $A_{m,l}=\{a_j^{m,l}\}_{j=1}^{N_{m,l}}$ for labeller $L_l$ having delimited $N_{m,l}$ segments for movie $M_m$. An annotation is a tuple $a_j^{m,l}=(s_j^{m,l},e_j^{m,l},l_j^{m,l},\mathbf{c}_j^{m,l})$ corresponding to start frame $s_j^{m,l}$, end frame $e_j^{m,l}$, annotated level of objectification $l_j^{m,l}$ and list of concepts $\mathbf{c}_j^{m,l}$.

We can hence consider several tasks: (i) detect the presence of objectifying concepts in given video segments, (ii) detect for each video segment the presence (or even level S or HN) of objectification corresponding to the task label, or (iii) localize segments with objectification.
In the rest of the article, we first consider task (i) of detecting the presence of specific objectifying concepts (Sec. \ref{sec:visual_concept}), then task (ii) of detecting the presence of objectification (Sec. \ref{sec:obj_detec}). We therefore use the term ``clip'' instead of ``video segment'' to emphasize that we do not make use of the segmenting information in what follows, instead considering each video segment as an individual video clip (except for the split definition discussed below). We show the feasibility of the localization task (iii) in App. \ref{sec:localization}. 


\subsection{Common experimental settings}\label{sec:common_settings}
We provide below the main aspects of our experimental protocol. All the model details and data preparation for multimodal feature extraction are provided in App. \ref{suppl:models}.

\paragraph{Train/validation/test splits}
To properly test the generalization of the models, we consider no overlap in the movies of the train, validation and test sets. To maximize the size of the training set, we proceed by leave-4-movies-out cross-validation, creating 5 folds each with 4 movies for test, 2 movies for validation and the remaining 14 movies for train (exact movie split in Table \ref{table:datasheet_films} in App. \ref{suppl:docu}). To have a cleaner dataset, we choose to discard NS samples, which correspond to cases where the labeller is uncertain and only represent 4.8\% of the samples (see Fig. \ref{fig:distr_obj}).
To account for the class imbalance, data balancing is performed in the training set by oversampling the minority class, and we report PR AUC from the perspective of the positive class, of the negative class and the macro average. 

\paragraph{Baselines}
To assess the feasibility of a task by a ML model, we compare with a trivial random baseline where the positive class is predicted with probability 0.5.
We adopt the recommended practices in AI experimental design and analysis (\cite{neurips_tuto}) and following \cite{Loftus1994}, we report the ``adjusted confidence intervals'' for the metric scores when several models are compared. We consider 5 seeds, additionally to the 5 folds, and the adjustment is meant to remove the (fold, seed) variability and enables a more reliable visualization of model effect. Multiple models can then be compared by considering that non-overlapping confidence intervals over a certain metric indicate a statistically significant difference between the models' performance. When a single model is considered, we report the standard (Student) confidence interval. We give more details on our approach of hypothesis testing for model comparison in App. \ref{app:exp_settings}.

\paragraph{Annotator variety}
For interpretive tasks, we often cannot assume the existence of a single gold label (absolute single truth being accessible or to be approximated from the label statistics) and \cite{bucarelli_leveraging_nodate, uma_learning_2021} have studied how to best consider label diversity (i.e., different annotators assigning different labels for the same item) in both model training and evaluation.
When the number of labels per item is insufficient, or the level of noise/disagreement is high, \cite{wei_aggregate_2023} show that label separation is preferable to label aggregation for training. 
\revised{For this reason, and to avoid any confusion due to data aggregation in the analyses, we generally consider only one annotator after having verified model performance consistency across annotators. Specifically, we verify in Tables \ref{table:res_intra-inter} and \ref{table:res_intra-inter_differential} that:}\\
$\bullet$ we obtain similar results over both annotators when we aim to generalize across movies only and have the same annotator in train and test;\\
$\bullet$ the models can generalize across different annotators in train and test.

\subsection{Concept detection}\label{sec:concept_detec}
Over each of the three modalities taken independently, we assess the feasibility of detecting each concept defined in the thesaurus. \revised{The model choices presented below build on the extensive literature in multimodal tasks, e.g. \cite{das_hatemm_2023, koushik_towards_2025}.}

\subsubsection{Vision modality}\label{sec:visual_concept}
We first investigate whether objectifying concepts can be detected from the MObyGaze dataset (task (i) in Sec. \ref{sec:tasks}).
For the vision modality, we consider the following models.\\
\noindent\textbf{X-CLIP+MLP}: We adapt X-CLIP (\cite{Ni_X-CLIP}), an extension of CLIP for videos (\cite{CLIP}). We keep the pre-trained model frozen and extract a 512-dimensional feature vector on every window of 16 frames, with a stride of 16, for each input video segment. The obtained vectors are mean-pooled, and the resulting vector fed to an MLP with 2 hidden layers with a final softmax layer of classification. This model is trained using fully-supervised learning with a cross-entropy loss.\\
\noindent\textbf{X-CLIP+Transf}: The X-CLIP feature vectors above are first projected to a 128-dimensional space, and appended with a CLS (classification) token to be learned. This concatenation is then passed through two self-attention layers, each followed by two linear layers with a hidden dimension of 1024 and GeLU activations. Finally, a classification head composed of two linear layers with ReLU activations is applied to the CLS token. The model is trained using cross-entropy loss.

\noindent\textbf{Task 1 (Coarse-grained/Overall concept detection)}: We first consider the binary classification task of detecting the presence of any visual concept, in which case the positive samples correspond to the clips annotated with at least one visual concept. 

Table \ref{table:res_all_visual} shows the performance of both models.
The first observation is that both models outperform the random baseline statistically significantly and with a very large effect size -- large Cohen's $d$'s around 1 for the three metrics considered (as $\hat{\sigma^{2 \prime}}$, defined in \ref{app:exp_settings}, is systematically smaller than 0.001, resulting in narrow confidence intervals and very large Cohen's $d$). 
The second observation is that the Transformer model is statistically significantly better than the MLP model, with an absolute difference in PR AUC of about 0.02.

\revised{We show in Table \ref{table:res_intra-inter} that these results hold for both annotators, and when we train on one annotator and test on the other. It is another indicator of the consistency of the MObyGaze dataset.}

These results show that it is feasible for vision models to tackle the detection of visual objectification, even if they also show there is significant room for improvement and interesting research challenges.

\begin{table*}[]
\small
\centering
\caption{Performance for the task of detecting components of visual objectification, i.e., the presence of at least one visual concept (with adjusted confidence intervals)}
\label{table:res_all_visual}
{\scriptsize
\begin{tabular}{llll}
\toprule
Model & PR AUC Positive & PR AUC Negative & PR AUC Macro \\
\cmidrule(rl){1-4}
X-CLIP+MLP & $0.703 [\pm 0.005]$ & $0.731 [\pm 0.008]$ & $0.717 [\pm 0.006]$\\
X-CLIP+Transf & $0.719 [\pm 0.005]$ & $0.758 [\pm 0.008]$ & $0.739 [\pm 0.006]$\\
random & $0.47$ & $0.53$ & $0.50$\\
\bottomrule
\end{tabular}
}
\end{table*}

\begin{table*}[]
\small
\centering
\caption{\revised{Performance for the task of detecting components of visual objectification, intra- and inter-annotator comparison, results of the X-CLIP+Transf model on PR AUC Macro (with adjusted confidence intervals)}}
\label{table:res_intra-inter}
{\scriptsize
\begin{tabular}{lll}
\toprule
Train annotator \textbackslash Test annotator & Annotator 1 & Annotator 2 \\
\cmidrule(rl){1-3}
Annotator 1 & $0.697 [\pm] 0.006]$ & $0.719 [\pm 0.005]$ \\
Annotator 2 & $0.7 [\pm 0.006]$ & $0.739 [\pm 0.005]$ \\
\bottomrule
\end{tabular}
}
\end{table*}

\noindent\textbf{Task 2 (Fine-grained/Individual concept detection)}: Second, we consider the task of detecting each individual visual concept (as many tasks as concepts), in which case the positive samples correspond to the clips annotated with this visual concept. 

Table \ref{table:res_pervisualconcept} shows the performance of detecting each individual visual concept. We observe that several concepts can be detected with non-trivial performance, but only 3 visual concepts out of 8 with a large gain in Macro PR AUC (>0.65) of X-CLIP+Transf over the random baseline (for which Macro PR AUC is 0.5). These concepts are Body, Clothing and Posture and, interestingly, are the most common in the dataset (see Fig. \ref{fig:distr_concepts}). Other well-represented concepts are Activity, Expr. of emotion and Type of shot : they are detected with non-trivial performance (statistical significance), but with a small difference compared to the random baseline. We hypothesize that such concepts are more subtle (see the Thesaurus in Fig. \ref{fig_thes}), and relevant information may not be (sufficiently) present in the X-CLIP features for Transformer layers to be trained to properly detect them with the low amount of positives. It also shows the limits of positive oversampling to counter the strong class imbalance. Lastly, Appearance and Look are poorly detected, and scarcely represented in the dataset. 

\begin{table*}[]
\small
\centering
\caption{Performance of model X-CLIP+Transf for the task of detecting individual visual concepts (with standard confidence intervals)}
\label{table:res_pervisualconcept}
{\scriptsize
\begin{tabular}{lllllllll}
\toprule
 & Activity & Appearance & Body & Clothing \\
\cmidrule(rl){1-5}
PR AUC Positive & $0.293 [\pm0.107]$ & $0.078 [\pm0.08]$ & $0.516 [\pm0.113]$ & $0.375 [\pm0.101]$ \\
PR AUC Negative & $0.936 [\pm0.053]$ & $0.956 [\pm0.061]$ & $0.912 [\pm0.084]$ & $0.926 [\pm0.056]$ \\
PR AUC Macro & $0.614 [\pm0.08]$ & $0.517 [\pm0.054]$ & $0.714 [\pm0.07]$ & $0.65 [\pm0.071]$ \\
\cmidrule(rl){1-5}
& Expr. emotion & Look & Posture & Type of shot \\
\cmidrule(rl){1-5}
PR AUC Positive & $0.222 [\pm0.088]$ & $0.143 [\pm0.096]$ & $0.391 [\pm0.11]$ & $0.27 [\pm0.114]$\\
PR AUC Negative & $0.958 [\pm0.048]$ & $0.937 [\pm0.062]$ & $0.917 [\pm0.062]$ & $0.9 [\pm0.08]$\\
PR AUC Macro & $0.59 [\pm0.059]$ & $0.54 [\pm0.062]$ & $0.654 [\pm0.071]$ & $0.585 [\pm0.079]$\\
\bottomrule
\end{tabular}
}
\end{table*}

\subsubsection{Textual modality}

For the textual modality, we consider the following models.

\noindent\textbf{BERT+Transf}: We choose to benchmark a masked language model of type bidirectional encoder. We specifically consider a frozen BERT model to extract contextualized embeddings, which are then passed into a lightweight Transformer-based classifier similar to the one presented for the vision modality. This model is referred to as BERT+Transf. We only present these results for the sake of concision, but we also considered a distilled version of RoBERTa \cite{liu2019roberta}, namely DistillRoBERTa\footnote{\url{https://huggingface.co/distilbert/distilroberta-base}}, with two fine-tuning strategies (one keeping frozen only the embedding layers and the other tuning only the 769 parameters for the classification head applied to the CLS token). Results were quantitatively similar to BERT+Transf.\\
\noindent\textbf{BERT+MLP}: Same BERT features as above are fed into the MLP similar to the one for the visual modality.

\noindent\textbf{Task (Textual Speech concept detection)}: We consider the task of detecting the textual concept (when speech is annotated as objectifying owing to what is said), in which case the positive samples correspond to the clips annotated with the speech concept. 

Table \ref{table:res_textualconcept} shows both models outperform the random baseline significantly. However, there is no significant difference between having a Transformer or an MLP processing BERT features. Again, this demonstrates the feasibility of detecting objectification in text but also shows room for improvement.

\begin{table*}[]
\small
\centering\caption{Performance of model BERT+Transf for the task of detecting the (textual) speech concept (with adjusted confidence intervals)}
\label{table:res_textualconcept}
{\scriptsize
\begin{tabular}{llll}
\toprule
Model & PR AUC Positive & PR AUC Negative & PR AUC Macro \\
\cmidrule(rl){1-4}
BERT+MLP & $0.411[\pm 0.07]$ & $0.895[\pm 0.003]$ & $0.653[\pm 0.004]$\\
BERT+Transf & $0.407[\pm 0.07]$ & $0.893[\pm 0.003]$ & $0.650[\pm 0.004]$\\
random & $0.214$ & $0.786$ & $0.500$\\
\bottomrule
\end{tabular}
}
\end{table*}

\subsubsection{Audio modality}

For the audio modality, we consider the following models.

\noindent\textbf{AST+Transf}: Audio Spectrogram Transformer (AST) \cite{ast} is used to extract audio features (see App. \ref{suppl:models} for details). These features are then fed as input into a Transformer model similar to the ones for the visual and textual tasks.\\
\noindent\textbf{AST+MLP}: Same AST features as above are fed into the MLP similar to the one for the visual modality.

\noindent\textbf{Task}: We consider the task of detecting the presence of any audio concept (voice or soundtrack), in which case the positive samples correspond to the clips annotated any of each.

The results reported in Table \ref{table:audio} show no performance improvement compared to the random baseline. 
\revised{This negative result must however be put in perspective with the choice of the feature extractor: while we wanted to have a homogeneous choice of pre-extractor architecture over the modalities, whereby the choice of AST feature, the results in Sec. \ref{sec:ext_cons} indicate that lower-level features, albeit simpler, would provide more useful information compared to AST (which has been trained on non-human sounds).}

\begin{table*}[]
\small
\centering
\caption{Performance of model AST+Transf for the task of detecting the audio concept (with adjusted confidence intervals)}
\label{table:audio}
{\scriptsize
\begin{tabular}{llll}
\toprule
Model & PR AUC Positive & PR AUC Negative & PR AUC Macro \\
\cmidrule(rl){1-4}
AST+MLP & $0.190[\pm 0.05]$ & $0.835[\pm 0.006]$ & $0.513[\pm 0.005]$\\
AST+Transf & $0.183[\pm 0.05]$ & $0.833[\pm 0.006]$ & $0.508[\pm 0.005]$\\
random & $0.174$ & $0.826$ & $0.500$\\
\bottomrule
\end{tabular}
}
\end{table*}

\subsection{Objectification classification}\label{sec:obj_detec}
We now investigate the classification of the ojectification levels from the MObyGaze dataset (task (ii) in Sec. \ref{sec:tasks}).
We first work with the objectification rating labels (EN, HN, S) to analyze the impact of HN (Sec. \ref{sec:hardneg}), then we study multimodal objectification detection (Sec. \ref{sec:mm}), and finally biases and model failures (Sec. \ref{sec:biases}) to propose improvements (Sec. \ref{sec:model_improv}).

\subsubsection{Impact of hard negative samples on objectification classification}\label{sec:hardneg}
In Sec. \ref{sec:visual_concept}, we have considered binary classification for the visual concepts where any clip annotated with a visual concept is considered positive. However, a number of such clips are labeled Hard Negative (HN), meaning that objectifying visual elements are present but are not judged sufficient by the annotator to consider a Sure level of objectification. Here we focus on the X-CLIP+Transf visual model and analyze the impact on model performance of assigning HN clips to the negative class, leaving the positive clips being S only with at least a visual concept.

Table \ref{table:vis_HN_neg} shows that the difference between the X-CLIP+Transf model and the random baseline is still statistically significant both for a binary task where HN clips are in the negative class, and a 3-class problem where HN makes up a separate class. Comparing the results with the case when HN are in the positive class (since we consider the presence of any visual concept as a positive in Table \ref{table:res_all_visual}), we however observe that HN are strong confusers, which shows the importance of fine-grained annotation for interpretive tasks, as also shown by \cite{samory_call_nodate}.

\begin{table*}[]
\small
\centering
\caption{Performance for classification of objectification levels from the visual modality only (with adjusted confidence intervals)}
\label{table:vis_HN_neg}
{\scriptsize
\begin{tabular}{lllll}
\toprule
Task & Model & PR AUC Positive & PR AUC Negative & PR AUC Macro \\
\cmidrule(rl){1-5}
EN$\cup$HN vs S & X-CLIP+Transf & $0.337[\pm 0.008]$ & $0.890[\pm 0.006]$ & $0.613[\pm 0.006]$\\
& random & $0.198[\pm 0.008]$ & $0.802[\pm 0.006]$ & $0.500[\pm 0.006]$\\
\cmidrule(rl){1-5}
EN vs HN vs S & X-CLIP+Transf & - & - & $0.489[\pm 0.003]$\\
& random & - & - & $0.333[\pm 0.003]$\\
\bottomrule
\end{tabular}
}
\end{table*}

\subsubsection{Multimodal objectification detection}\label{sec:mm}
We now consider whether combining information from multiple modalities could benefit objectification detection.

\noindent\textbf{Models}: We consider the single-modality models for vision (named V) and text (named T) introduced in Sec. \ref{sec:concept_detec}. We consider a multimodal model named VT where, for each clip, the visual (X-CLIP) and textual (BERT) token vectors are projected to a common feature dimension, concatenated along the token axis and fed to a Transformer model similar to that also described in Sec. \ref{sec:concept_detec} for the individual modalities. \revised{We consider feature concatenation following \cite{koushik_towards_2025} who found that attention-based fusion never obtains better performance than concatenation for multimodal hate detection.} More details are provided in App. \ref{app:multimodalmodel}.

\noindent\textbf{Task (Multimodal Objectification detection EN vs HN$\cup$S)}: The task is to detect HN or S samples from both the visual and textual modality. All three models, which differ in their inputs, are trained and tested on the dataset (still leaving movies out between folds) where the positive class is made of the samples with at least a visual or textual concept present.

Table \ref{table:multimodal} shows the performance of the models. We observe that model VT outperforms model T and obtains results similar to model V. We also observe with VT that such an early fusion of modalities does not bring a quantitative advantage over having the visual modality only. This is a known challenge in multimedia fusion that properly weighing the importance of each modality for each sample is not trivial to learn (\cite{zheng_multi-level_2023,du_ldw_2025}). We indeed show in Sec. \ref{sec:ICIPElisa} how per-modality training based on the additional concept information can be leveraged to improve multimodal models.

\begin{table*}[]
\small
\centering
\caption{Performance of multimodal models on the task EN vs HN$\cup$S (with adjusted confidence intervals)}
\label{table:multimodal}
{\scriptsize
\begin{tabular}{llll}
\toprule
Model & PR AUC Positive & PR AUC Negative & PR AUC Macro \\
\cmidrule(rl){1-4}
V & $0.781[\pm 0.006]$ & $0.701[\pm 0.009]$ & $0.741[\pm 0.006]$\\
T & $0.696 [\pm 0.006]$ & $0.636 [\pm 0.009]$ & $0.666 [\pm 0.006]$\\
VT & $0.769[\pm 0.006]$ & $0.711[\pm 0.009]$ & $0.740[\pm 0.006]$\\
\bottomrule
\end{tabular}
}
\end{table*}

\subsubsection{Analysis of annotator and model biases}\label{sec:biases}

We now further analyze the performance of the vision model with disaggregated quantitative and qualitative analyses. We then propose model improvements in Sec. \ref{sec:model_improv}.

\paragraph{Analysis of annotator bias}

\begin{figure}
    \centering
    \begin{minipage}{0.40\textwidth}
        \centering
\begin{tikzpicture}[scale=0.6]

\definecolor{darkgray176}{RGB}{176,176,176}
\definecolor{darkslategray66}{RGB}{66,66,66}
\definecolor{lightgray204}{RGB}{204,204,204}
\definecolor{peru22412844}{RGB}{224,128,44}
\definecolor{steelblue49115161}{RGB}{49,115,161}

\begin{axis}[
legend cell align={left},
legend style={
  fill opacity=0.8,
  draw opacity=1,
  text opacity=1,
  at={(0.03,0.97)},
  anchor=north west,
  draw=lightgray204
},
tick align=outside,
tick pos=left,
unbounded coords=jump,
x grid style={darkgray176},
xmin=-0.5, xmax=9.5,
xtick style={color=black},
xtick={0,1,2,3,4,5,6,7,8,9},
xticklabel style={rotate=90.0, font=\footnotesize},
xticklabels={
  Crash,
  Forrest Gump,
  Gone Girl,
  Indiana Jones,
  Pulp Fiction,
  Silver Linings P.,
  Sleepless in Seattle,
  The Social Network,
  The Ugly Truth,
  Up in the Air
},
y grid style={darkgray176},
ylabel={accuracy},
ymin=0, ymax=1,
ytick style={color=black}
]
\draw[draw=none,fill=steelblue49115161] (axis cs:-0.4,0) rectangle (axis cs:0,0.837);
\draw[draw=none,fill=steelblue49115161] (axis cs:0.6,0) rectangle (axis cs:1,0.871);
\draw[draw=none,fill=steelblue49115161] (axis cs:1.6,0) rectangle (axis cs:2,0.772);
\draw[draw=none,fill=steelblue49115161] (axis cs:2.6,0) rectangle (axis cs:3,0.806);
\draw[draw=none,fill=steelblue49115161] (axis cs:3.6,0) rectangle (axis cs:4,0.465);
\draw[draw=none,fill=steelblue49115161] (axis cs:4.6,0) rectangle (axis cs:5,0.561);
\draw[draw=none,fill=steelblue49115161] (axis cs:5.6,0) rectangle (axis cs:6,0.717);
\draw[draw=none,fill=steelblue49115161] (axis cs:6.6,0) rectangle (axis cs:7,0.655);
\draw[draw=none,fill=steelblue49115161] (axis cs:7.6,0) rectangle (axis cs:8,0.42);
\draw[draw=none,fill=steelblue49115161] (axis cs:8.6,0) rectangle (axis cs:9,0.894);
\draw[draw=none,fill=peru22412844] (axis cs:5.55111512312578e-17,0) rectangle (axis cs:0.4,0.545);
\draw[draw=none,fill=peru22412844] (axis cs:1,0) rectangle (axis cs:1.4,0.407);
\draw[draw=none,fill=peru22412844] (axis cs:2,0) rectangle (axis cs:2.4,0.696);
\draw[draw=none,fill=peru22412844] (axis cs:3,0) rectangle (axis cs:3.4,0.781);
\draw[draw=none,fill=peru22412844] (axis cs:4,0) rectangle (axis cs:4.4,0.846);
\draw[draw=none,fill=peru22412844] (axis cs:5,0) rectangle (axis cs:5.4,0.758);
\draw[draw=none,fill=peru22412844] (axis cs:6,0) rectangle (axis cs:6.4,0.535);
\draw[draw=none,fill=peru22412844] (axis cs:7,0) rectangle (axis cs:7.4,0.691);
\draw[draw=none,fill=peru22412844] (axis cs:8,0) rectangle (axis cs:8.4,0.856);
\draw[draw=none,fill=peru22412844] (axis cs:9,0) rectangle (axis cs:9.4,0.517);
\draw[draw=none,fill=steelblue49115161] (axis cs:0,0) rectangle (axis cs:0,0);
\addlegendimage{only marks, mark=square*,fill=steelblue49115161}
\addlegendentry{Neg.}

\draw[draw=none,fill=peru22412844] (axis cs:0,0) rectangle (axis cs:0,0);
\addlegendimage{only marks, mark=square*,fill=peru22412844}
\addlegendentry{Pos.}

\addplot [line width=0.9pt, darkslategray66, forget plot]
table {%
-0.2 nan
-0.2 nan
};
\addplot [line width=0.9pt, darkslategray66, forget plot]
table {%
0.8 nan
0.8 nan
};
\addplot [line width=0.9pt, darkslategray66, forget plot]
table {%
1.8 nan
1.8 nan
};
\addplot [line width=0.9pt, darkslategray66, forget plot]
table {%
2.8 nan
2.8 nan
};
\addplot [line width=0.9pt, darkslategray66, forget plot]
table {%
3.8 nan
3.8 nan
};
\addplot [line width=0.9pt, darkslategray66, forget plot]
table {%
4.8 nan
4.8 nan
};
\addplot [line width=0.9pt, darkslategray66, forget plot]
table {%
5.8 nan
5.8 nan
};
\addplot [line width=0.9pt, darkslategray66, forget plot]
table {%
6.8 nan
6.8 nan
};
\addplot [line width=0.9pt, darkslategray66, forget plot]
table {%
7.8 nan
7.8 nan
};
\addplot [line width=0.9pt, darkslategray66, forget plot]
table {%
8.8 nan
8.8 nan
};
\addplot [line width=0.9pt, darkslategray66, forget plot]
table {%
0.2 nan
0.2 nan
};
\addplot [line width=0.9pt, darkslategray66, forget plot]
table {%
1.2 nan
1.2 nan
};
\addplot [line width=0.9pt, darkslategray66, forget plot]
table {%
2.2 nan
2.2 nan
};
\addplot [line width=0.9pt, darkslategray66, forget plot]
table {%
3.2 nan
3.2 nan
};
\addplot [line width=0.9pt, darkslategray66, forget plot]
table {%
4.2 nan
4.2 nan
};
\addplot [line width=0.9pt, darkslategray66, forget plot]
table {%
5.2 nan
5.2 nan
};
\addplot [line width=0.9pt, darkslategray66, forget plot]
table {%
6.2 nan
6.2 nan
};
\addplot [line width=0.9pt, darkslategray66, forget plot]
table {%
7.2 nan
7.2 nan
};
\addplot [line width=0.9pt, darkslategray66, forget plot]
table {%
8.2 nan
8.2 nan
};
\addplot [line width=0.9pt, darkslategray66, forget plot]
table {%
9.2 nan
9.2 nan
};
\end{axis}

\end{tikzpicture}
        \vspace*{-1cm}
        \subcaption{ }
        \label{fig:vit_barplot_accuracy_val}
    \end{minipage}
    \hfill
    \begin{minipage}{0.55\textwidth}
        \centering
        \includegraphics[width=1.\linewidth]{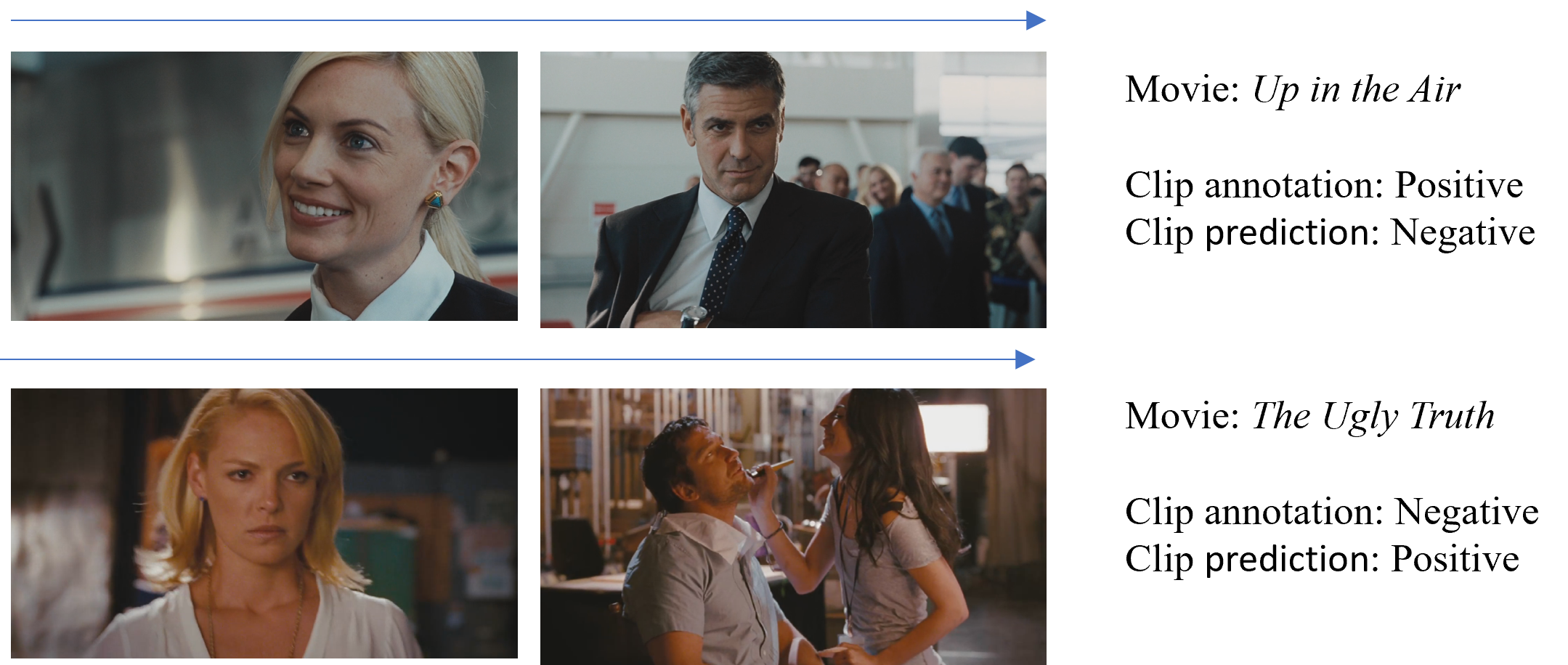}
        \vspace*{1.2cm}
        \subcaption{ }
        \label{fig:regul_UITA-UT}
    \end{minipage}
    \caption{(a) Accuracy of the X-CLIP+Transf model, disaggregated by movie and by true label, on the movies in the validation sets (note that the accuracy on positive corresponds to recall). (b): Illustration of the dependence of annotations on the movie context: clips from different movies annotated differently while reverse labels may have been assigned if the clips were considered outside their respective movies.}
    \label{fig:acc_val-regul}
\end{figure}

Fig. \ref{fig:vit_barplot_accuracy_val} shows the performance of model X-CLIP+Transf on task EN vs HN$\cup$S for visual concepts only, disaggregated over the 10 movies used in the validation sets of all the folds. Colored bars further depict the accuracy disaggregated over the positive and negative classes. 
We observe that certain movies have highly heterogeneous results, such as \textit{Crash}, \textit{Forrest Gump} and \textit{Up in the air} for which positive clips are more often misclassified as negative, while for others like \textit{The Ugly Truth}, negative clips are frequently misclassified as positive. Let us note that \textit{The Ugly Truth} is highly stereotypical with the highest fraction of positive clips (75\%, see Table \ref{tab:positive-proportion} in App. \ref{sec:suppl_dataset}).

Fig. \ref{fig:regul_UITA-UT} shows qualitative examples of such misclassifications in these movies. Two clips, which, taken independently may seem similar, but have been annotated (by the same annotator) differently (positive for \textit{Up in the Air} and negative for \textit{The Ugly Truth}). From there, we can suppose that the annotators, who have annotated an entire movie at once while watching it, make contextual and contrastive annotations: they tend to self-regulate by trying to balance the number of positive and negative labels they assign within a movie, limiting the number of positives in stereotypical movies in comparison of strongly objectifying scenes in the same movie, while annotating as positive in the less stereotypical movies scenes that are less obviously positive when taken independently but are labelled positive in contrast with other scenes within that movie. 
\revised{This finding of movie-dependent annotation aligns well with important knowledge established in psychophysics according to which humans naturally assess quality through comparison rather than in absolute terms, and this has been a central consideration in several domains based on human assessments, specifically driving the methodologies in image/video quality assessment (IQA/VQA).}
This supposition is ground for our hypothesis on how to improve model in Sec. \ref{sec:model_improv}.

\paragraph{Analysis of model bias}

\begin{figure}
    \centering
    \includegraphics[width=0.9\linewidth]{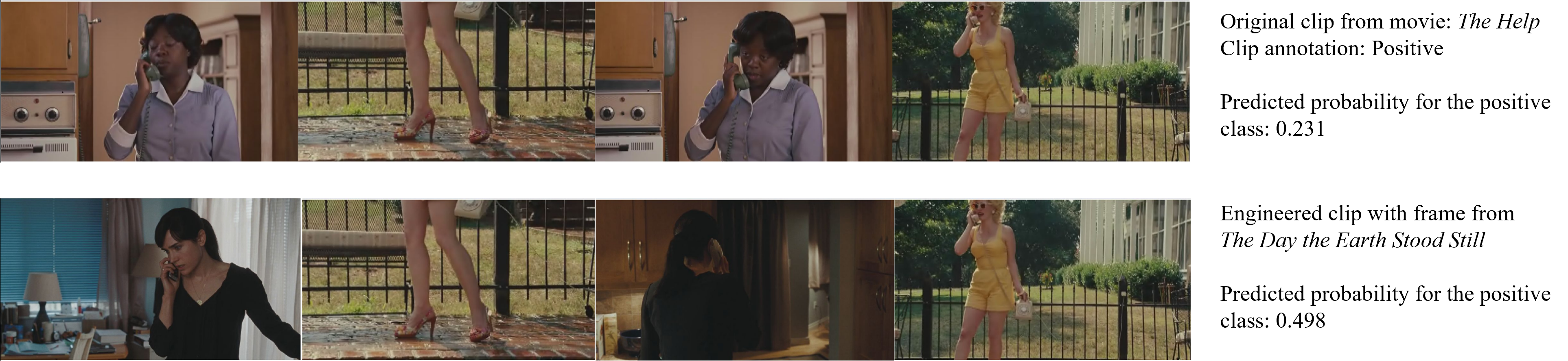}
    \caption{Illustration of context bias learned by the model. Top: The original clip is annotated positive but the predicted probability for the positive class is 0.231. Bottom: Replacing frames where visual objectification does not occur (that depict the character of color) with semantically equivalent frames from a train movie (movie with the least fraction of positive clips) significantly increases the probability of the positive class.}
    \label{fig:bias_TH-TDTEST}
\end{figure}

We make a second observation of how the movie context may bias the model performance. In Fig. \ref{fig:bias_TH-TDTEST}, a clip from the movie \textit{The Help} is annotated as positive, but the predicted probability of the positive class is only $0.231$, while being typically positive and not a corner case according to the thesaurus. To check whether the model may be confused by elements of context not seen in train, we engineer (bottom part of the figure) a new clip preserving the objectifying frames and replacing the non-objectifying frames with semantically equivalent frames from a train movie (movie with the least fraction of positive clips). This operation increases the probability of the positive class from $0.231$ to $0.498$. We consider this another indication that differences in context, aside from the very type of objectification, between train and validation/test can (expectedly) confuse the model.

\subsubsection{Improving objectification detection with differential representation}\label{sec:model_improv}
Based on the observation of biases above, we propose the following model improvement.

\noindent\textbf{Hypothesis}: We hypothesize that a model that can contrast a clip feature against the entire movie to then make the class prediction will be able to better generalize over movies.

\noindent\textbf{Model}: To test this hypothesis, we design a new model that we call \textbf{Differential}, defined as follows.
Let $f$ be the representation component of the X-CLIP+Transf model, returning the CLS-token output that is fed to the MLP classifier. For a clip $x$, $f(x) \in \mathbb{R}^{128}$ thus denotes the task features extracted by the trained Transformer from input $x$.
For a given movie $M$, we define the movie context as:
\[
c_M = \mbox{mean}_{\{x \in M\}} f(x)
\]
The differential paradigm simply consists in introducing an additional operation before the classification head, outputting a \textit{differential representation} 
\[
f_{\mbox{diff}}(x, c_M) := f(x) - c_M
\]
that is fed to the MLP instead of $f(x)$.
%
%
During the training stage, $c_M$ is replaced by its batch counterpart:
\[
c_M(\mbox{batch}) = \mbox{mean}_{\{x' \in M \cap \mbox{batch} \}} f(x')
\]
In practice, in order to ensure that $c_M(\mbox{batch})$ is close enough to $c_M$, we construct the batches so that they contain 32 clips per movie: with 2 different movies per batch, this represents a total batch size of 64.

\noindent\textbf{Results}: Table \ref{table:res_all_visual_differential} shows the test results in terms of classification performance (PR AUC) but also in terms of heterogeneity in performance between movies (right-most column), defined as the average over the folds of the standard deviation of the accuracy over all four movies of a fold test set. The larger the heterogeneity, the more differently the model performs from a movie to another; thus, a model with lower heterogeneity is likely to better generalize to new movies.
We observe that the differential paradigm significantly improves the performance over the standard X-CLIP+Transformer model in terms of PR AUC, but also significantly decreases the movie performance heterogeneity, hence verifying the hypothesis we have emitted from our previous observations on the validation sets.
\revised{Table \ref{table:res_intra-inter_differential} shows that this holds for both annotators (we omit confidence intervals for the sake of clarity).
Under this new paradigm, we verify that models maintain cross-annotator generalization. Table \ref{table:res_intra-inter_differential} shows that this is the case: a model trained to suppress the movie bias for one annotator relieves the movie bias for another annotator in test: the differential model maintains or improves the average classification performance in every case, and decreases movie performance heterogeneity in 3 cases in 4.}
Fig. \ref{fig:barplots_heterpermovie_test} further shows the performance evolution between both models, over the different movies and classes, clearly showing the improved heterogeneity of the differential model without sacrificing classification accuracy.
The better performance of the differential model supports the assumption underlying it, according to which human annotations are contextual.

\begin{table*}[!h]
\small
\centering
\caption{Performance for visual objectification detection (adjusted confidence intervals)}
\label{table:res_all_visual_differential}
{\scriptsize
\begin{tabular}{lllll}
\toprule
Model & PR AUC Pos. ($\uparrow)$ & PR AUC Neg. ($\uparrow)$ & PR AUC Macro ($\uparrow)$ & Heterogeneity ($\downarrow)$ \\
\cmidrule(rl){1-5}
X-CLIP+Transf & $0.719 [\pm 0.005]$ & $0.758 [\pm 0.007]$ & $0.739 [\pm 0.003]$ & $0.073 [\pm 0.005]$\\
Differential X-CLIP+Transf & $0.730 [\pm 0.005]$ & $0.784 [\pm 0.007]$ & $0.757 [\pm 0.005]$ & $0.061 [\pm 0.005]$\\
\bottomrule
\end{tabular}
}
\end{table*}

\begin{table*}[!h]
\small
\centering
\caption{\revised{Performance for visual objectification detection, model comparison with intra- and inter-annotator comparison, results as PR AUC Macro ($\uparrow)$| Heterogeneity ($\downarrow)$}}
\label{table:res_intra-inter_differential}
{\scriptsize
\begin{tabular}{lcccc}
\toprule
Train annotator \textbackslash Test annotator & \multicolumn{2}{c}{Annotator 1} & \multicolumn{2}{c}{Annotator 2} \\
\cmidrule(rl){1-5}
\multirow{2}{*}{Annotator 1} & \centering Non-diff & Diff & Non-diff & Diff \\
  & 0.697 | 0.079 & \darkgreen{0.704 | 0.068} & 0.719 | 0.063 & \darkgreen{0.721 | 0.055} \\
\multirow{2}{*}{Annotator 2} & Non-diff & Diff & Non-diff & Diff \\
  & 0.7 | 0.057 & \darkgreen{0.711 | 0.064} & 0.739 | 0.073 & \darkgreen{0.757 | 0.061} \\
\bottomrule
\end{tabular}
}
\end{table*}

\begin{figure}[!h]
    \centering
    \begin{minipage}[t]{0.45\textwidth}
        \centering
\begin{tikzpicture}[scale=0.8]

\definecolor{darkgray176}{RGB}{176,176,176}
\definecolor{darkslategray66}{RGB}{66,66,66}
\definecolor{lightgray204}{RGB}{204,204,204}
\definecolor{peru22412844}{RGB}{224,128,44}
\definecolor{steelblue49115161}{RGB}{49,115,161}

\begin{axis}[
legend cell align={left},
legend style={
  fill opacity=0.8,
  draw opacity=1,
  text opacity=1,
  at={(0.03,0.97)},
  anchor=north west,
  draw=lightgray204
},
tick align=outside,
tick pos=left,
unbounded coords=jump,
x grid style={darkgray176},
xmin=-0.5, xmax=19.5,
xtick style={color=black},
xtick={0,1,2,3,4,5,6,7,8,9,10,11,12,13,14,15,16,17,18,19},
xticklabel style={rotate=90.0, font=\tiny},
xticklabels={
  As Good as It Gets,
  Crash,
  {Crazy,Stupid,Love},
  Flight,
  Forrest Gump,
  Friends with Benefits,
  Gone Girl,
  Indiana Jones,
  Juno,
  Marley \& Me,
  Pulp Fiction,
  Silver Linings Playbook,
  Sleepless in Seattle,
  The Day the Earth Stood Still,
  The Girl with the Dragon Tattoo,
  The Help,
  The Social Network,
  The Ugly Truth,
  Titanic,
  Up in the Air
},
y grid style={darkgray176},
ylabel={accuracy},
ymin=0, ymax=1,
ytick style={color=black}
]
\draw[draw=none,fill=steelblue49115161] (axis cs:-0.4,0) rectangle (axis cs:0,0.571);
\draw[draw=none,fill=steelblue49115161] (axis cs:0.6,0) rectangle (axis cs:1,0.765);
\draw[draw=none,fill=steelblue49115161] (axis cs:1.6,0) rectangle (axis cs:2,0.532);
\draw[draw=none,fill=steelblue49115161] (axis cs:2.6,0) rectangle (axis cs:3,0.871);
\draw[draw=none,fill=steelblue49115161] (axis cs:3.6,0) rectangle (axis cs:4,0.774);
\draw[draw=none,fill=steelblue49115161] (axis cs:4.6,0) rectangle (axis cs:5,0.376);
\draw[draw=none,fill=steelblue49115161] (axis cs:5.6,0) rectangle (axis cs:6,0.641);
\draw[draw=none,fill=steelblue49115161] (axis cs:6.6,0) rectangle (axis cs:7,0.806);
\draw[draw=none,fill=steelblue49115161] (axis cs:7.6,0) rectangle (axis cs:8,0.846);
\draw[draw=none,fill=steelblue49115161] (axis cs:8.6,0) rectangle (axis cs:9,0.787);
\draw[draw=none,fill=steelblue49115161] (axis cs:9.6,0) rectangle (axis cs:10,0.632);
\draw[draw=none,fill=steelblue49115161] (axis cs:10.6,0) rectangle (axis cs:11,0.683);
\draw[draw=none,fill=steelblue49115161] (axis cs:11.6,0) rectangle (axis cs:12,0.804);
\draw[draw=none,fill=steelblue49115161] (axis cs:12.6,0) rectangle (axis cs:13,0.885);
\draw[draw=none,fill=steelblue49115161] (axis cs:13.6,0) rectangle (axis cs:14,0.887);
\draw[draw=none,fill=steelblue49115161] (axis cs:14.6,0) rectangle (axis cs:15,0.959);
\draw[draw=none,fill=steelblue49115161] (axis cs:15.6,0) rectangle (axis cs:16,0.611);
\draw[draw=none,fill=steelblue49115161] (axis cs:16.6,0) rectangle (axis cs:17,0.353);
\draw[draw=none,fill=steelblue49115161] (axis cs:17.6,0) rectangle (axis cs:18,0.511);
\draw[draw=none,fill=steelblue49115161] (axis cs:18.6,0) rectangle (axis cs:19,0.909);
\draw[draw=none,fill=peru22412844] (axis cs:5.55111512312578e-17,0) rectangle (axis cs:0.4,0.687);
\draw[draw=none,fill=peru22412844] (axis cs:1,0) rectangle (axis cs:1.4,0.701);
\draw[draw=none,fill=peru22412844] (axis cs:2,0) rectangle (axis cs:2.4,0.791);
\draw[draw=none,fill=peru22412844] (axis cs:3,0) rectangle (axis cs:3.4,0.718);
\draw[draw=none,fill=peru22412844] (axis cs:4,0) rectangle (axis cs:4.4,0.64);
\draw[draw=none,fill=peru22412844] (axis cs:5,0) rectangle (axis cs:5.4,0.956);
\draw[draw=none,fill=peru22412844] (axis cs:6,0) rectangle (axis cs:6.4,0.93);
\draw[draw=none,fill=peru22412844] (axis cs:7,0) rectangle (axis cs:7.4,0.671);
\draw[draw=none,fill=peru22412844] (axis cs:8,0) rectangle (axis cs:8.4,0.714);
\draw[draw=none,fill=peru22412844] (axis cs:9,0) rectangle (axis cs:9.4,0.742);
\draw[draw=none,fill=peru22412844] (axis cs:10,0) rectangle (axis cs:10.4,0.818);
\draw[draw=none,fill=peru22412844] (axis cs:11,0) rectangle (axis cs:11.4,0.727);
\draw[draw=none,fill=peru22412844] (axis cs:12,0) rectangle (axis cs:12.4,0.372);
\draw[draw=none,fill=peru22412844] (axis cs:13,0) rectangle (axis cs:13.4,0.543);
\draw[draw=none,fill=peru22412844] (axis cs:14,0) rectangle (axis cs:14.4,0.509);
\draw[draw=none,fill=peru22412844] (axis cs:15,0) rectangle (axis cs:15.4,0.012);
\draw[draw=none,fill=peru22412844] (axis cs:16,0) rectangle (axis cs:16.4,0.67);
\draw[draw=none,fill=peru22412844] (axis cs:17,0) rectangle (axis cs:17.4,0.875);
\draw[draw=none,fill=peru22412844] (axis cs:18,0) rectangle (axis cs:18.4,0.775);
\draw[draw=none,fill=peru22412844] (axis cs:19,0) rectangle (axis cs:19.4,0.5);
\draw[draw=none,fill=steelblue49115161] (axis cs:0,0) rectangle (axis cs:0,0);
\addlegendimage{only marks, mark=square*,fill=steelblue49115161}
\addlegendentry{Neg.}

\draw[draw=none,fill=peru22412844] (axis cs:0,0) rectangle (axis cs:0,0);
\addlegendimage{only marks, mark=square*,fill=peru22412844}
\addlegendentry{Pos.}

\addplot [line width=0.9pt, darkslategray66, forget plot]
table {%
-0.2 nan
-0.2 nan
};
\addplot [line width=0.9pt, darkslategray66, forget plot]
table {%
0.8 nan
0.8 nan
};
\addplot [line width=0.9pt, darkslategray66, forget plot]
table {%
1.8 nan
1.8 nan
};
\addplot [line width=0.9pt, darkslategray66, forget plot]
table {%
2.8 nan
2.8 nan
};
\addplot [line width=0.9pt, darkslategray66, forget plot]
table {%
3.8 nan
3.8 nan
};
\addplot [line width=0.9pt, darkslategray66, forget plot]
table {%
4.8 nan
4.8 nan
};
\addplot [line width=0.9pt, darkslategray66, forget plot]
table {%
5.8 nan
5.8 nan
};
\addplot [line width=0.9pt, darkslategray66, forget plot]
table {%
6.8 nan
6.8 nan
};
\addplot [line width=0.9pt, darkslategray66, forget plot]
table {%
7.8 nan
7.8 nan
};
\addplot [line width=0.9pt, darkslategray66, forget plot]
table {%
8.8 nan
8.8 nan
};
\addplot [line width=0.9pt, darkslategray66, forget plot]
table {%
9.8 nan
9.8 nan
};
\addplot [line width=0.9pt, darkslategray66, forget plot]
table {%
10.8 nan
10.8 nan
};
\addplot [line width=0.9pt, darkslategray66, forget plot]
table {%
11.8 nan
11.8 nan
};
\addplot [line width=0.9pt, darkslategray66, forget plot]
table {%
12.8 nan
12.8 nan
};
\addplot [line width=0.9pt, darkslategray66, forget plot]
table {%
13.8 nan
13.8 nan
};
\addplot [line width=0.9pt, darkslategray66, forget plot]
table {%
14.8 nan
14.8 nan
};
\addplot [line width=0.9pt, darkslategray66, forget plot]
table {%
15.8 nan
15.8 nan
};
\addplot [line width=0.9pt, darkslategray66, forget plot]
table {%
16.8 nan
16.8 nan
};
\addplot [line width=0.9pt, darkslategray66, forget plot]
table {%
17.8 nan
17.8 nan
};
\addplot [line width=0.9pt, darkslategray66, forget plot]
table {%
18.8 nan
18.8 nan
};
\addplot [line width=0.9pt, darkslategray66, forget plot]
table {%
0.2 nan
0.2 nan
};
\addplot [line width=0.9pt, darkslategray66, forget plot]
table {%
1.2 nan
1.2 nan
};
\addplot [line width=0.9pt, darkslategray66, forget plot]
table {%
2.2 nan
2.2 nan
};
\addplot [line width=0.9pt, darkslategray66, forget plot]
table {%
3.2 nan
3.2 nan
};
\addplot [line width=0.9pt, darkslategray66, forget plot]
table {%
4.2 nan
4.2 nan
};
\addplot [line width=0.9pt, darkslategray66, forget plot]
table {%
5.2 nan
5.2 nan
};
\addplot [line width=0.9pt, darkslategray66, forget plot]
table {%
6.2 nan
6.2 nan
};
\addplot [line width=0.9pt, darkslategray66, forget plot]
table {%
7.2 nan
7.2 nan
};
\addplot [line width=0.9pt, darkslategray66, forget plot]
table {%
8.2 nan
8.2 nan
};
\addplot [line width=0.9pt, darkslategray66, forget plot]
table {%
9.2 nan
9.2 nan
};
\addplot [line width=0.9pt, darkslategray66, forget plot]
table {%
10.2 nan
10.2 nan
};
\addplot [line width=0.9pt, darkslategray66, forget plot]
table {%
11.2 nan
11.2 nan
};
\addplot [line width=0.9pt, darkslategray66, forget plot]
table {%
12.2 nan
12.2 nan
};
\addplot [line width=0.9pt, darkslategray66, forget plot]
table {%
13.2 nan
13.2 nan
};
\addplot [line width=0.9pt, darkslategray66, forget plot]
table {%
14.2 nan
14.2 nan
};
\addplot [line width=0.9pt, darkslategray66, forget plot]
table {%
15.2 nan
15.2 nan
};
\addplot [line width=0.9pt, darkslategray66, forget plot]
table {%
16.2 nan
16.2 nan
};
\addplot [line width=0.9pt, darkslategray66, forget plot]
table {%
17.2 nan
17.2 nan
};
\addplot [line width=0.9pt, darkslategray66, forget plot]
table {%
18.2 nan
18.2 nan
};
\addplot [line width=0.9pt, darkslategray66, forget plot]
table {%
19.2 nan
19.2 nan
};
\end{axis}

\end{tikzpicture}
        \vspace*{-1cm}
        \subcaption{Standard X-CLIP+Transf}
        \label{fig:barplots_heterpermovie_standard_test}
    \end{minipage}
    \hfill
    \begin{minipage}[t]{0.45\textwidth}
        \centering
\begin{tikzpicture}[scale=0.8]

\definecolor{darkgray176}{RGB}{176,176,176}
\definecolor{darkslategray66}{RGB}{66,66,66}
\definecolor{lightgray204}{RGB}{204,204,204}
\definecolor{peru22412844}{RGB}{224,128,44}
\definecolor{steelblue49115161}{RGB}{49,115,161}

\begin{axis}[
legend cell align={left},
legend style={fill opacity=0.8, draw opacity=1, text opacity=1, draw=lightgray204},
tick align=outside,
tick pos=left,
unbounded coords=jump,
x grid style={darkgray176},
xmin=-0.5, xmax=19.5,
xtick style={color=black},
xtick={0,1,2,3,4,5,6,7,8,9,10,11,12,13,14,15,16,17,18,19},
xticklabel style={rotate=90.0, font=\tiny},
xticklabels={
  As Good as It Gets,
  Crash,
  {Crazy,Stupid,Love},
  Flight,
  Forrest Gump,
  Friends with Benefits,
  Gone Girl,
  Indiana Jones,
  Juno,
  Marley \& Me,
  Pulp Fiction,
  Silver Linings Playbook,
  Sleepless in Seattle,
  The Day the Earth Stood Still,
  The Girl with the Dragon Tattoo,
  The Help,
  The Social Network,
  The Ugly Truth,
  Titanic,
  Up in the Air
},
y grid style={darkgray176},
ylabel={accuracy},
ymin=0, ymax=1,
ytick style={color=black}
]
\draw[draw=none,fill=steelblue49115161] (axis cs:-0.4,0) rectangle (axis cs:0,0.633);
\draw[draw=none,fill=steelblue49115161] (axis cs:0.6,0) rectangle (axis cs:1,0.724);
\draw[draw=none,fill=steelblue49115161] (axis cs:1.6,0) rectangle (axis cs:2,0.662);
\draw[draw=none,fill=steelblue49115161] (axis cs:2.6,0) rectangle (axis cs:3,0.806);
\draw[draw=none,fill=steelblue49115161] (axis cs:3.6,0) rectangle (axis cs:4,0.699);
\draw[draw=none,fill=steelblue49115161] (axis cs:4.6,0) rectangle (axis cs:5,0.734);
\draw[draw=none,fill=steelblue49115161] (axis cs:5.6,0) rectangle (axis cs:6,0.694);
\draw[draw=none,fill=steelblue49115161] (axis cs:6.6,0) rectangle (axis cs:7,0.791);
\draw[draw=none,fill=steelblue49115161] (axis cs:7.6,0) rectangle (axis cs:8,0.846);
\draw[draw=none,fill=steelblue49115161] (axis cs:8.6,0) rectangle (axis cs:9,0.8);
\draw[draw=none,fill=steelblue49115161] (axis cs:9.6,0) rectangle (axis cs:10,0.681);
\draw[draw=none,fill=steelblue49115161] (axis cs:10.6,0) rectangle (axis cs:11,0.805);
\draw[draw=none,fill=steelblue49115161] (axis cs:11.6,0) rectangle (axis cs:12,0.696);
\draw[draw=none,fill=steelblue49115161] (axis cs:12.6,0) rectangle (axis cs:13,0.654);
\draw[draw=none,fill=steelblue49115161] (axis cs:13.6,0) rectangle (axis cs:14,0.734);
\draw[draw=none,fill=steelblue49115161] (axis cs:14.6,0) rectangle (axis cs:15,0.619);
\draw[draw=none,fill=steelblue49115161] (axis cs:15.6,0) rectangle (axis cs:16,0.673);
\draw[draw=none,fill=steelblue49115161] (axis cs:16.6,0) rectangle (axis cs:17,0.62);
\draw[draw=none,fill=steelblue49115161] (axis cs:17.6,0) rectangle (axis cs:18,0.632);
\draw[draw=none,fill=steelblue49115161] (axis cs:18.6,0) rectangle (axis cs:19,0.742);
\draw[draw=none,fill=peru22412844] (axis cs:5.55111512312578e-17,0) rectangle (axis cs:0.4,0.627);
\draw[draw=none,fill=peru22412844] (axis cs:1,0) rectangle (axis cs:1.4,0.792);
\draw[draw=none,fill=peru22412844] (axis cs:2,0) rectangle (axis cs:2.4,0.727);
\draw[draw=none,fill=peru22412844] (axis cs:3,0) rectangle (axis cs:3.4,0.846);
\draw[draw=none,fill=peru22412844] (axis cs:4,0) rectangle (axis cs:4.4,0.814);
\draw[draw=none,fill=peru22412844] (axis cs:5,0) rectangle (axis cs:5.4,0.709);
\draw[draw=none,fill=peru22412844] (axis cs:6,0) rectangle (axis cs:6.4,0.878);
\draw[draw=none,fill=peru22412844] (axis cs:7,0) rectangle (axis cs:7.4,0.753);
\draw[draw=none,fill=peru22412844] (axis cs:8,0) rectangle (axis cs:8.4,0.762);
\draw[draw=none,fill=peru22412844] (axis cs:9,0) rectangle (axis cs:9.4,0.79);
\draw[draw=none,fill=peru22412844] (axis cs:10,0) rectangle (axis cs:10.4,0.762);
\draw[draw=none,fill=peru22412844] (axis cs:11,0) rectangle (axis cs:11.4,0.682);
\draw[draw=none,fill=peru22412844] (axis cs:12,0) rectangle (axis cs:12.4,0.535);
\draw[draw=none,fill=peru22412844] (axis cs:13,0) rectangle (axis cs:13.4,0.761);
\draw[draw=none,fill=peru22412844] (axis cs:14,0) rectangle (axis cs:14.4,0.719);
\draw[draw=none,fill=peru22412844] (axis cs:15,0) rectangle (axis cs:15.4,0.535);
\draw[draw=none,fill=peru22412844] (axis cs:16,0) rectangle (axis cs:16.4,0.638);
\draw[draw=none,fill=peru22412844] (axis cs:17,0) rectangle (axis cs:17.4,0.731);
\draw[draw=none,fill=peru22412844] (axis cs:18,0) rectangle (axis cs:18.4,0.736);
\draw[draw=none,fill=peru22412844] (axis cs:19,0) rectangle (axis cs:19.4,0.667);
\draw[draw=none,fill=steelblue49115161] (axis cs:0,0) rectangle (axis cs:0,0);
\addlegendimage{only marks, mark=square*,fill=steelblue49115161}
\addlegendentry{Neg.}

\draw[draw=none,fill=peru22412844] (axis cs:0,0) rectangle (axis cs:0,0);
\addlegendimage{only marks, mark=square*,fill=peru22412844}
\addlegendentry{Pos.}

\addplot [line width=0.9pt, darkslategray66, forget plot]
table {%
-0.2 nan
-0.2 nan
};
\addplot [line width=0.9pt, darkslategray66, forget plot]
table {%
0.8 nan
0.8 nan
};
\addplot [line width=0.9pt, darkslategray66, forget plot]
table {%
1.8 nan
1.8 nan
};
\addplot [line width=0.9pt, darkslategray66, forget plot]
table {%
2.8 nan
2.8 nan
};
\addplot [line width=0.9pt, darkslategray66, forget plot]
table {%
3.8 nan
3.8 nan
};
\addplot [line width=0.9pt, darkslategray66, forget plot]
table {%
4.8 nan
4.8 nan
};
\addplot [line width=0.9pt, darkslategray66, forget plot]
table {%
5.8 nan
5.8 nan
};
\addplot [line width=0.9pt, darkslategray66, forget plot]
table {%
6.8 nan
6.8 nan
};
\addplot [line width=0.9pt, darkslategray66, forget plot]
table {%
7.8 nan
7.8 nan
};
\addplot [line width=0.9pt, darkslategray66, forget plot]
table {%
8.8 nan
8.8 nan
};
\addplot [line width=0.9pt, darkslategray66, forget plot]
table {%
9.8 nan
9.8 nan
};
\addplot [line width=0.9pt, darkslategray66, forget plot]
table {%
10.8 nan
10.8 nan
};
\addplot [line width=0.9pt, darkslategray66, forget plot]
table {%
11.8 nan
11.8 nan
};
\addplot [line width=0.9pt, darkslategray66, forget plot]
table {%
12.8 nan
12.8 nan
};
\addplot [line width=0.9pt, darkslategray66, forget plot]
table {%
13.8 nan
13.8 nan
};
\addplot [line width=0.9pt, darkslategray66, forget plot]
table {%
14.8 nan
14.8 nan
};
\addplot [line width=0.9pt, darkslategray66, forget plot]
table {%
15.8 nan
15.8 nan
};
\addplot [line width=0.9pt, darkslategray66, forget plot]
table {%
16.8 nan
16.8 nan
};
\addplot [line width=0.9pt, darkslategray66, forget plot]
table {%
17.8 nan
17.8 nan
};
\addplot [line width=0.9pt, darkslategray66, forget plot]
table {%
18.8 nan
18.8 nan
};
\addplot [line width=0.9pt, darkslategray66, forget plot]
table {%
0.2 nan
0.2 nan
};
\addplot [line width=0.9pt, darkslategray66, forget plot]
table {%
1.2 nan
1.2 nan
};
\addplot [line width=0.9pt, darkslategray66, forget plot]
table {%
2.2 nan
2.2 nan
};
\addplot [line width=0.9pt, darkslategray66, forget plot]
table {%
3.2 nan
3.2 nan
};
\addplot [line width=0.9pt, darkslategray66, forget plot]
table {%
4.2 nan
4.2 nan
};
\addplot [line width=0.9pt, darkslategray66, forget plot]
table {%
5.2 nan
5.2 nan
};
\addplot [line width=0.9pt, darkslategray66, forget plot]
table {%
6.2 nan
6.2 nan
};
\addplot [line width=0.9pt, darkslategray66, forget plot]
table {%
7.2 nan
7.2 nan
};
\addplot [line width=0.9pt, darkslategray66, forget plot]
table {%
8.2 nan
8.2 nan
};
\addplot [line width=0.9pt, darkslategray66, forget plot]
table {%
9.2 nan
9.2 nan
};
\addplot [line width=0.9pt, darkslategray66, forget plot]
table {%
10.2 nan
10.2 nan
};
\addplot [line width=0.9pt, darkslategray66, forget plot]
table {%
11.2 nan
11.2 nan
};
\addplot [line width=0.9pt, darkslategray66, forget plot]
table {%
12.2 nan
12.2 nan
};
\addplot [line width=0.9pt, darkslategray66, forget plot]
table {%
13.2 nan
13.2 nan
};
\addplot [line width=0.9pt, darkslategray66, forget plot]
table {%
14.2 nan
14.2 nan
};
\addplot [line width=0.9pt, darkslategray66, forget plot]
table {%
15.2 nan
15.2 nan
};
\addplot [line width=0.9pt, darkslategray66, forget plot]
table {%
16.2 nan
16.2 nan
};
\addplot [line width=0.9pt, darkslategray66, forget plot]
table {%
17.2 nan
17.2 nan
};
\addplot [line width=0.9pt, darkslategray66, forget plot]
table {%
18.2 nan
18.2 nan
};
\addplot [line width=0.9pt, darkslategray66, forget plot]
table {%
19.2 nan
19.2 nan
};
\end{axis}

\end{tikzpicture}
        \vspace*{-1cm}
        \subcaption{Differential model}
        \label{fig:barplots_heterpermovie_diff_test}
    \end{minipage}
    \caption{Accuracy on the test set disaggregated by movie and by true label. (a) Original model X-CLIP+Transf. (b) Differential X-CLIP+Transf.}
    \label{fig:barplots_heterpermovie_test}
\end{figure}

\subsubsection{MLLM}\label{sec:MLLM}
\revised{We now provide preliminary results with Multimodal Large Language Models (MLLMs).
We consider the following open-source models:
\href{https://huggingface.co/OpenGVLab/InternVideo2_5_Chat_8B}{\intern{}},
\href{https://huggingface.co/OpenGVLab/InternVL3-8B}{\internvl{}},
\href{https://huggingface.co/DAMO-NLP-SG/VideoLLaMA3-7B}{\llama{}},
\href{https://huggingface.co/lmms-lab/LLaVA-Video-7B-Qwen2}{\llava{}} and
\href{https://huggingface.co/Qwen/Qwen2.5-VL-7B-Instruct}{\qwen{}}.
We provide the details on prompts, formatting and hyper-parameter choices in App. \ref{app:MLLM}.}

\paragraph{\revised{Zero-shot inference}}

\begin{table*}[!h]
\small
\centering
\caption{\revised{Zero-shot performance on the task EN vs HN$\cup$S}}
\label{table:res_EN_HNS}
{\scriptsize
\begin{tabular}{cccc}
\toprule
Model & Accuracy & Macro F1 & \makecell{proportion of\\predicted positive} \\
\cmidrule(rl){1-4}
X-CLIP+Transf & 0.685 & 0.677 & 0.51\\
\cmidrule(rl){1-4}
\internvl{} & 0.640 & 0.612 & 0.28\\
\intern{} & 0.615 & 0.554 & 0.17\\
\llava{} & 0.548 & 0.547 & 0.49\\
\qwen{} & 0.538 & 0.383 & 0.025\\
\llama{} & 0.535 & 0.381 & 0.03\\
\bottomrule
\end{tabular}
}
\end{table*}

\begin{table*}[!h]
\small
\centering
\caption{\revised{Comparison of zero-shot performance of \qwen{} for different input modalities}}
\label{table:mllm_modalities}
{\scriptsize
\begin{tabular}{cccc}
\toprule
Modalities & Accuracy & Macro F1 & \makecell{proportion of\\predicted positive} \\
\cmidrule(rl){1-4}
text & 0.79 [$\pm$ 0.025] & 0.537 [$\pm$ 0.02] & 0.052 [$\pm$ 0.013]\\
text+video & 0.454 [$\pm$ 0.025] & 0.361 [$\pm$ 0.02] & 0.052 [$\pm$ 0.013]\\
video & 0.538 [$\pm$ 0.025] & 0.383 [$\pm$ 0.02] & 0.025 [$\pm$ 0.013]\\
\bottomrule
\end{tabular}
}
\end{table*}

\revised{Table \ref{table:res_EN_HNS} reports Macro F1 (as the answers provide discrete classification results) and shows the performance obtained for the task EN vs HN$\cup$S. For comparison, the first line shows the performance of a X-CLIP+Transf model.
While this task is balanced (48\% of samples are annotated as positive), we observe a general tendency of MLLMs to predict the negative class. In particular, the models \qwen{} and \llama{} almost systematically predict the negative class, leading to a F1-macro performance worse than random. \llava{} is the only MLLM model predicting positive in a similar proportion to the true labels, but performs hardly better than random in terms of Macro F1. The best MLLMs models are \intern{} and \internvl{}, which achieve non-trivial performance; however, because of the aforementioned tendency, they still perform worse.
Table \ref{table:mllm_modalities} compare results of \qwen{} when fed with different modalities, without changing the above conclusion.}

\paragraph{\revised{Few-shot inference}}

\revised{We inspire on recent works by \cite{fujii_viola_2026} and \cite{kim_videoicl_2025} to test few-shot learning with MLLMs.
The choice of few-shot demonstrations is characterized by the initial demonstration pool, and the selection strategy from this demonstration pool.
For the demonstration pool, we consider the training and validation set of each fold.
For the selection strategy, we use the X-CLIP embeddings of the clips to compute the cosine similarity between the test sample and all observations from the demonstration pool. We select the 4 demonstration examples with the largest similarity under the condition the example classes are balanced. The order in which examples are presented is random.
In few-shot, compared to the zero-shot setting, for a given task description there is more flexibility in the structure of the prompt: whether few-shot examples are presented at the beginning, or at the end, and how they are described.}

\revised{Table \ref{table:res_fewshot} shows that few-shot learning does not readily improve the results.
Qualitative analysis shows that, in few-shot settings, models can hardly segment the different video clips that they are given. As a consequence, it is not surprising that their Macro F1 is hardly better than random.
An interesting observation, however, is that while \llama{} systematically predicts negative, in the few-shot setting the proportion of positives predicted by \qwen{} increases.}
\begin{table*}[!h]
\small
\centering
\caption{Performance on EN vs HN$\cup$S with few-shot inference, with examples first}
\label{table:res_fewshot}
{\scriptsize
\begin{tabular}{cccc}
\toprule
Model & Accuracy & Macro F1 & \makecell{proportion of\\predicted positive} \\
\cmidrule(rl){1-4}
\qwen{} & 0.595 & 0.552 & 0.22\\
\intern{} & 0.543 & 0.517 & 0.30\\
\llama{} & 0.527 & 0.371 & 0.03\\
\bottomrule
\end{tabular}
}
\end{table*}

\revised{While MLLMs are interesting models for the new video interpretive task we introduce, this section shows that they do not readily provide satisfying results and require more work. In the line of recent works for video anomaly detection by \cite{vad-r1-neurips}, approaches with supervised fine-tuning and reinforcement learning leveraging rich concept supervision, are promising perspectives.}

\subsection{Interest of MObyGaze to improve multimodal and explainable models}\label{sec:interest}
We now demonstrate two applications of interest for MObyGaze. In Sec. \ref{sec:ICIPElisa}, we show how rich concept annotations can be leveraged to improve multimodal models for our video interpretation task. In Sec. \ref{sec:julieMM}, we show how MObyGaze raises new challenges for explainable AI and exemplify a contribution to achieve better accuracy-interpretability trade-off for our task.

\subsubsection{Improving multimodal models with concept supervision}\label{sec:ICIPElisa}

\revised{The objective here is to show that concept supervision can help multimodal training. For this, we first introduce the idea of Concept Modality Specific Dataset (CMSD) below. The models used for the experiments below are the same as in Sec. \ref{sec:concept_detec} with the same unimodal pre-extractors. Only, for the sake of computational intensity and to relate to the approach taken by \cite{bose_mm-autowards_2023}, we consider a Perceiver IO architecture instead of the usual Transformer in X-CLIP+Transf or BERT+Transf (we highlight that Perceiver IO is hence not fed with raw data but with pre-extracted features). This work is more detailed in \cite{elisaArxivICIP}.}

As we have seen in Sec.~\ref{sec:mm}, benefiting from several modalities with varying levels of relevance is challenging.
Here we show that the concept annotations can be used to improve multimodal models. Depending on how concept information is considered, we define two dataset versions: the Concept Modality Agnostic Dataset (CMAD) and the Concept Modality Specific Dataset (CMSD). In CMAD, a sample is labeled as positive regardless of the modality in which the concept appears. In contrast, CMSD labels a sample as positive only if the annotation contains at least one concept that is relevant to the specific modality.
We study the interest of training unimodal and multimodal models on CMAD or CMSD, while always testing on CMSD. Indeed, CMSD can be seen as the truthful label we desire the model to output as, e.g., we want a vision model to predict the negative class if a clip is labelled as objectifying but not owing to visual factors.

When considering a single modality (vision or text), we verify in Fig.~\ref{fig:video_mod} and Fig.~\ref{fig:text_mod} that unimodal models trained on CMSD (``Concept'' in the figure) perform significantly better than those trained on CMAD (``No concept'' in the figure). This already highlights the interest of concept annotation in multimodal tasks.

Fig. \ref{fig:video_audio}-\ref{fig:text_audio} present the performance of bimodal models, and Fig. \ref{fig:all_modalities} that of a trimodal model. We consider two major fusion strategies: \revised{mid-}Early fusion (``\revised{mid-}Early'') where modality tokens are concatenated at input as for our model in Sec. \ref{sec:mm}, and Late fusion (``Late'') where unimodal models are first pre-trained (on CMAD and CMSD), with their individual decision later fused with a simple max operation. For the bimodal models (Fig. \ref{fig:video_audio}-\ref{fig:text_audio}), we again observe the interest to use concept information (CMSD) at training time, regarless of the fusion type. For the trimodal model with \revised{mid-}Early fusion (Fig. \ref{fig:all_modalities}) can only be trained on CMAD. 
However, Late fusion, while performing worse that \revised{mid-}Early fusion when trained on CMAD, is able to catch up to \revised{mid-}Early fusion performance when its composing unimodal models are trained on CMSD.
This is all the more interesting that Late fusion architectures are more explainable that \revised{mid-}Early fusion because the information on what modality drives the decision is readily accessible.

While these results show the interest of exploiting concept annotation in the design of multimodal models for MObyGaze, we further the question of designing explainable models for MObyGaze in the next section.

\begin{figure}[!ht]
    \centering
    
    \hspace*{-0.5cm} 
    \begin{minipage}[t]{0.32\textwidth}
        \centering
        \resizebox{\linewidth}{!}{
\resizebox{0.48\columnwidth}{!}{
\begin{tikzpicture}

\definecolor{darkgray176}{RGB}{176,176,176}
\definecolor{gray}{RGB}{128,128,128}
\definecolor{mediumseagreen7617580}{RGB}{76,175,80}
\definecolor{peachpuff}{RGB}{255,218,185}

\begin{axis}[
tick align=outside,
tick pos=left,
label style={font=\LARGE},
tick label style={font=\Large},
x grid style={darkgray176},
xmin=0.25, xmax=0.4,
xtick style={color=black},
xtick={0.3,0.35},
xticklabels={No concept,Concept},
y grid style={darkgray176},
ylabel={PR AUC},
ymajorgrids,
ymin=0.6, ymax=0.75,
ytick style={color=black}
]
\draw[draw=none,fill=blue,fill opacity=0.3] (axis cs:0.285,0) rectangle (axis cs:0.315,0.669);

\draw[draw=none,fill=blue,fill opacity=0.7] (axis cs:0.335,0) rectangle (axis cs:0.365,0.711);

\path [draw=black, semithick]
(axis cs:0.3,0.65)
--(axis cs:0.3,0.688);

\path [draw=black, semithick]
(axis cs:0.35,0.696)
--(axis cs:0.35,0.726);

\path [draw=gray, semithick]
(axis cs:0.3,0.65)
--(axis cs:0.3,0.688);

\path [draw=gray, semithick]
(axis cs:0.35,0.696)
--(axis cs:0.35,0.726);

\addplot [semithick, black, mark=-, mark size=5, mark options={solid}, only marks]
table[row sep=crcr]{%
0.3 0.65\\
};
\addplot [semithick, black, mark=-, mark size=5, mark options={solid}, only marks]
table[row sep=crcr]{%
0.3 0.688\\
};
\addplot [semithick, black, mark=-, mark size=5, mark options={solid}, only marks]
table[row sep=crcr]{%
0.35 0.696\\ 
};
\addplot [semithick, black, mark=-, mark size=5, mark options={solid}, only marks]
table[row sep=crcr]{%
0.35 0.726
\\ };
\addplot [semithick, gray, mark=-, mark size=5, mark options={solid}, only marks]
table[row sep=crcr]{%
0.3 0.65\\ 
};
\addplot [semithick, gray, mark=-, mark size=5, mark options={solid}, only marks]
table[row sep=crcr]{%
0.3 0.688\\ 
};
\addplot [semithick, gray, mark=-, mark size=5, mark options={solid}, only marks]
table[row sep=crcr]{%
0.35 0.696\\ 
};
\addplot [semithick, gray, mark=-, mark size=5, mark options={solid}, only marks]
table[row sep=crcr]{%
0.35 0.726\\ 
};

\end{axis}

\end{tikzpicture}
}}
        \subcaption{Video modality}
        \label{fig:video_mod}
    \end{minipage}
    \hspace{0.5cm} 
    \begin{minipage}[t]{0.32\textwidth}
        \centering
        \resizebox{\linewidth}{!}{
\resizebox{0.48\columnwidth}{!}{
\begin{tikzpicture}

\definecolor{darkgray176}{RGB}{176,176,176}
\definecolor{gray}{RGB}{128,128,128}
\definecolor{mediumseagreen7617580}{RGB}{76,175,80}
\definecolor{peachpuff}{RGB}{255,218,185}

\begin{axis}[
tick align=outside,
tick pos=left,
label style={font=\LARGE},
tick label style={font=\Large},
x grid style={darkgray176},
xmin=0.25, xmax=0.4,
xtick style={color=black},
xtick={0.3,0.35},
xticklabels={No concept,Concept},
y grid style={darkgray176},
ylabel={PR AUC},
ymajorgrids,
ymin=0.2, ymax=0.5,
ytick style={color=black}
]
\draw[draw=none,fill=yellow,fill opacity=0.3] (axis cs:0.285,0) rectangle (axis cs:0.315,0.267);

\draw[draw=none,fill=yellow,fill opacity=0.7] (axis cs:0.335,0) rectangle (axis cs:0.365,0.4365);

\path [draw=black, semithick]
(axis cs:0.3,0.229)
--(axis cs:0.3,0.305);

\path [draw=black, semithick]
(axis cs:0.35,0.386)
--(axis cs:0.35,0.487);

\path [draw=gray, semithick]
(axis cs:0.3,0.229)
--(axis cs:0.3,0.305);

\path [draw=gray, semithick]
(axis cs:0.35,0.386)
--(axis cs:0.35,0.487);

\addplot [semithick, black, mark=-, mark size=5, mark options={solid}, only marks]
table[row sep=crcr]{%
0.3 0.229\\
};
\addplot [semithick, black, mark=-, mark size=5, mark options={solid}, only marks]
table[row sep=crcr]{%
0.3 0.305\\
};
\addplot [semithick, black, mark=-, mark size=5, mark options={solid}, only marks]
table[row sep=crcr]{%
0.35 0.386\\
};
\addplot [semithick, black, mark=-, mark size=5, mark options={solid}, only marks]
table[row sep=crcr]{%
0.35 0.487\\
};
\addplot [semithick, gray, mark=-, mark size=5, mark options={solid}, only marks]
table[row sep=crcr]{%
0.3 0.229\\
};
\addplot [semithick, gray, mark=-, mark size=5, mark options={solid}, only marks]
table[row sep=crcr]{%
0.3 0.305\\
};
\addplot [semithick, gray, mark=-, mark size=5, mark options={solid}, only marks]
table[row sep=crcr]{%
0.35 0.386\\
};
\addplot [semithick, gray, mark=-, mark size=5, mark options={solid}, only marks]
table[row sep=crcr]{%
0.35 0.487\\
};

\end{axis}

\end{tikzpicture}
represented}}
        \subcaption{Text modality}
        \label{fig:text_mod}
    \end{minipage}
    \hfill
    
    \vspace{0.3cm} 
    
    \begin{minipage}[t]{0.32\textwidth}
        \centering
        \resizebox{\linewidth}{!}{
\resizebox{0.48\columnwidth}{!}{
\begin{tikzpicture}

\definecolor{darkgray176}{RGB}{176,176,176}
\definecolor{gray}{RGB}{128,128,128}
\definecolor{lightgray204}{RGB}{204,204,204}
\definecolor{magenta}{RGB}{255,0,255}

\begin{axis}[
legend cell align={left},
legend columns=2,
legend style={
  fill opacity=0.8,
  draw opacity=1,
  text opacity=1,
  at={(0.5,1)},
  anchor=north,
  draw=lightgray204
},
tick align=outside,
tick pos=left,
label style={font=\LARGE},
tick label style={font=\Large},
x grid style={darkgray176},
xmin=-0.5, xmax=1.5,
xtick style={color=black},
xtick={0,1},
xticklabels={Late,Early},
y grid style={darkgray176},
ylabel={PR AUC},
ymajorgrids,
ymin=0.6, ymax=0.8,
ytick style={color=black}
]

\draw[draw=none,fill=magenta,fill opacity=0.3] (axis cs:-0.2,0) rectangle (axis cs:0,0.7085);
\draw[draw=none,fill=magenta,fill opacity=0.3] (axis cs:0.8,0) rectangle (axis cs:1,0.725);
\addlegendimage{ybar,area legend,draw=none,fill=magenta,fill opacity=0.3} 
\addlegendentry{No Concept}

\draw[draw=none,fill=magenta,fill opacity=0.7] (axis cs:-1.38777878078145e-17,0) rectangle (axis cs:0.2,0.736);
\draw[draw=none,fill=magenta,fill opacity=0.7] (axis cs:1,0) rectangle (axis cs:1.2,0.755);
\addlegendimage{ybar,area legend,draw=none,fill=magenta,fill opacity=1} 
\addlegendentry{Concept}



\path [draw=gray, semithick]
(axis cs:-0.1,0.69)
--(axis cs:-0.1,0.727);

\path [draw=gray, semithick]
(axis cs:0.9,0.705)
--(axis cs:0.9,0.745);

\path [draw=gray, semithick]
(axis cs:0.1,0.715)
--(axis cs:0.1,0.757);

\path [draw=gray, semithick]
(axis cs:1.1,0.733)
--(axis cs:1.1,0.777);

\addplot [semithick, gray, mark=-, mark size=5, mark options={solid}, only marks]
table[row sep=crcr]{%
-0.1 0.69\\
-0.1 0.727\\
0.9 0.705\\
0.9 0.745\\
0.1 0.715\\
0.1 0.757\\
1.1 0.733\\
1.1 0.777\\
};

\end{axis}

\end{tikzpicture}
}}
        \subcaption{Video+Audio}
        \label{fig:video_audio}
    \end{minipage}
    \hfill
    \begin{minipage}[t]{0.32\textwidth}
        \centering
        \resizebox{\linewidth}{!}{
\resizebox{0.48\columnwidth}{!}{
\begin{tikzpicture}

\definecolor{darkgray176}{RGB}{176,176,176}
\definecolor{gray}{RGB}{128,128,128}
\definecolor{lightgray204}{RGB}{204,204,204}
\definecolor{orange}{RGB}{255,165,0}

\begin{axis}[
legend cell align={left},
legend columns=3,
legend style={
  fill opacity=0.8,
  draw opacity=1,
  text opacity=1,
  at={(0.5,0.98)},
  anchor=north,
  draw=lightgray204
},
tick align=outside,
tick pos=left,
label style={font=\LARGE},
tick label style={font=\Large},
x grid style={darkgray176},
xmin=-0.5, xmax=1.5,
xtick style={color=black},
xtick={0,1},
xticklabels={Late,Early},
y grid style={darkgray176},
ylabel={PR AUC},
ymajorgrids,
ymin=0.2, ymax=0.6,
ytick style={color=black}
]


\draw[draw=none,fill=orange,fill opacity=0.3] (axis cs:-0.2,0) rectangle (axis cs:0,0.375);
\draw[draw=none,fill=orange,fill opacity=0.3] (axis cs:0.8,0) rectangle (axis cs:1,0.373);
\addlegendimage{ybar,area legend,draw=none,fill=orange,fill opacity=0.3}
\addlegendentry{No Concept}

\draw[draw=none,fill=orange,fill opacity=0.7] (axis cs:-1.38777878078145e-17,0) rectangle (axis cs:0.2,0.4715);
\draw[draw=none,fill=orange,fill opacity=0.7] (axis cs:1,0) rectangle (axis cs:1.2,0.4845);
\addlegendimage{ybar,area legend,draw=none,fill=orange,fill opacity=1} 
\addlegendentry{Concept}



\path [draw=gray, semithick]
(axis cs:-0.1,0.332)
--(axis cs:-0.1,0.418);

\path [draw=gray, semithick]
(axis cs:0.9,0.331)
--(axis cs:0.9,0.415);

\path [draw=gray, semithick]
(axis cs:0.1,0.419)
--(axis cs:0.1,0.524);

\path [draw=gray, semithick]
(axis cs:1.1,0.437)
--(axis cs:1.1,0.532);

\addplot [semithick, gray, mark=-, mark size=5, mark options={solid}, only marks]
table[row sep=crcr]{%
-0.1 0.332\\
-0.1 0.418\\
0.9 0.331\\
0.9 0.415\\
0.1 0.419\\
0.1 0.524\\
1.1 0.437\\
1.1 0.532\\
};

\end{axis}

\end{tikzpicture}
}}
        \subcaption{Text+Audio}
        \label{fig:text_audio}
    \end{minipage}
    \hfill
    \begin{minipage}[t]{0.32\textwidth}
        \centering
        \resizebox{\linewidth}{!}{
\resizebox{0.9\columnwidth}{!}{
\begin{tikzpicture}

\definecolor{darkgray176}{RGB}{176,176,176}
\definecolor{gray}{RGB}{128,128,128}
\definecolor{lightgray204}{RGB}{204,204,204}
\definecolor{green}{RGB}{0,255,0}

\begin{axis}[
legend cell align={left},
legend columns=2,
legend style={
  fill opacity=1,
  draw opacity=1,
  text opacity=1,
  at={(0.5,0.98)},
  anchor=north,
  draw=lightgray204
},
tick align=outside,
tick pos=left,
label style={font=\LARGE},
tick label style={font=\Large},
x grid style={darkgray176},
xmin=-0.5, xmax=1.5,
xtick style={color=black},
xtick={0,1},
xticklabels={Late,Early},
y grid style={darkgray176},
ylabel={PR AUC},
ymajorgrids,
ymin=0.6, ymax=0.9,
ytick style={color=black}
]

\draw[draw=none,fill=green,fill opacity=0.3] (axis cs:-0.25,0) rectangle (axis cs:0,0.754);
\draw[draw=none,fill=green,fill opacity=0.3] (axis cs:0.75,0) rectangle (axis cs:1,0.813);
\addlegendimage{ybar,area legend,draw=none,fill=green,fill opacity=0.3} 
\addlegendentry{No Concept}

\draw[draw=none,fill=green,fill opacity=0.7] (axis cs:0,0) rectangle (axis cs:0.25,0.7885);
\draw[draw=none,fill=green,fill opacity=0.7] (axis cs:1,0) rectangle (axis cs:1.25,0.8165);
\addlegendimage{ybar,area legend,draw=none,fill=green,fill opacity=0.7} 
\addlegendentry{Concept}


\path [draw=gray, semithick]
(axis cs:-0.125,0.718)
--(axis cs:-0.125,0.79);

\path [draw=gray, semithick]
(axis cs:0.875,0.802)
--(axis cs:0.875,0.824);

\path [draw=gray, semithick]
(axis cs:0.125,0.765)
--(axis cs:0.125,0.812);

\path [draw=gray, semithick]
(axis cs:1.125,0.804)
--(axis cs:1.125,0.829);

\addplot [semithick, gray, mark=-, mark size=5, mark options={solid}, only marks]
table[row sep=crcr]{%
-0.125 0.718\\
-0.125 0.79\\
0.875 0.802\\
0.875 0.824\\
0.125 0.765\\
0.125 0.812\\
1.125 0.804\\
1.125 0.829\\
};

\end{axis}

\end{tikzpicture}
}}
        \subcaption{All three}
        \label{fig:all_modalities}
    \end{minipage}
    
    \caption{Performance of unimodal (a, b) bimodal (c, d) and trimodal (e) models trained without or with concept supervision. Results are on the test set of CMSD.}
    \label{fig:modality_analysis}
    \vspace{-0.2cm}
\end{figure}

\subsubsection{Improving explainable models for challenging video interpretation}\label{sec:julieMM}

In this section we exemplify some challenges our MOByGaze dataset and task raise for existing explainable AI approaches. The results below are more extensively presented in \cite{julieACMMM}. \revised{We also show another example of how concept supervision can help for the downstream classification task, and which challenge remain.}

The use of auxiliary conceptual information in image classification tasks is at the heart of important work in explainable deep learning, especially with so-called Concept Bottleneck Models (CBMs) introduced by \cite{pmlr-v119-koh20a}. They have given rise to a line of subsequent works, such as Concept Embedding Models (CEMs) by \cite{CEM}.
These methods are classically developed for image classification tasks and datasets with simple conceptual information (e.g., the color of a bird's beak).
\cite{ramaswamy_overlooked_2023} have analyzed the factors that participate in the success of concept-based explanations: explanatory concepts should be limited (no more than 32) to be truly useful for human interpretation, and they must be easy to learn. 
MObyGaze, which centers on a complex multimodal video interpretation task aided with high-level concepts, therefore challenges this validity domain for concept-based models because the complexity of the end label would require numerous concepts if these were as simple as color or texture areas (such as in \cite{cub} where there are 112 concepts).
Instead, we choose with MObyGaze to have a limited number of concepts (11 in total, 8 visual) by having each concept itself interpretive and hence possibly not easy to learn. We investigate the ensuing challenges for concept-based models.

\paragraph{Analysis of MObyGaze for existing concept-based models}
We first position MObyGaze with respect to two classical concept-based image datasets on the dimensions of \textit{Concept-task determination} and \textit{Concept detection}, shown in the y-axis and x-axis of Fig. \ref{fig:AUC-CAS_2}.a, respectively. The \textit{Concept-task determination} quantifies how much the knowledge of which concepts are present is sufficient to determine the end label, and is measured by the performance of a reference CBM fed with the ground-truth presence of each concept. The \textit{Concept detection} score is the mean F1-score of concept detectors jointly trained within the reference CEM.
We consider three datasets: two reference datasets annotated with concepts, CUB (\cite{cub}) and CelebA (\cite{CelebA}), and MObyGaze with visual concepts only. CUB is a dataset of bird images annotated for their species and with 112 curated concepts (such as beak color). CelebA is an image dataset of human faces annotated with their ID and a set of 40 concepts (including ``eyeglasses'' but also ethically questionable ones such as ``attractive'' or ``narrow eyes'').

Fig. \ref{fig:julie_datasets} shows that, while CEM achieves good concept detection on CUB and CelebA, its detection performance on MObyGaze (despite being adapted with proper loss weights to balance heterogeneous concept representations) is significantly lower, and is worse than concept detectors trained outside CEM and shown in Sec. \ref{sec:visual_concept}. This illustrates a difficulty of legacy concept-based models such as CBM and CEM to deal with complex tasks with relatively more complex concepts.

\paragraph{A new concept-based model for MObyGaze}
To tackle this challenge, in \cite{julieACMMM}, we introduce a novel CEM-type architecture that we name Concept Pre-trained Transformer Frozen (CPT-F). It replaces MLP-based layers by Transformer-based layers and leverages frozen pre-trained concept detectors, specifically incorporated to address the severe class imbalance present in the MObyGaze dataset.

Fig. \ref{fig:EN_HNS_CEM_vs_CPT-F} and Fig. \ref{fig:ENHN_S_CEM_vs_CPT-F} show the \textbf{End-task performance - Interpretability trade-off}. Performance (y-axis) is measured in PR AUC. Interpretability (x-axis) is measured with the Concept Alignment Score (CAS) introduced by \cite{CEM}. CAS represents the homogeneity of concept labels amongst clusters of samples in the latent space used for end-task decision. High values of CAS indicate a latent space shaped by concepts, hence a decision relying heavily on concept presence but represented in a latent space, without constraining the decision to a binary concept bottleneck like with CBMs. 

We observe that for the end-task of objectification classification EN vs HN$\cup$S (Center (b)), which is the easiest in that it boils down to detecting the presence of at least one concept, CPT-F does not allow to gain in task performance compared with CEM, but it increases interpretability thanks to improved concept detection. 
For the harder end-task of objectification classification EN$\cup$HN vs S (Left (c)), CPT-F allows to gain both in interpretability and performance, as the difference between HN and S requires finer concept counts.

\revised{It is worth noting that, despite the concepts being high-level and hence harder to detect on an individual basis, they nonetheless help for the downstream classification task with the proposed architecture: results in PR AUC Macro in Table \ref{table:res_all_visual} and Fig. \ref{fig:AUC-CAS_2} can be compared and}\\
\revised{$\bullet$ on the EN vs HN$\cup$S task, PR AUC Macro: X-CLIP+Transf 0.739, CEM 0.785,}\\
\revised{$\bullet$ on the EN$\cup$HN vs S task, PR AUC Macro: X-CLIP+Transf 0.613, CEM 0.628.}

We have therefore shown that MObyGaze represents a valuable asset to design better and more effective explainable models for more interpretive tasks.

\begin{figure}
    \centering
    \begin{minipage}[t]{0.32\textwidth}
        \centering
\begin{tikzpicture}[scale=0.5]

\definecolor{crimson2143940}{RGB}{214,39,40}
\definecolor{darkgray176}{RGB}{176,176,176}
\definecolor{darkorange25512714}{RGB}{255,127,14}
\definecolor{forestgreen4416044}{RGB}{44,160,44}
\definecolor{gray}{RGB}{128,128,128}
\definecolor{lightgray204}{RGB}{204,204,204}
\definecolor{steelblue31119180}{RGB}{31,119,180}

\begin{groupplot}[group style={group size=1 by 1}]

\nextgroupplot[
scale=1.0,
        grid=major,
        xlabel={Concept Detection (F1)},
        ylabel={Concept-task determination (F1)}, 
        scatter/classes={
            A={mark=*, fill={rgb:red,0.25;green,1;blue,0}},
            B={mark=*, fill={rgb:red,0.1;green,0.5;blue,1}}, 
            C={mark=*, fill={rgb:red,0.4;green,0.75;blue,0}},
            D={mark=*, fill={rgb:red,1;green,0.1;blue,0.6}}, 
            E={mark=*, fill={rgb:red,0.5;green,0.5;blue,0}}
        },
        legend style={at={(0, -0.4)}, anchor=west, legend columns=2},%
        xmin=0.4, xmax=1, 
        ymin=0.1, ymax=1.2,  
]
\addplot[scatter,only marks, scatter src=explicit symbolic, mark size=5pt]
    table [meta=label] {
        y    z    label
        0.45  0.99  A
        0.84  1    B
        0.46  0.738 C
        0.85  0.26 D
        0.45  0.783 E

    };
\legend{MObyGaze EN vs HN$\cup$S, CUB, MObyGaze EN$\cup$HN vs S, CelebA, MObyGaze EN vs HN vs S}

\end{groupplot}

\end{tikzpicture}
        \vspace*{-0.6cm}
        \subcaption{Datasets}
        \label{fig:julie_datasets}
    \end{minipage}
    \hfill
    \begin{minipage}[t]{0.32\textwidth}
        \centering
\begin{tikzpicture}[scale=0.6]

\definecolor{darkgray176}{RGB}{176,176,176}
\definecolor{darkorange25512714}{RGB}{255,127,14}
\definecolor{lightgray204}{RGB}{204,204,204}
\definecolor{steelblue31119180}{RGB}{31,119,180}

\begin{axis}[
legend cell align={left},
legend style={at={(0.1, 0.2)}, anchor=west},%
tick align=outside,
tick pos=left,
x grid style={darkgray176},
xlabel={CAS},
xmin=60, xmax=67,
xtick style={color=black},
y grid style={darkgray176},
ylabel={PR AUC Macro},
ymin=0.77, ymax=0.80,
ytick style={color=black},
grid=major
]
\path [draw=steelblue31119180, semithick]
(axis cs:61.196,0.787)
--(axis cs:61.724,0.787);

\path [draw=steelblue31119180, semithick]
(axis cs:61.46,0.784)
--(axis cs:61.46,0.79);

\addplot [semithick, steelblue31119180, mark=*, mark size=3, mark options={solid}, only marks]
table {%
61.46 0.787
};
\addlegendentry{CEM}
\addplot [semithick, darkorange25512714, mark=*, mark size=3, mark options={solid}, only marks]
table {%
65.296 0.784
};
\addlegendentry{CPT-F}

\addplot [semithick, steelblue31119180, mark=|, mark size=5, mark options={solid}, only marks]
table {%
61.196 0.787
};
\addplot [semithick, steelblue31119180, mark=|, mark size=5, mark options={solid}, only marks]
table {%
61.724 0.787
};
\addplot [semithick, steelblue31119180, mark=-, mark size=5, mark options={solid}, only marks]
table {%
61.46 0.784
};
\addplot [semithick, steelblue31119180, mark=-, mark size=5, mark options={solid}, only marks]
table {%
61.46 0.79
};
\path [draw=darkorange25512714, semithick]
(axis cs:65.032,0.784)
--(axis cs:65.56,0.784);

\path [draw=darkorange25512714, semithick]
(axis cs:65.296,0.781)
--(axis cs:65.296,0.787);

\addplot [semithick, darkorange25512714, mark=|, mark size=5, mark options={solid}, only marks]
table {%
65.032 0.784
};
\addplot [semithick, darkorange25512714, mark=|, mark size=5, mark options={solid}, only marks]
table {%
65.56 0.784
};
\addplot [semithick, darkorange25512714, mark=-, mark size=5, mark options={solid}, only marks]
table {%
65.296 0.781
};
\addplot [semithick, darkorange25512714, mark=-, mark size=5, mark options={solid}, only marks]
table {%
65.296 0.787
};
]
\end{axis}

\end{tikzpicture}
        \vspace*{-0.4cm}
        \subcaption{EN vs. HN$\cup$S}
        \label{fig:EN_HNS_CEM_vs_CPT-F}
    \end{minipage}
    \hfill
    \begin{minipage}[t]{0.32\textwidth}
        \centering
\begin{tikzpicture}[scale=0.6]

\definecolor{darkgray176}{RGB}{176,176,176}
\definecolor{darkorange25512714}{RGB}{255,127,14}
\definecolor{lightgray204}{RGB}{204,204,204}
\definecolor{steelblue31119180}{RGB}{31,119,180}

\begin{axis}[
legend cell align={left},
legend style={at={(0.1, 0.2)}, anchor=west},%
tick align=outside,
tick pos=left,
x grid style={darkgray176},
xlabel={CAS},
xmin=60, xmax=67,
xtick style={color=black},
y grid style={darkgray176},
ylabel={PR AUC Macro},
ymin=0.60, ymax=0.65,
ytick style={color=black},
grid=major
]
\path [draw=steelblue31119180, semithick]
(axis cs:61.008,0.628)
--(axis cs:61.612,0.628);

\path [draw=steelblue31119180, semithick]
(axis cs:61.31,0.626)
--(axis cs:61.31,0.63);

\addplot [semithick, steelblue31119180, mark=*, mark size=3, mark options={solid}, only marks]
table {%
61.31 0.628
};
\addlegendentry{CEM}
\addplot [semithick, darkorange25512714, mark=*, mark size=3, mark options={solid}, only marks]
table {%
65.266 0.645
};
\addlegendentry{CPT-F}
\addplot [semithick, steelblue31119180, mark=|, mark size=5, mark options={solid}, only marks]
table {%
61.008 0.628
};
\addplot [semithick, steelblue31119180, mark=|, mark size=5, mark options={solid}, only marks]
table {%
61.612 0.628
};
\addplot [semithick, steelblue31119180, mark=-, mark size=5, mark options={solid}, only marks]
table {%
61.31 0.626
};
\addplot [semithick, steelblue31119180, mark=-, mark size=5, mark options={solid}, only marks]
table {%
61.31 0.63
};
\path [draw=darkorange25512714, semithick]
(axis cs:64.964,0.645)
--(axis cs:65.568,0.645);

\path [draw=darkorange25512714, semithick]
(axis cs:65.266,0.643)
--(axis cs:65.266,0.647);

\addplot [semithick, darkorange25512714, mark=|, mark size=5, mark options={solid}, only marks]
table {%
64.964 0.645
};
\addplot [semithick, darkorange25512714, mark=|, mark size=5, mark options={solid}, only marks]
table {%
65.568 0.645
};
\addplot [semithick, darkorange25512714, mark=-, mark size=5, mark options={solid}, only marks]
table {%
65.266 0.643
};
\addplot [semithick, darkorange25512714, mark=-, mark size=5, mark options={solid}, only marks]
table {%
65.266 0.647
};

\end{axis}

\end{tikzpicture}
        \vspace*{-0.4cm}
        \subcaption{EN$\cup$HN vs. S}
        \label{fig:ENHN_S_CEM_vs_CPT-F}
    \end{minipage}
    \caption{(a): Comparison of the datasets on the concept detection difficulty (x-axis) and determination of the downstream task (y-axis).
    (b) and (c): End-task performance-Interpretability trade-off in terms of PR AUC and CAS for the tasks of objectification detection.}
    \label{fig:AUC-CAS_2}
\end{figure}
\section{Broader Impact Statement}\label{sec:lims_appls}

\paragraph{Limitations}
The main limitation of the MObyGaze dataset is that the annotation has been performed at scene-level, not allowing for supervision to learn longer-term objectification patterns. These are known to relate to narratology and occur recurrently throughout a movie, such as tropes as studied by \cite{tropes}. We intend to extend the dataset with such longer-term patterns thanks to the annotation tool, which allows for a multi-level and evolving thesaurus. It will also be possible by the availability of the experts who have developed fine-grained memory of the 20 movies. \revised{Another limitation is the low number of annotators, albeit their expertise, and the limited set of movies tied to the original definition of male gaze within the Hollywood film industry.}

\paragraph{Applications}
We have shown that detecting objectification, as defined for the MObyGaze dataset, is accessible to current vision and text models, with significant room for improvement. 
Immediate challenges are hence in designing models able to learn better representations of complex concept instances. This includes in particular designing explainable models to better characterize complex temporal and multimodal objectification patterns, which can in turn enrich qualitative studies by media scholars. Sub-constructs detection is also an an interesting future in this interdisciplinary context.
The MObyGaze dataset can also be used to study the fairness of existing computer vision models: person detectors and human pose estimators may miss the presence of characters onscreen, compromising the study of how certain patterns correlate with certain human groups, if the humans are often mis-detected for these patterns (e.g., shots with headless body parts as shown by \cite{wu_evaluation_2022}).
Finally, the social purpose of this work is to feed public debate and reflection on the tangibility of subtle widespread audiovisual patterns conveying biased gender representations. Designing and making publicly available models to detect objectification patterns can help raise awareness, but also serve for filmmaking training.

\paragraph{Potential misuse and mitigation measures}
As we expand in the datasheet in App. \ref{suppl:docu} under section Uses, the MObyGaze dataset should not be used for tasks such as content filtering, defining regulatory standards, or censorship of media content. Another potential negative impact would lie in the interpretation of the outputs of models, being it by social science researchers assisting their qualitative studies with models trained on MObyGaze to detect objectification, or by amateurs. We detail these cases we envision, and the corresponding proactive measures, which involve the licensing prohibiting commercial usage and the dataset documentation highlighting that the models trained on the MObyGaze dataset must be used in a clear ethical framework: that of making subtle patterns of representation disparities visible and contributing responsible usages.


\acks{This work was supported by the French National Research Agency through the ANR TRACTIVE project ANR-21-CE38-0012-01, and through the France 2030 investment plan under reference number ANR-15-IDEX-01. This work was partly supported by EU Horizon 2020 project AI4Media, under contract no. 951911 (https://ai4media.eu/). The authors are grateful to the Université Côte d’Azur’s Center for High-Performance Computing (OPAL infrastructure) for providing resources and support.}

\renewcommand{\bibfont}{\small}
\bibliography{biblio}

\newpage
\appendix
\section{Supplementary material for ``MObyGaze: a film dataset of multimodal objectification densely annotated by experts''}\label{sec:suppl}

\subsection{Datasheet documentation for the MObyGaze dataset}\label{suppl:docu}

The datasheet documentation below provides the necessary information required in the checklist, specifically:
\begin{itemize}
  \item Dataset documentation and intended uses.
  \item URL to website where the dataset can be downloaded by the reviewers and
  \item URL to Croissant metadata record documenting the dataset:\\
  The MObyGaze dataset artifacts are provided along with a Croissant description at \url{https://github.com/husky-helen/MObyGaze}
  \item Author statement: We bear all responsibility for the MObyGaze dataset, which is shared under a CC BY-NC-SA licence.
  \item Hosting, licensing, and maintenance plan are described in the datasheet below.
\end{itemize}

\begin{multicols}{2}





\definecolor{darkblue}{RGB}{46,25, 110}

\newcommand{\dssectionheader}[1]{%
   \noindent\framebox[\columnwidth]{%
      {\fontfamily{phv}\selectfont \textbf{\textcolor{darkblue}{#1}}}
   }
}

\newcommand{\dsquestion}[1]{%
    \noindent\fontfamily{phv}\selectfont \textcolor{darkblue}{\textbf{#1}}%
}

\newcommand{\dsquestionex}[2]{%
    \noindent\fontfamily{phv}\selectfont \textcolor{darkblue}{\textbf{#1} #2}%
}

\newcommand{\dsanswer}[1]{%
   \noindent #1\par\medskip
}

\dssectionheader{Motivation}

\dsquestion{For what purpose was the dataset created?}

\dsanswer{
The purpose of the MObyGaze dataset is to advance computational approaches to help make subtle patterns of bias in audiovisual content visible and more tangible, and quantify their prevalence. 
We create the MObyGaze dataset to enable the AI community to design computational approaches to characterize and quantify complex temporal and multimodal patterns of character objectification in films.
For this, we devise a thesaurus of objectification by building on existing characterization in film studies and cognitive and social psychology. 
The thesaurus articulates visual, speech and audio components, which we denote as \textit{concepts} involved in the production of objectification. 
The annotation process then consists in 2 experts densely annotating 20 movies: they manually delimit all the segments they find relevant for objectification, and label each with a level of objectification. 
To allow for fine-grained data and model analysis, they also annotate which objectification concepts are present.
}

\dsquestion{Who created this dataset?}

\dsanswer{The multidisciplinary team of authors composed of gender and media studies researchers, data scientists, and AI researchers from multiple research institutes and universities.
}

\dsquestion{Who funded the creation of the dataset?}

\dsanswer{The project was supported through public research funds mentioned in the acknowledgments section: the French National Reasearch Agency (ANR) and the European Commission (EC).
}


\bigskip
\dssectionheader{Composition}

\dsquestion{What do the instances that comprise the dataset represent (e.g., documents, photos, people, countries)?}

\begin{figure*}[ht]
  \centering
  \includegraphics[width=1.0\linewidth]{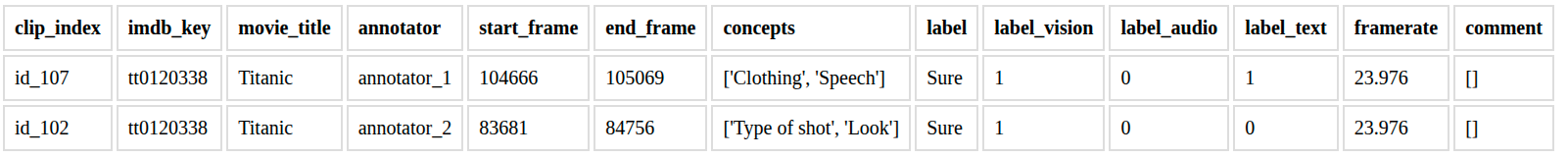}
  \caption{Example of two dataset instances.}
  \label{fig:datasheet_instance_example}
\end{figure*}

\dsanswer{
The dataset consists of annotations of 20 feature-length films, for which we consider the video track, the sound track and associated subtitles.
Each movie is annotated by 2 experts for a freely determined number of segments per movie. A dataset instance is therefore a video segment identified with its indices of start and end frame and start and end time-stamps, annotated with the objectification rating and thesaurus concepts tagged as present in the segment by the annotator. The annotator considered the image, sound dialogue modalities to annotate, and the dialogue transcript is also provided for each segment. Fig. \ref{fig:datasheet_instance_example} shows an example of two dataset instances.
The files we provide are:\\
\noindent $\bullet$ the list of films (mobygaze$\_$movielist.csv), also reported in Table \ref{table:datasheet_films};\\
\noindent $\bullet$ the objectification thesaurus (objectification-thesaurus.json) listing the concepts and their instances the annotators used to annotate the films;\\
\noindent $\bullet$ the entire table of annotated segments (mobygaze$\_$dataframe.csv). One segment corresponds to an interval of a movie delimited by a given annotator. Fig. \ref{fig:datasheet_instance_example} shows an example;\\
\noindent $\bullet$ the SQL description of the dataset as a database, with tables annotations, movies and subtitles (Neurips.sql).
}

\begin{table*}[ht]
\small
\caption{List of films annotated in the MObyGaze dataset (sorted by genre)}
\label{table:datasheet_films} 
{\scriptsize
\begin{tabular}{lllllll}
\toprule
IMDB key & Movie Title & Duration & Year & Genre & Test Fold & \makecell{Validation\\Fold}\\
\cmidrule(lr){1-7}
tt0097576 & \makecell{Indiana Jones and the \\Last Crusade} & 2h 6min & 1989 & adventure, action & 1 & 5\\
tt1454029 & The Help & 2h 26min & 2011 & drama & 2 & \\
tt1285016 & The Social Network & 2h 0min & 2010 & drama, biographical & 3 & 4\\
tt0467406 & Juno & 1h 36min & 2007 & drama, comedy & 4 & \\
tt0110912 & Pulp Fiction & 2h 34min & 1994 & drama, crime & 5 & 3\\
tt0822832 & Marley \& Me & 1h 55min & 2008 & drama, family & 1 & \\
tt1568346 & \makecell{The Girl with the \\Dragon Tattoo} & 2h 38min & 2011 & drama, mystery, crime & 2 & \\
tt2267998 & Gone Girl & 2h 29min & 2014 & drama, mystery, thriller & 3 & 2\\
tt0109830 & Forrest Gump & 2h 22min & 1994 & drama, romantic & 4 & 1\\
tt0120338 & Titanic & 3h 14min & 1997 & drama, romantic & 5 & \\
tt0108160 & Sleepless in Seattle & 1h 45min & 1993 & drama, romantic, comedy & 1 & 5\\
tt0119822 & As Good as It Gets & 2h 18min & 1997 & drama, romantic, comedy & 2 & \\
tt1193138 & Up in the Air & 1h 49min & 2009 & drama, romantic, comedy & 3 & 4\\
tt1570728 & Crazy,Stupid,Love. & 1h 58min & 2011 & drama, romantic, comedy & 4\\ 
tt1045658 & Silver Linings Playbook & 2h 2min & 2012 & drama, romantic, comedy & 5 & 3\\
tt0970416 & \makecell{The Day the Earth \\Stood Still} & 1h 43min & 2008 & drama, sci-fi, adventure & 1 & \\
tt1907668 & Flight & 2h 18min & 2012 & drama, thriller & 2 & \\
tt0375679 & Crash & 1h 55min & 2004 & drama, thriller, crime & 3 & 2\\
tt1142988 & The Ugly Truth & 1h 35min & 2009 & romantic, comedy & 4 & 1\\
tt1632708 & Friends with benefits & 1h 49min & 2011 & romantic, comedy & 5 & \\
\bottomrule
\end{tabular}
}
\end{table*}

\dsquestion{How many instances are there in total (of each type, if appropriate)?}

\dsanswer{
There are 20 films annotated by two annotators, yielding in total 6072 segments delimited and annotated.
}

\dsquestion{Does the dataset contain all possible instances or is it a sample (not necessarily random) of instances from a larger set?}

\dsanswer{The dataset contains all the instances produced by both annotators who have entirely annotated each movie.
The 20 movies of MObyGaze are a subset of the 51 movies of the pre-existing MovieGraphs dataset. The 20 movies have been chosen to maintain the ratio of represented genres and diversify to the maximum the actors/directors represented.
Table \ref{table:datasheet_films} provides details on the movies.
}

\dsquestion{What data does each instance consist of? “Raw” data (e.g., unprocessed text or images) or features?}

\dsanswer{
Each instance consists of the start and end frames and time stamps of each segment, along with the dialog transcript from the subtitle file, and annotator ratings of level of objectification and thesaurus concepts. The text of the subtitle was down-cased and HTML tags were removed.
}

\dsquestion{Is there a label or target associated with each instance?}

\dsanswer{
Each instance is a segment associated with an objectification rating (Easy Negative - EN, Hard Negative - HN, Sure - S, Not Sure - NS) and a set of concepts selected by the annotator as present, and chosen from the thesaurus. An additional free text box may be used by the annotator (column 'comment' in Fig. \ref{fig:datasheet_instance_example}).
}

\dsquestion{Is any information missing from individual instances?}

\dsanswer{
The visual and sound data are not provided in the dataset, but exact frame and time stamping are provided to correctly use the annotations of the MObyGaze dataset and compare to the benchmarked models accompanying the dataset. 
}

\dsquestion{Are relationships between individual instances made explicit (e.g., users’ movie ratings, social network links)?}

\dsanswer{
Any relationship between annotated segments is assured by the consistency in movie identifiers, frame and time stamping as well as by annotator identifiers.
}

\dsquestion{Are there recommended data splits (e.g., training, development/validation, testing)?}

\dsanswer{
In order to assess generic objectification patterns while maximizing the amount of data available for training and testing, we recommend to use leave-4-movies-out cross-validation, where no movie in train is used in test (even for different segments). The dataset therefore comes with 5 different folds.
Each fold is made of a train, validation and test split, each composed of 14, 2 and 4 movies, respectively. There is no overlap between the test sets, so that every movie appears exactly once in test.
The fold indices where each movie appears in test or validation are indicated in Table \ref{table:datasheet_films}. We recommend to use the same folds to generate results comparable with the models benchmarked in the article introducing MObyGaze (present article submitted to NeurIPS dataset and benchmarks track, to be edited after review). Results should be reported as average of model results over 5 folds, along with standard deviations.
The definition of the ML task on the MObyGaze dataset must be fully specified (classification, localization, constitution of the positive and negative classes from the objectification ratings and tagged concepts).
}

\dsquestion{Are there any errors, sources of noise, or redundancies in the dataset?}

\dsanswer{
N/A
}

\dsquestion{Is the dataset self-contained, or does it link to or otherwise rely on external resources (e.g., websites, tweets, other datasets)?}

\dsanswer{
The dataset only relies on the films, not shared for itellectual property reasons. Table \ref{table:datasheet_films} and file mobygaze$\_$dataframe.csv provide the necessary information for anyone to acquire the films and align the MObyGaze annotations onto it. We integrate the subtitles in the dataset.
}

\dsquestion{Does the dataset contain data that might be considered confidential?}

\dsanswer{N/A
}

\dsquestion{Does the dataset contain data that, if viewed directly, might be offensive, insulting, threatening, or might otherwise cause anxiety?}

\dsanswer{
The annotated movies can contain offensive and otherwise disturbing content. Detailed information on the movie age suitability rating per country is available on the IMDB page under parental guide \footnote{www.imdb.com/title/[IMDBkey]/parentalguide}
This page also lists the types of scenes under each category on non-mild content.
The free text field of the annotations provided in MObyGaze may contain speech segments or descriptions of visual content linked to the labels that can be offensive.
}

\dsquestion{Does the dataset relate to people?}

\dsanswer{All annotated films are fictitious, and do not depict real persons. The characters are played by human actors. The dataset is meant to train models to study disparities in gender representation in cinema.
}

\dsquestion{Does the dataset identify any subpopulations (e.g., by age, gender)?}

\dsanswer{The MObyGaze dataset does not provide annotation of gender or other demographics of the characters. However, character gender can be obtained from the MovieGraphs dataset for the corresponding movies, or easily inferred from the actor cast and face recognition technologies applied to the content to detect the actor.
}

\dsquestion{Is it possible to identify individuals (i.e., one or more natural persons), either directly or indirectly (i.e., in combination with other data) from the dataset?}

\dsanswer{The actors can be identified from their appearance in the movies, but the MObyGaze dataset is not the enabler.
}

\dsquestion{Does the dataset contain data that might be considered sensitive in any way?}

\dsanswer{N/A
}

\bigskip

\dssectionheader{Collection Process}

\dsquestion{How was the data associated with each instance acquired?}

\dsanswer{Each movie is annotated by 2 experts (with background in computer science, film studies and cognitive psychology), 
who watch it entirely, setting temporal boundaries of each segment where at least one objectifying concept is deemed present. For each such segment, they rate objectification on one of four levels:\\
\noindent$\bullet$ Easy Negative (EN): no objectifying concept is present;\\
\noindent$\bullet$ Hard Negative (HN): one or some concepts are present, are annotated, but are deemed insufficient to produce a perception of objectification;\\
\noindent$\bullet$ Sure (S): objectification is perceived and explained by the annotated concepts from the thesaurus;\\
\noindent$\bullet$ Not Sure (NS): objectification is perceived and concepts are annotated but the annotator considers they do not sufficiently explain the perception of objectification.\\
}

\dsquestion{What mechanisms or procedures were used to collect the data (e.g., hardware apparatus or sensor, manual human curation, software program, software API)?}

\dsanswer{Fig. \ref{fig:tagging_tool} shows the tool specifically designed for densely annotating objectification levels and concepts. It can be seen that the tool provides a free-text field that the annotators can choose to use.
The annotators first annotated 2 movies. The obtained annotations were then aligned and colored for the annotators to identify their major divergences. They convened and identified that the agreement was generally high on the rating of objectification. 
Analyzing the annotation differences for the 11 concepts, the annotators specifically identified under-determination of concepts Actitivies and Appearance. They expanded Activities to include all types of actions contributing to objectification, particularly momentary actions by a character onto another (including aspects of domination and violence). They trimmed the concept of Appearance, which initially consisted of instances deploying over the entire film (such as age of character not matching age or appearance of actress) to restrict it to scene-level features. Fig. \ref{fig_thes} shows the resulting thesaurus. They then carried out individually the annotation over the rest of the 20 movies.
}

\dsquestion{If the dataset is a sample from a larger set, what was the sampling strategy (e.g., deterministic, probabilistic with specific sampling probabilities)?}

\dsanswer{The dataset contains all the instances produced by both annotators having entirely annotated each movie.
The 20 movies of MObyGaze are a subset of the 51 movies of the pre-existing MovieGraphs dataset. The 20 movies have been chosen to maintain the ratio of represented genres and diversify to the maximum the actors/directors represented.
Table \ref{table:datasheet_films} provides details on the movies.
}

\dsquestion{Who was involved in the data collection process (e.g., students, crowdworkers, contractors) and how were they compensated (e.g., how much were crowdworkers paid)?}

\dsanswer{
Both experts that produced all the annotations are project members (tenured scholars) and worked on annotation as part of their research tasks.
}

\dsquestion{Over what timeframe was the data collected? Does this timeframe match the creation timeframe of the data associated with the instances (e.g., recent crawl of old news articles)?}

\dsanswer{The annotations took place between May 2023 and May 2024. The annotated films were produced between 1989 and 2014.
}

\dsquestion{Were any ethical review processes conducted (e.g., by an institutional review board)?}

\dsanswer{The data collection neither involved intervention nor interpersonal contact with subjects, or collection of data on subjects. The dataset creation therefore did not require an institutional review board or ethical committee review.
}

\dsquestion{Does the dataset relate to people?}

\dsanswer{The annotations provided do not relate to people. The annotated films depict fictitious characters.
}

\dsquestion{Did you collect the data from the individuals in question directly, or obtain it via third parties or other sources (e.g., websites)?}

\dsanswer{The data has been collected directly by project members using the annotation tool on their local computers.
}

\dsquestion{Were the individuals in question notified about the data collection?}

\dsanswer{N/A
}

\dsquestion{Did the individuals in question consent to the collection and use of their data?}

\dsanswer{The annotators were project members. The film data have been lawfully used as per the French and European regulations.
}

\dsquestion{If consent was obtained, were the consenting individuals provided with a mechanism to revoke their consent in the future or for certain uses?}

\dsanswer{N/A
}

\dsquestion{Has an analysis of the potential impact of the dataset and its use on data subjects (e.g., a data protection impact analysis) been conducted?}

\dsanswer{N/A
}

\bigskip
\dssectionheader{Preprocessing/cleaning/labeling}

\dsquestion{Was any preprocessing/cleaning/labeling of the data done (e.g., discretization or bucketing, tokenization, part-of-speech tagging, SIFT feature extraction, removal of instances, processing of missing values)?}

\dsanswer{
The annotations created and provided in MObyGaze are the raw transcription from the annotation tool, which generates, for each annotator annotating a movie, json files sharing indices of start and end frame of each segment, objectification level, objectification concepts, and free text. These are re-formatted without any approximation in the mobygaze$\_$dataframe.csv provided in the dataset.
}

\dsquestion{Was the “raw” data saved in addition to the preprocessed/cleaned/labeled data (e.g., to support unanticipated future uses)?}

\dsanswer{
The raw data is saved but does not bring additional information compared to mobygaze$\_$dataframe.csv and is not shared to preserve anonymity. Indeed, the json files produced by the annotation tool contain folder paths of the local machines of the annotators.
}

\dsquestion{Is the software used to preprocess/clean/label the instances available?}

\dsanswer{
Yes, the python scripts used to create the database from the json files generated by the annotation tool, and the python scripts used to generate mobygaze$\_$dataframe.csv from the database, will be made available along with the publication of the annotation tool to the community.
}

\bigskip
\dssectionheader{Uses}

\dsquestion{Has the dataset been used for any tasks already?}

\dsanswer{
The dataset has been used for classification of objectification knowing the true segment boundaries, and localization of objectififcation in fixed-length segments. All vision, speech and audio modalities have been used separately.
}

\dsquestion{Is there a repository that links to any or all papers or systems that use the dataset?}

\dsanswer{
\url{https://github.com/husky-helen/MObyGaze}
}

\dsquestion{What (other) tasks could the dataset be used for?}

\dsanswer{
The MObyGaze dataset is also meant to design explainable models to better characterize complex temporal and multimodal objectification patterns, which can in turn enrich qualitative studies by media scholars.
The MObyGaze dataset can also be used to study the fairness of existing computer vision models: person detectors and human pose estimators may miss the presence of characters onscreen, compromising the study of how certain patterns correlate with certain human groups, if the humans are often mis-detected for these patterns (e.g., shots with headless body parts).

}

\dsquestion{Is there anything about the composition of the dataset or the way it was collected and preprocessed/cleaned/labeled that might impact future uses?}{}

\dsanswer{Any future user must be aware that the movies selected for annotating objectification were in no case chosen for their specific crew or other production affiliation, but on the sole basis of preserving the genre distribution when sampling from the pre-existing MovieGraphs dataset.}

\dsquestionex{Are there tasks for which the dataset should not be used?}{}
\dsanswer{The MObyGaze dataset should not be used for tasks such as content filtering, defining regulatory standards, or censorship of media content. Another potential negative impact would lie in the interpretation of the outputs of models, being it by social science researchers assisting their qualitative studies with models trained on MObyGaze to detect objectification, or by amateurs. For example, if the models to detect objectification trained on filmic fictional content are used on real-user videos (such as those uploaded on online social networks), one may (1) establish and highlight correlations between the level of objectification and demographics of users that put on a performance and upload their videos, then (2) claim that the users of these demographic groups (most likely historically under-privileged or submitted, e.g., women) are the sole responsible for choosing to objectify themselves, somehow discarding and making invisible the social root causes leading to self-objectification. While the risk of fallacious interpretation already exists from qualitative analyses not assisted by AI models, a misleading narrative can be reinforced with an AI model allowing larger-scale analysis and a veneer of objectivity.The first proactive measure is the choice of the dataset license prohibiting commercial usage, to address the risk of censorship mentioned first. The second proactive measure is to highlight in the dataset documentation that the models trained on the MObyGaze dataset must be used in a clear ethical framework: that of making subtle patterns of representation disparities visible and contribute responsible usages. These may be discussions on the extent of these patterns in certain contexts and their causes (discussions in social sciences, or by associations raising awareness on representation disparities in the media). This can also be for training purpose, e.g., to assist students in audiovisual production in the analysis of their production trials and develop reflexivity in their practice. We will also advertise this message in the public website under construction where we will demo the dataset and possible models. We will indicate that every further developed model should come with proper standardized documentation for this task, relaying the need for responsible interpretation and usage.}

\bigskip
\dssectionheader{Distribution}

\dsquestion{Will the dataset be distributed to third parties outside of the entity (e.g., company, institution, organization) on behalf of which the dataset was created?}

\dsanswer{Yes, the dataset will be made publicly available.
}

\dsquestion{How will the dataset will be distributed (e.g., tarball on website, API, GitHub)}

\dsanswer{The dataset is made available on Github: \url{https://github.com/husky-helen/MObyGaze}], and will also be made available on Zenodo, from where a DOI will be obtained, and long-term storage ensured.
}

\dsquestion{When will the dataset be distributed?}

\dsanswer{The dataset is already available online, but will be hosted on Zenodo and attributed a DOI after the double-blind review process.
}

\dsquestion{Will the dataset be distributed under a copyright or other intellectual property (IP) license, and/or under applicable terms of use (ToU)?}

\dsanswer{The dataset will be made available under an open CC BY-NC-SA license.
}

\dsquestion{Have any third parties imposed IP-based or other restrictions on the data associated with the instances?}

\dsanswer{N/A
}

\dsquestion{Do any export controls or other regulatory restrictions apply to the dataset or to individual instances?}

\dsanswer{N/A
}

\bigskip
\dssectionheader{Maintenance}

\dsquestion{Who will be supporting/hosting/maintaining the dataset?}

\dsanswer{
The dataset will be hosted on permanent public storage Zenodo. The dataset will be supported and maintained by the project team. The team is led by tenured researchers who will dedicate the necessary resources, after project funding ends, to maintain the dataset.
}

\dsquestion{How can the owner/curator/manager of the dataset be contacted (e.g., email address)?}

\dsanswer{By institutional email lucile.sassatelli@univ-cotedazur.fr
}

\dsquestion{Is there an erratum?}

\dsanswer{
N/A
}

\dsquestion{Will the dataset be updated (e.g., to correct labeling errors, add new instances, delete instances)?}

\dsanswer{New versions will be added to the dataset in the case of inclusion of a completely new session of annotation with modifications to the thesaurus, film set, and/or annotators. The dataset will be updated in the case of adding individual annotations on new or existing films using the same thesaurus. Regular updates (every 3-6 months) to address minor issues in the dataset will be provided based on requests. Next version previewed is in Dec 2026 from another annotation session on television series and historical drama.
By 2027, we intend to make the thesaurus evolve to annotate tropes both at the film and at the segment levels, creating a new version of the dataset.
}

\dsquestion{If the dataset relates to people, are there applicable limits on the retention of the data associated with the instances (e.g., were individuals in question told that their data would be retained for a fixed period of time and then deleted)?}

\dsanswer{N/A
}

\dsquestion{Will older versions of the dataset continue to be supported/hosted/maintained?}

\dsanswer{The older version of the dataset will continue to be hosted and maintained, thanks to the resources described above where tenured researchers responsible for the research funding will dedicate the necessary resources to maintenance.
}

\dsquestion{If others want to extend/augment/build on/contribute to the dataset, is there a mechanism for them to do so?}

\dsanswer{
Under the terms of the chosen license CC BY-NC-SA, any user can fork the dataset and extend, modify, and share it as desired.
}

\end{multicols}

\newpage


\subsection{Details on the MObyGaze dataset}\label{sec:suppl_dataset}

\subsubsection{Films}

Table \ref{table:films} shows the list of films in the MObyGaze dataset, while Table \ref{table:genre_distr} shows how this list reproduces the genre distribution of the original MovieGraphs dataset \cite{vicol_moviegraphs_2018}.

\begin{table*}[ht]
\small
\caption{List of films annotated in the MObyGaze dataset (sorted by genre)}
\label{table:films} 
{\scriptsize
\begin{tabular}{lllllll}
\toprule
IMDB key & Movie Title & Duration & Year & Genre & Test Fold & \makecell{Validation\\Fold}\\
\cmidrule(lr){1-7}
tt0097576 & \makecell{Indiana Jones and the \\Last Crusade} & 2h 6min & 1989 & adventure, action & 1 & 5\\
tt1454029 & The Help & 2h 26min & 2011 & drama & 2 & \\
tt1285016 & The Social Network & 2h 0min & 2010 & drama, biographical & 3 & 4\\
tt0467406 & Juno & 1h 36min & 2007 & drama, comedy & 4 & \\
tt0110912 & Pulp Fiction & 2h 34min & 1994 & drama, crime & 5 & 3\\
tt0822832 & Marley \& Me & 1h 55min & 2008 & drama, family & 1 & \\
tt1568346 & \makecell{The Girl with the \\Dragon Tattoo} & 2h 38min & 2011 & drama, mystery, crime & 2 & \\
tt2267998 & Gone Girl & 2h 29min & 2014 & drama, mystery, thriller & 3 & 2\\
tt0109830 & Forrest Gump & 2h 22min & 1994 & drama, romantic & 4 & 1\\
tt0120338 & Titanic & 3h 14min & 1997 & drama, romantic & 5 & \\
tt0108160 & Sleepless in Seattle & 1h 45min & 1993 & drama, romantic, comedy & 1 & 5\\
tt0119822 & As Good as It Gets & 2h 18min & 1997 & drama, romantic, comedy & 2 & \\
tt1193138 & Up in the Air & 1h 49min & 2009 & drama, romantic, comedy & 3 & 4\\
tt1570728 & Crazy,Stupid,Love. & 1h 58min & 2011 & drama, romantic, comedy & 4\\ 
tt1045658 & Silver Linings Playbook & 2h 2min & 2012 & drama, romantic, comedy & 5 & 3\\
tt0970416 & \makecell{The Day the Earth \\Stood Still} & 1h 43min & 2008 & drama, sci-fi, adventure & 1 & \\
tt1907668 & Flight & 2h 18min & 2012 & drama, thriller & 2 & \\
tt0375679 & Crash & 1h 55min & 2004 & drama, thriller, crime & 3 & 2\\
tt1142988 & The Ugly Truth & 1h 35min & 2009 & romantic, comedy & 4 & 1\\
tt1632708 & Friends with benefits & 1h 49min & 2011 & romantic, comedy & 5 & \\
\bottomrule
\end{tabular}
}
\end{table*}

\begin{table*}[ht]
\small
\centering
\caption{Distribution of movie genres between MovieGraphs (51 movies) \cite{vicol_moviegraphs_2018} and MObyGaze (subset of 20 movies). Genre source: IMDB (note: a movie has several genres).}
\label{table:genre_distr} 
{\footnotesize
\begin{tabular}{lll}
\toprule
Genre & MoviGraphs & MObyGaze \\
\cmidrule(lr){1-3}
Action & 0.02 & 0.05\\
Adventure & 0.08 & 0.1\\
Biography & 0.06 & 0.05\\
Comedy & 0.43 & 0.4\\
Crime & 0.2 & 0.15\\
Drama & 0.76 & 0.85\\
Family & 0.04 & 0.05\\
Fantasy & 0.02 & 0\\
Film noir & 0.02 & 0\\
History & 0.02 & 0\\
Mystery & 0.12 & 0.1\\
Romance & 0.49 & 0.45\\
Sci-Fi & 0.08 & 0.05\\
Thriller & 0.16 & 0.1\\
\bottomrule
\end{tabular}
}
\end{table*}

\subsubsection{Annotation tool}
\begin{figure*}[ht]
  \centering
  \includegraphics[width=0.9\linewidth]{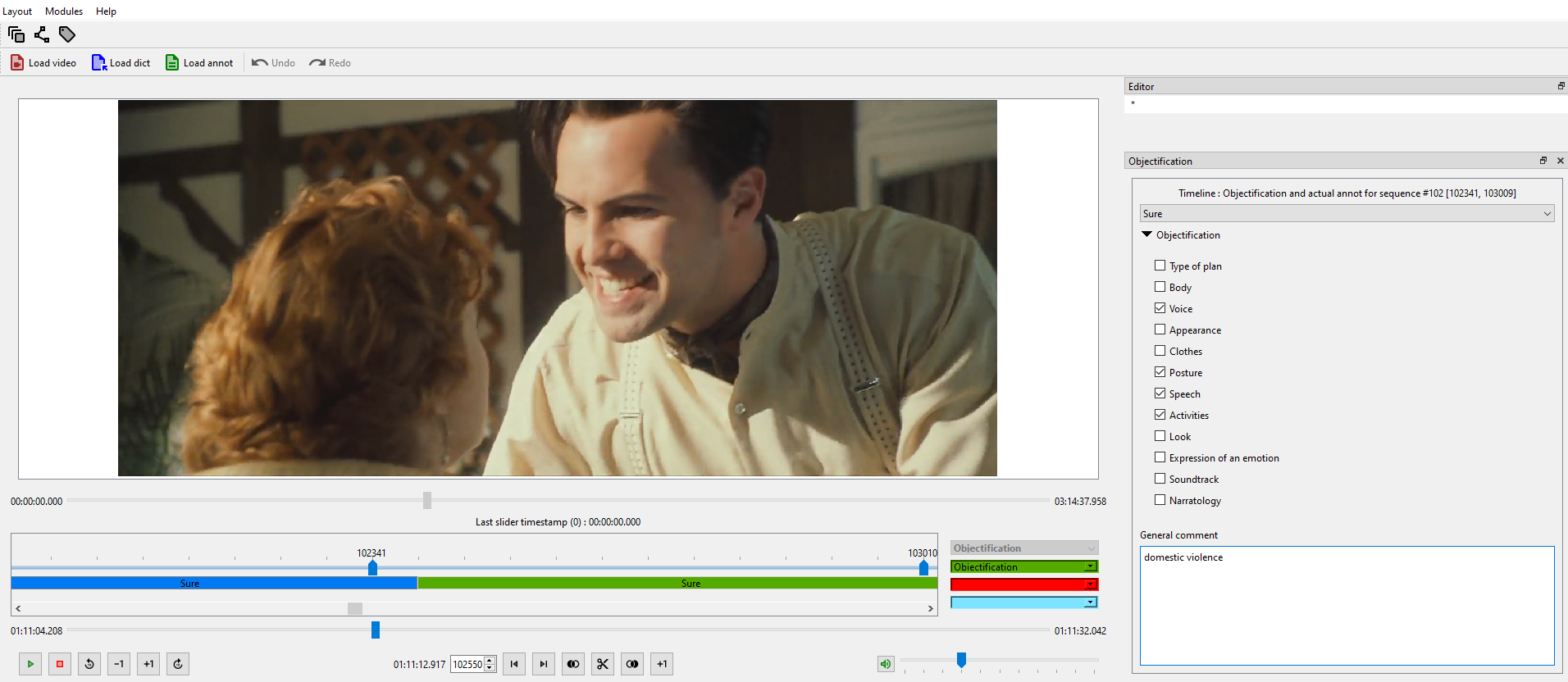}
  \caption{The annotation interface.}
  \label{fig:tagging_tool}
\end{figure*}

Fig. \ref{fig:tagging_tool} shows the tool specifically designed for densely annotating objectification levels and concepts. It can be seen that the tool allows for a free text field that the annotators are free to use.

\subsubsection{Details on the annotation procedure}
\revised{The annotators are experts who are both tenured scholars at public institutions and are the instigators of the nationally-funded project funding this work through the involvement of 3 laboratories in social sciences and 3 laboratories in computer science. They have coordinated the multidisciplinary team for every phase of the work for this article. One annotator holds a B.A. in English literature, a M.Sc in Communications and a PhD in computer science with expertise in cinematic discourse, the other holds a PhD in computer science, has led projects at the intersection of CS and social psychology for gender inequalities, and is the project PI. They both identify as women, in their late 30s-early40s, they both work in Europe, one is from Asian descent and the other from European descent.}

The annotators first annotated 2 movies. The obtained annotations were then aligned and colored for the annotators to identify their major divergences. They convened and identified that the agreement was generally high on the rating of objectification. Noticeable differences were in annotation of NS with levels of narratology spanning over more than a single segment. The thesaurus is indeed targeted at annotating sailient concepts in segments, and we discuss this limitation in Sec. \ref{sec:lims_appls}.
They specifically identified under-determination of concepts Actitivies and Appearance. They expanded Activities to include all types of actions contributing to objectification, particularly momentary actions by a character onto another (including aspects of domination and violence). They trimmed the concept of Appearance, which initially consisted of instances deploying over the entire film (such as age of character not matching age or appearance of actress) to restrict it to scene-level features. Fig. \ref{fig_thes} shows the resulting thesaurus. They then carried out individually the annotation over the rest of the 20 movies.

\begin{figure*}[t]
  \centering
  \includegraphics[width=0.9\linewidth]{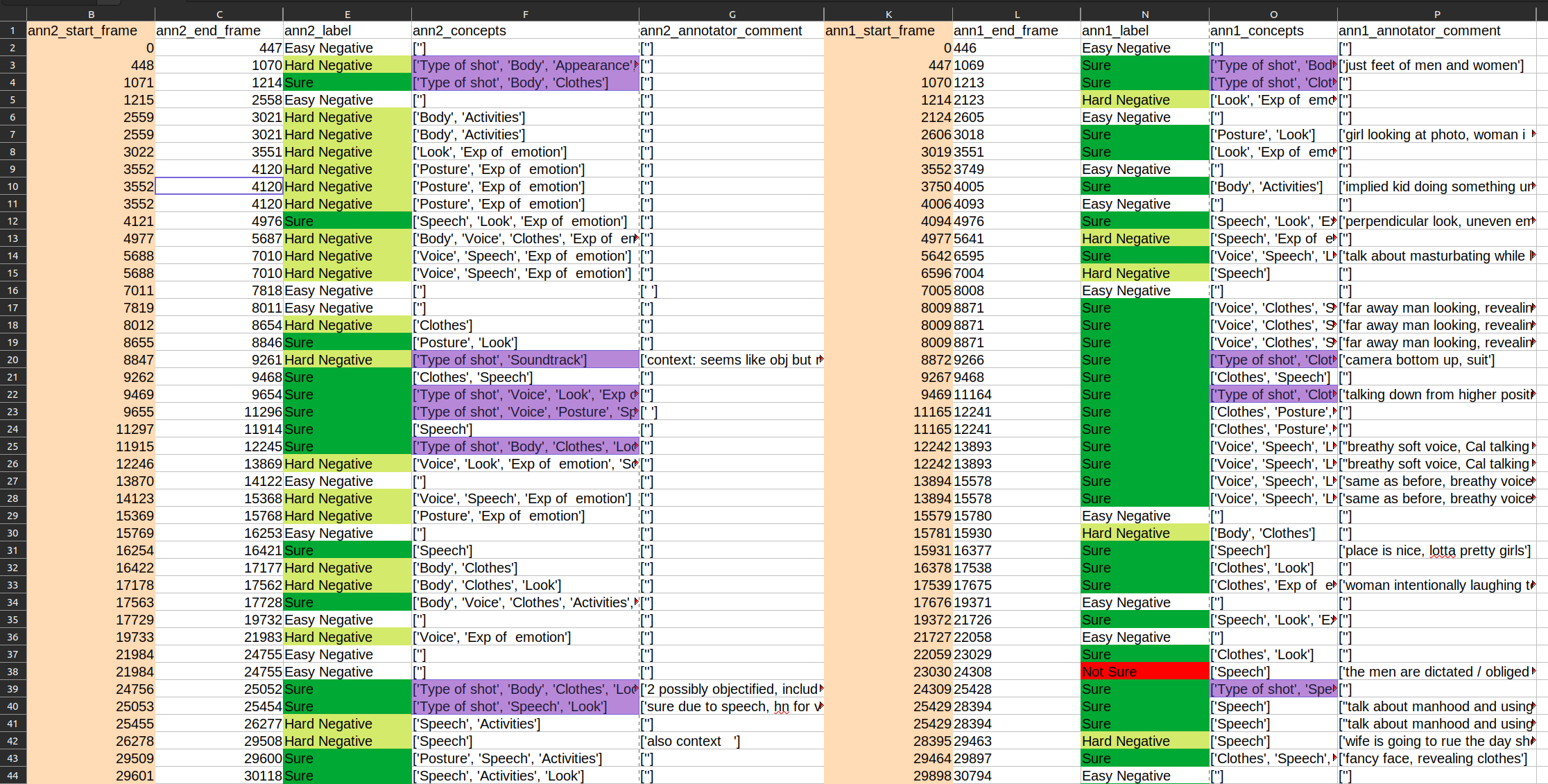}
  \caption{\revised{Excerpt of an alignment file created after the pilot annotation by both experts so support the discussion leading to invalidate or modify certain concepts or aligning their interpretations.}}
  \label{fig:example_alignment}
\end{figure*}

\subsubsection{Alternate representation of the thesaurus}\label{app:thes_colorblind}
\begin{figure*}[t]
  \centering
  \includegraphics[width=0.9\linewidth]{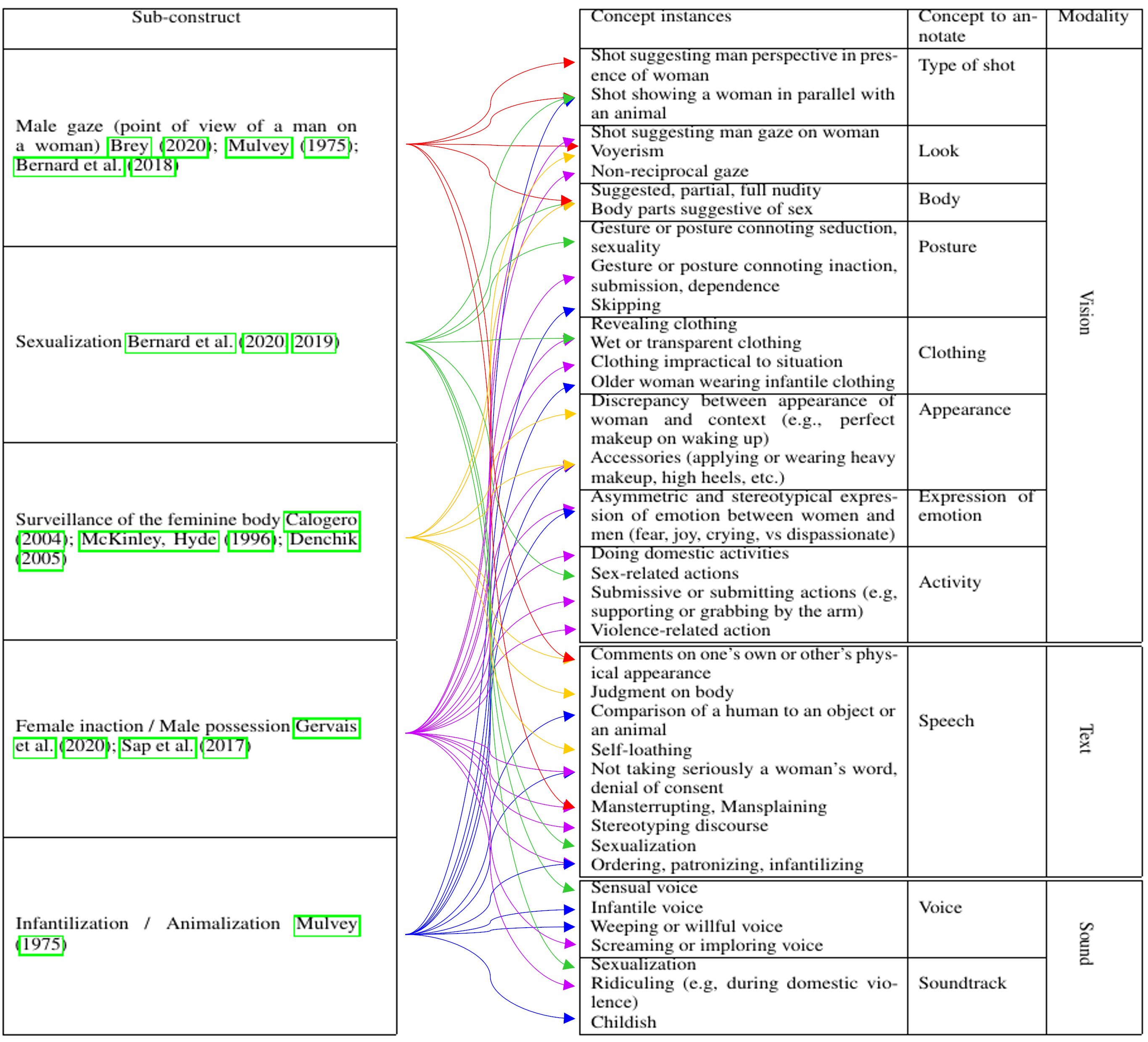}
  \caption{Thesaurus for the construct of objectification: 5 sub-constructs (left table) manifested through 11 concepts spanning 3 modalities (right table).}
  \label{fig:thes_cb}
\end{figure*}

A color-blind version of the thesaurus is shown in Fig. \ref{fig:thes_cb}.

\subsubsection{Examples of annotations}

\begin{figure*}[]
  \centering
  \includegraphics[width=0.9\linewidth]{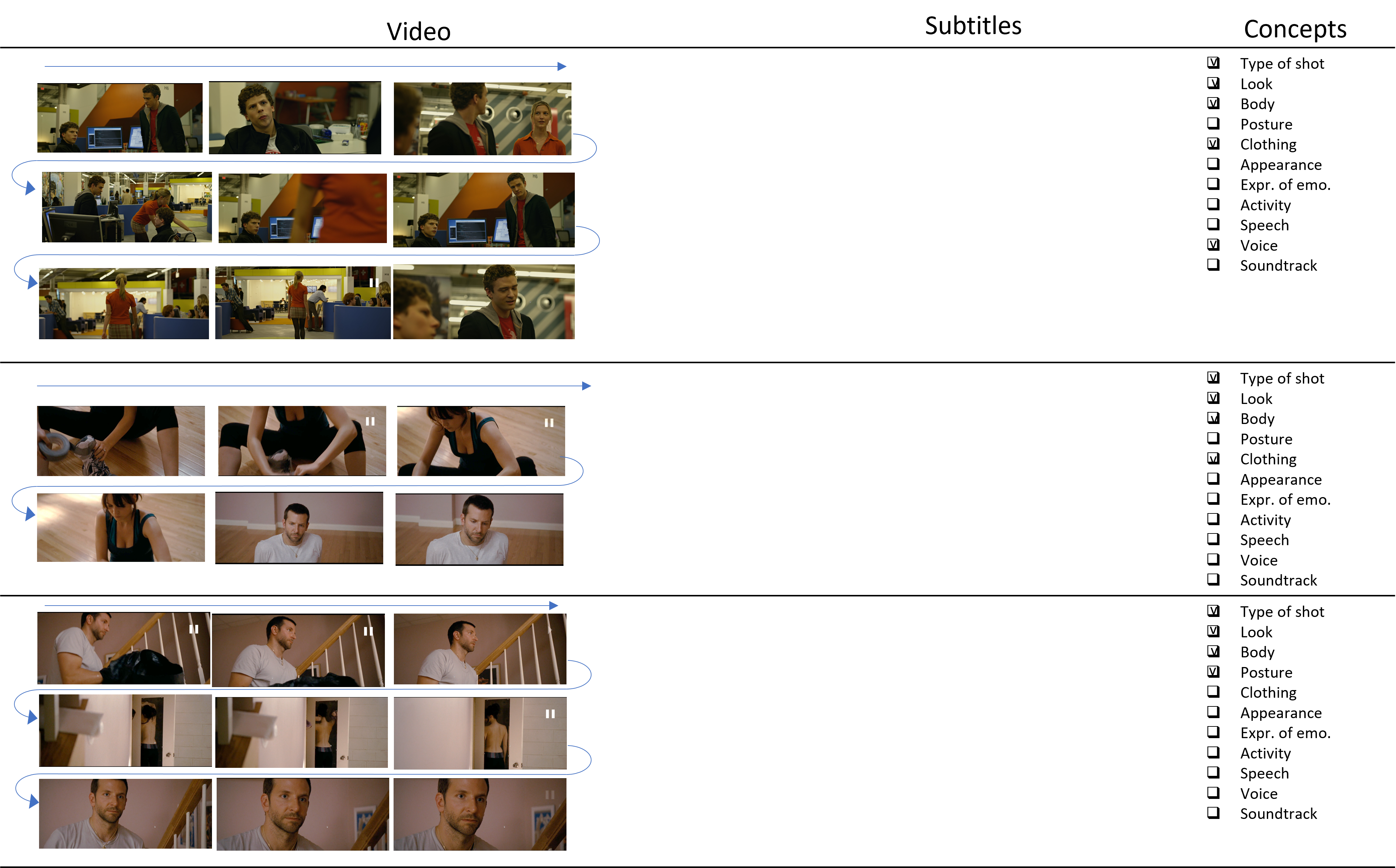}
  \caption{Examples of segments delimited and tagged with a Sure level of objectification, produced by only or mainly visual concepts.}
  \label{fig:examples_visual}
\end{figure*}

\begin{figure*}[]
  \centering
  \includegraphics[width=0.9\linewidth]{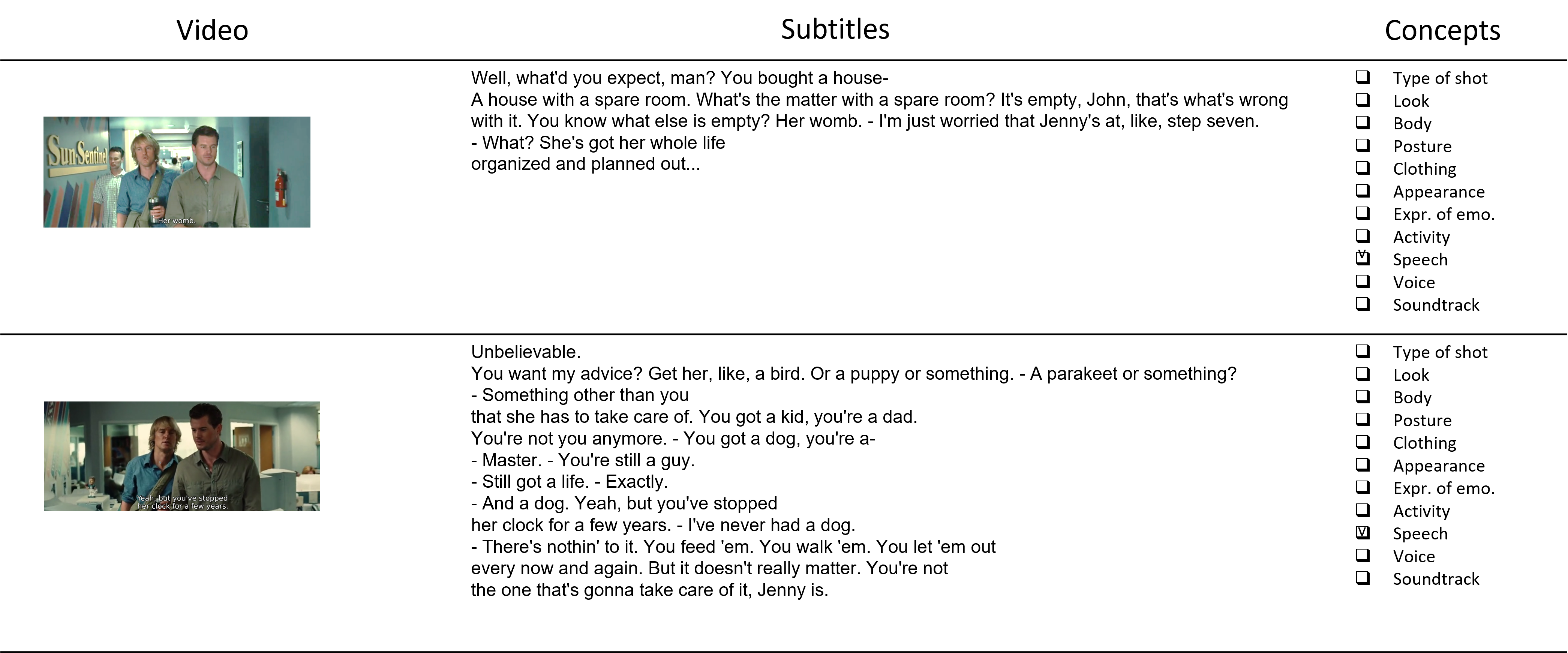}
  \caption{Examples of segments delimited and tagged with a Sure level of objectification, produced by only  textual concept.}
  \label{fig:examples_textual}
\end{figure*}

\begin{figure*}[]
  \centering
  \includegraphics[width=0.9\linewidth]{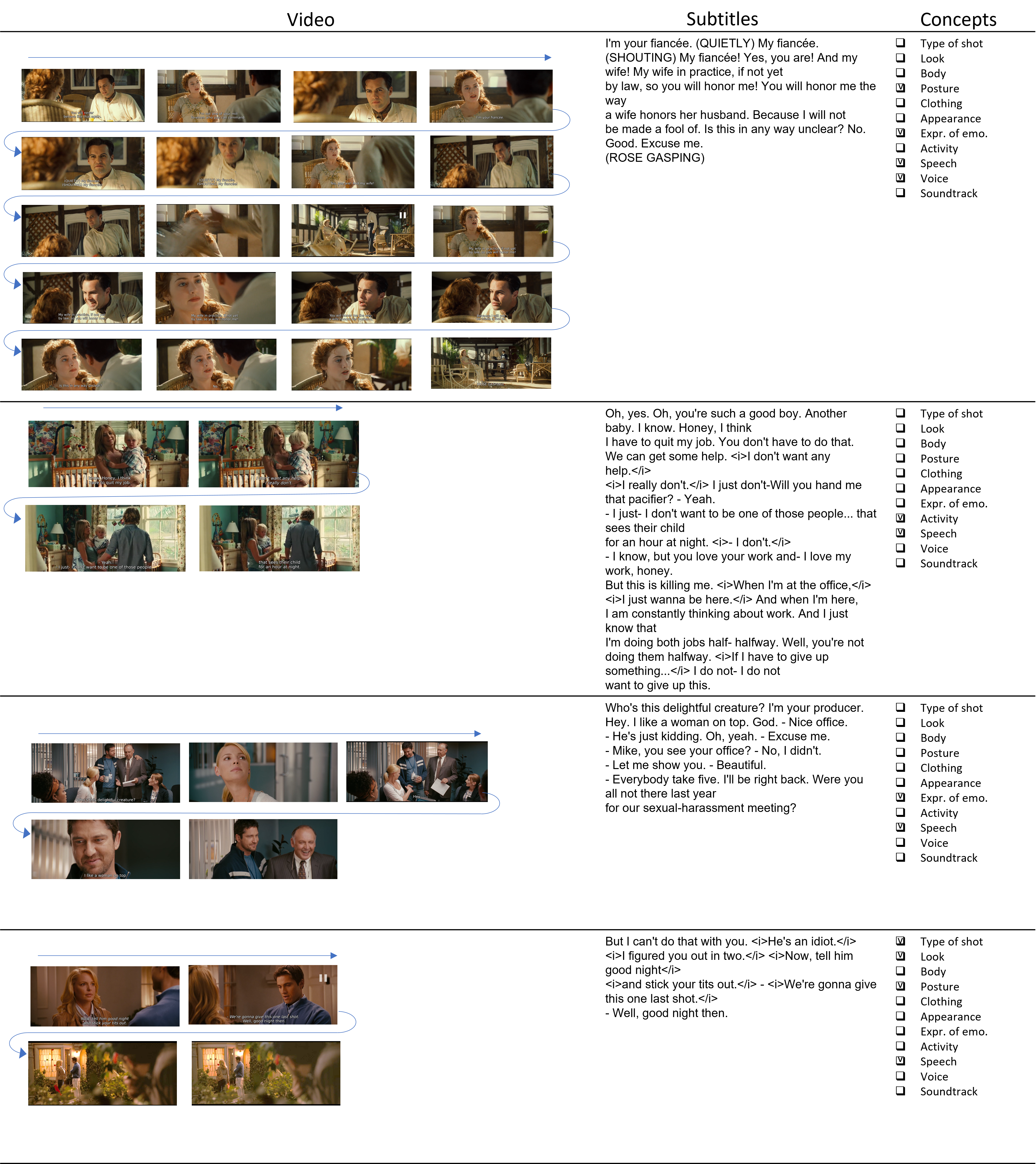}
  \caption{Examples of segments delimited and tagged with a Sure level of objectification, produced by a combination of concepts of different modalities.}
  \label{fig:examples_multimodal}
\end{figure*}

Examples of annotated segments are detailed in the case where objectification is produced by visual concepts mainly in Fig. \ref{fig:examples_visual}, by textual concept only in Fig. \ref{fig:examples_textual}, and by a multimodal combination of visual, textual and audio concepts in Fig. \ref{fig:examples_multimodal}.

\subsubsection{Details on inter-annotator agreement analysis}\label{app:IAA}

We provide the code for computing IAA metrics in the Github repository.

\paragraph{Definition of dist@IoUthresh}
\revised{As expected distances are used for chance correction, we take them as distances between any annotations (regardless of annotator) between 2 different movies. Between 2 movie annotations $\mathbf{a}$ and $\mathbf{b}$, we defined the distance itself as follows. Considering $\mathbf{a}$ as GT, for each segment $s_a\in\mathbf{a}$, we select the subset $S_b$ of segments in $\mathbf{b}$ so that IoU$\geq$IoUthresh. For each such segment $s_b$, we set dist($s_a,s_b$)=$D_0$ if the objectification labels are the same, $=D_1$ if the labels are one hop apart (we set the order EN - HN- S), $=D_2$ if 2 hops apart. We discard all NS segments that represent less than 5\%. We then take dist$(s_a,S_b)$ as the minimum of dist($s_a,s_b$) for all $s_b\in S_b$, then dist($S_a,S_b$) as the average of dist$(s_a,S_b)$ for all $s_a\in S_a$, and finally $D(S_a,S_b)=(dist(S_a,S_b)+dist(S_b,S_a))/2$. $D_0$, $D_1$ and $D_2$ are hyper-parameters we set to 0, 1 and 2, respectively. We verify their values do not impact the IAA metrics provided the order remains the same (as it can be understood from the following explanation).}

\paragraph{Formal intuition for the difference between Krippendorff's $\alpha$ and $\sigma$ with unitizing}
\revised{Let us consider that each segment label can take $V=2$ values, instead of $V=3$ (EN, HN, S, and we discard distance value $D_2$). Let us assume for now that every segment $s_a$ of an annotation sequence can be matched with exactly $K=1$ other annotation segment $s_b$. Let us assume that the annotators agree on a fraction $F_A$ of the samples, and chance correction is made by matching with another random segment with a label picked uniformly at random. Then Pr$(s_a\neq s_b)=1-F_A$ if $s_a$ and $s_b$ are from the same item, otherwise Pr$(s_a\neq s_b)=V(1/V)(1-V)$. Taking $V=2$ and $F_A=0.8$ yields $\alpha = 1-$ Pr$(s_a\neq s_b$|same item) $D_1$/( Pr$(s_a\neq s_b$|different items) $D_1$) = 1-0.2/0.5 = 0.6 as expected.
Let us now consider a non-perfect agreement on the segment boundaries, and that under a certain IoUthresh constraint, we have exactly $K=2$ segments $s^1_b$ and $s^2_b$ that can be matched with $s_a$ for different items, while this is possible only with probability $f$ for same items (to consider a non-trivial random agreement on boundaries for a given movie). Then we can express:}
\revised{Pr($s_a\neq s^1_b$ and $s_a\neq s^2_b$|same item) = $(1-f)[V(1/V(1-F_A))]+f[V(1/V(1-F_A)(1-1/V))]$ 
= $(1-F_A)(1-1/V f)$,}\\
\revised{and Pr($s_a\neq s^1_b$ and $s_a\neq s^2_b$|different items) = $V(1/V(1-1/V)^K) = (1-1/V)^K$}\\
\revised{Taking $V=2$ and $K=2$:}\\
\revised{$\alpha = 1- (1-F_A)/0.5 (2-f)$}\\
\revised{so for $f=1$: $\alpha = 0.6$ again}\\
\revised{but for $f<1$, e.g., $f=0.7$: $\alpha = 0.22$.}

\revised{This explains the partly opposite trends of $\alpha$ and $\sigma$ we observe in Fig. \ref{fig:iaa_obj} (remind that there in the real case V=3): increasing IoUthresh yields a monotonously decreasing $\alpha$ while $\sigma$ stays constant or slightly increases, before decreasing sharply after some high IoUthresh (above the boundary agreement rate), as expected.
Therefore, as an IAA metric should describe how much the overall data is better than chance, after observing the weakness of Kripendorff's $\alpha$@IoU to describe this phenomenon, we opt for reporting $\sigma$ instead.}

\subsubsection{\revised{Details on computation for external consistency verification}}\label{app:external_cons}
\revised{For visual concepts, we do contrastive scoring precisely as follows:}\\
\revised{1. Cosine similarity is computed between every window embedding in the segment and every prompt embedding for that concept, giving a *(windows × prompts)* similarity matrix.}\\
\revised{2. This is aggregated across windows — either by the mean, or by a chosen quantile (e.g., the 90th percentile, capturing "peak" rather than "average" presence across the segment) — giving one similarity value per prompt.}\\
\revised{3. This is then aggregated with max-pooling across the concept's multiple prompts, giving a single raw concept score for the segment.}\\
\revised{4. CLIP-family text embeddings are known to be anisotropic. We correct this with embedding centering and similarity offset. First for embedding centering, a single "background" vector is computed as the mean of all prompt embeddings (every concept's prompts and the negative prompts, pooled together), and this background is subtracted from every individual prompt embedding before re-normalizing to unit length. The same idea is applied on the video side: each movie's window embeddings are centered by that movie's own average window embedding (removing shared, movie-specific bias such as color grading or cinematographic style) before re-normalizing. Second for similarity offset, the identical two-step aggregation above is applied to the negative/background prompts, giving a single background score for the segment.}\\
\revised{5. The final concept score = raw concept score - background score.} \\

\begin{figure*}[]
  \centering
  \includegraphics[width=0.9\linewidth]{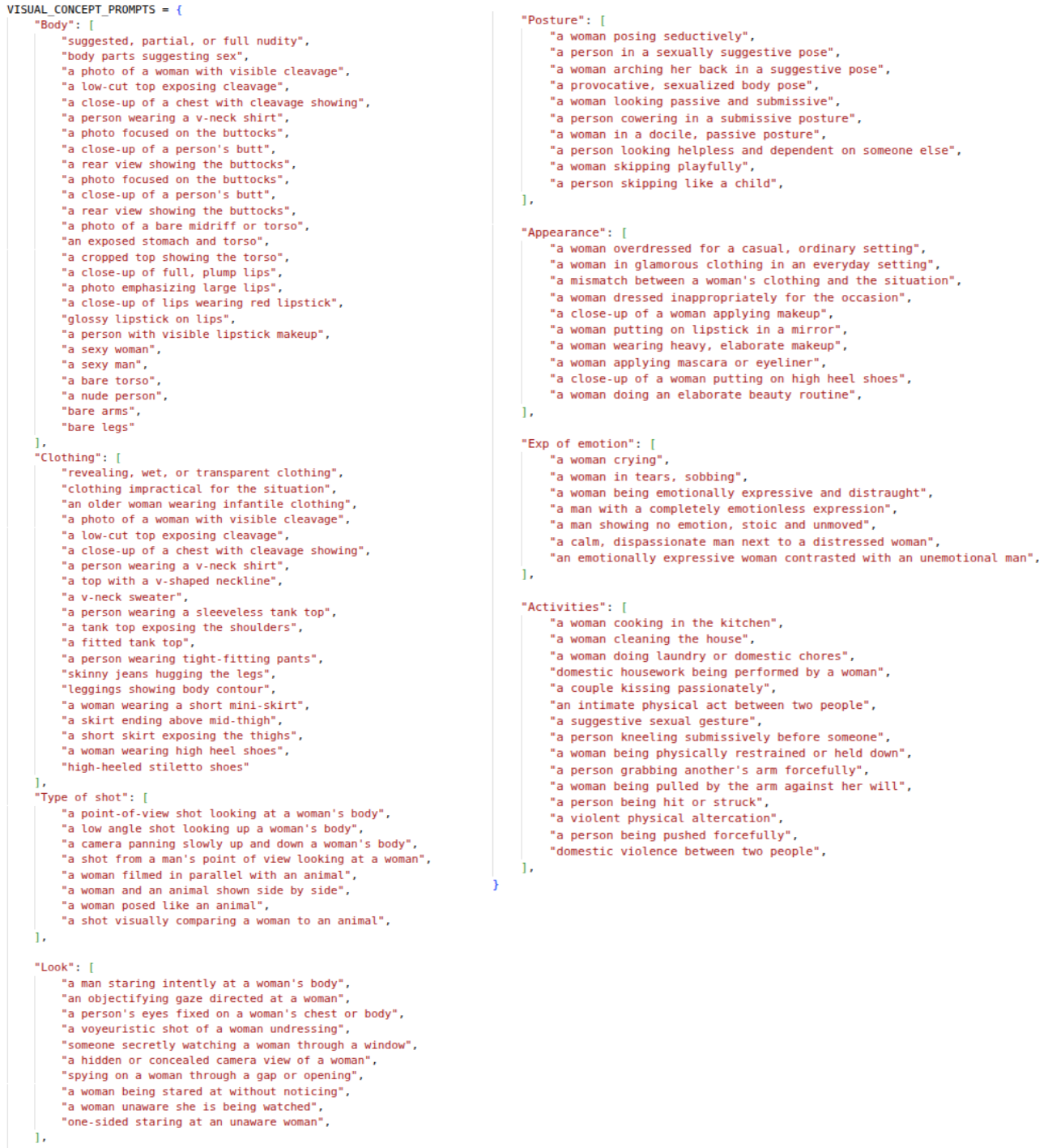}
  \caption{\revised{The text prompt used for the X-CLIP similarities.}}
  \label{fig:vis_concept_prompt}
\end{figure*}

\begin{figure*}[]
  \centering
  \includegraphics[width=0.9\linewidth]{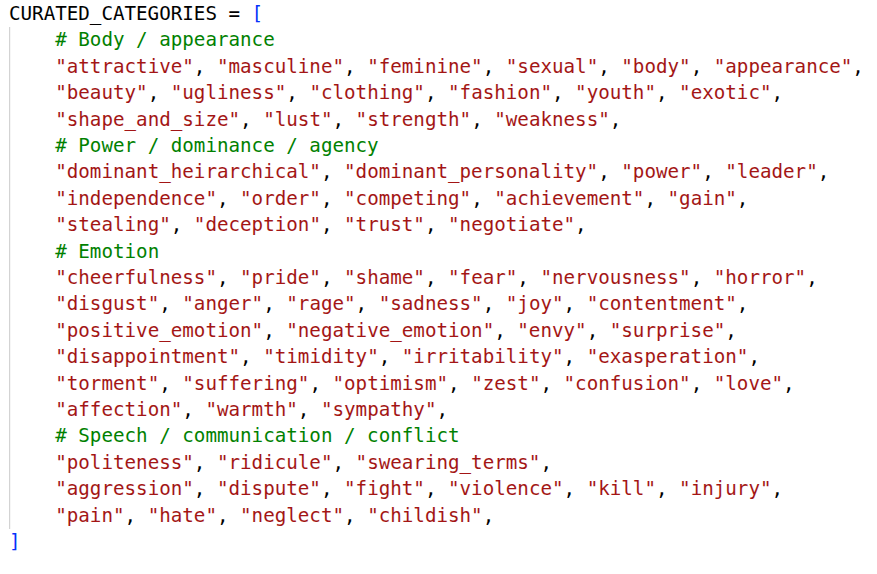}
  \caption{\revised{The semantic categories used for the Empath analysis of Speech concept.}}
  \label{fig:speech_concept_prompt}
\end{figure*}

\begin{table*}[]
\small
\centering
\caption{Proportion of positives (in EN vs HN$\cup$S perspective) in each movie}
\label{tab:positive-proportion}
\begin{tabular}{lcc}
\toprule
Movie title & \makecell{fraction of positive\\(in duration)} & \makecell{fraction of positive\\(in nb of clips)} \\
\midrule
As Good as It Gets & 0.30 & 0.68 \\
Crash & 0.32 & 0.59 \\
Crazy,Stupid,Love & 0.68 & 0.77 \\
Flight & 0.31 & 0.64 \\
Forrest Gump & 0.23 & 0.57 \\
Friends with Benefits & 0.38 & 0.67 \\
Gone Girl & 0.33 & 0.58 \\
Indiana Jones and the Last Crusade & 0.13 & 0.54 \\
Juno & 0.12 & 0.68 \\
Marley \& Me & 0.36 & 0.52 \\
Pulp Fiction & 0.35 & 0.59 \\
Silver Linings Playbook & 0.30 & 0.64 \\
Sleepless in Seattle & 0.55 & 0.65 \\
The Day the Earth Stood Still & 0.15 & 0.34 \\
The Girl with the Dragon Tattoo & 0.32 & 0.58 \\
The Help & 0.56 & 0.66 \\
The Social Network & 0.26 & 0.59 \\
The Ugly Truth & 0.63 & 0.75 \\
Titanic & 0.27 & 0.57 \\
Up in the Air & 0.42 & 0.64 \\
\bottomrule
\end{tabular}
\end{table*}

\subsection{Experimental analysis}\label{app:exp_settings}
Below we define formally the confidence intervals (CIs) we use for model comparison, building on \cite{neurips_tuto, Loftus1994, Masson2003}.

We want to compare $K$ types of models on a given metric $X$, each type of model being trained and evaluated on $n$ (fold, seed) pairs, referred to as \textit{subjects}. (Here $n = 5 \times 5 = 25$.) In total, we have $N = n K$ observations. For $k \in [\![ 1,K ]\!]$, for $i \in [\![ 1,n ]\!]$, we note $x_{ik}$ the metric for the $i$-th instance of model $k$. For $k \in [\![ 1,K ]\!]$, we note $\bar{x}_k$ the average metric for model $k$, while $\bar{x}$ represents the average metric over the whole population of $nK$ models.

While still corresponding to the outcome of Student tests, the following confidence intervals rely on the usual instrumental ANOVA quantities named sums of squares (SS):


\[ SS_{total} = \sum_{k=1}^{K} \sum_{i=1}^{n} (x_{ik} - \bar{x})^{2} \]

\[ SS_{between} = \sum_{k=1}^{K} n (\bar{x}_{k} - \bar{x})^{2} \]

\[ SS_{within} = \sum_{k=1}^{K} \sum_{i=1}^{n} (x_{ik} - \bar{x}_k)^{2} \]

In particular, the following relationship holds :
$SS_{total} = SS_{between} + SS_{within}$.

When the subjects are paired across the groups, which is the case here, we can further decompose the variance: $ SS_{within} = SS_{subject} + SS_{error}$, where 
\[ SS_{subject} = K \sum_{i=1}^{n} (\bar{x}_{i} - \bar{x})^{2}\mbox{ , and} \]
\[SS_{error} = SS_{within} - SS_{subject}\]

Corresponding mean sums of squares are obtained by dividing sums of squares by their corresponding degrees of freedom. In particular:

\[ MS_{within} = \frac{SS_{within}}{df_{within}} \mbox{  , where $df_{within} = N - K$} \]

Assuming the homogeneity of variance across groups (this has to be verified in practice), $MS_{within}$ is an estimate of $\hat{\sigma}^2$ the (common) variance of $X$ in each group. Similarly:

\[ MS_{error} = \frac{SS_{error}}{df_{error}} \mbox{  , where $df_{error} = (K - 1)(n-1)$} \]

$MS_{error}$ is nothing else but an estimate of $\hat{\sigma^\prime}^2$ the (common) variance of each group on the normalized data, i.e., the data we removed the inter-subject variability of (for more details, see \cite{Loftus1994}).

As proven in \cite{Loftus1994}, under the assumption that the (fold, seed) variability plays no role in the comparison of models, the $(1-\alpha$)-confidence interval corresponding to the Student test comparing models $k$ and $j$ is the following:

\[
CI_{kj} = \bar{x}_k \pm \sqrt{\frac{2~MS_{error}}{n}}~t_{1-\alpha/2}(df_{error})
\]

In other words : if $\bar{x}_j$ does not belong to $CI_{kj}$, with a level of confidence $(1-\alpha)$, the models $k$ and $j$ are significantly different. Because the width of this confidence interval is the same for all $(k,j)$ pairs, we can derive, for each model $k$, the confidence interval:

\[
CI_{diff~k} = \frac{1}{2} CI_{kj} = \bar{x}_k \pm \sqrt{\frac{MS_{error}}{2n}}~t_{1-\alpha/2}(df_{error})
\]

so that, for any two models $k$ and $j$, non-overlapping intervals $CI_{diff~k}$ and $CI_{diff~j}$ can be interpreted as a significant difference.\\

We also comment on effect sizes (between such populations $k$ and $j$) considered as Cohen's $d$, which is nothing else than their standardized difference. Consistently with the work above, we compare the normalized populations, i.e. the population we removed the inter-subject variability of, so we have:
\[
d = \frac{\bar{x}_k - \bar{x}_j}{\hat{\sigma^\prime}}
\]

An effect size larger than $0.9$ is considered to be very large (\cite{Sawilowsky2009-jj}, \cite{Funder2019-zo}).

\subsection{Details on the models}\label{suppl:models}

\subsubsection{Common elements of the training procedure over all the models}
We proceed with cross-fold validation with 5 folds also shown in Table \ref{table:films}. The folds are made so as to have an even representation of the genres.
The classes for the learning tasks are determined as described in the main article (Sec. \ref{sec:tasks}). 
Training is always done with random oversampling on the minority class. The validation set is used for early stopping with patience of 10.

\subsubsection{X-CLIP+Transf.}\label{subsubsec:transformer}

\paragraph{Data Preparation} We adapt X-CLIP (\cite{Ni_X-CLIP}), an extension of CLIP for videos (\cite{CLIP}). We keep the pre-trained model frozen and extract a 512-dimensional feature vector on every window of 16 frames, with a stride of 16, for each input video segment. The obtained vectors are mean-pooled

\paragraph{Model} Our baseline architecture consists of two main components: a self-attention encoder and a classification head. Input visual features are first adjusted to a fixed sequence length of 160 tokens using either zero-padding (for shorter sequences) or uniform down-sampling (for longer ones). A corresponding binary mask is generated to indicate valid token positions within each sequence. 
The self-attention encoder begins with a patch embedding module that projects the 512-dimensional input features into a 128-dimensional embedding space through a sequence of normalization and linear layers. A special classification token (CLS token) is appended to each input sequence and is treated as a learnable embedding.
Then the resulting embeddings, including the CLS token, are given as input to a Transformer encoder comprising 2 layers, each with 8 attention heads. Each layer includes a multi-head self-attention mechanism, followed by a feed-forward network that expands the hidden dimension to 1024. GELU activations and dropout are applied within the feed-forward sublayers. The classification head processes the latest CLS token through two linear layers with a ReLU activation in between. The final output is a set of logits, to which a sigmoid activation is applied for binary classification. 

\paragraph{Training} The model is trained with a batch size of 32 for 100 epochs. Early stopping and learning rate scheduling (with reduction on plateau) are applied based on validation loss performance. Training on a single fold takes approximately 30 minutes on an NVIDIA GeForce GTX 1080 Ti GPU.

\subsubsection{Language model Bert Embeddings + Transf}
\label{subsubsec:bert-embeddings-transf}

\paragraph{Data Preparation} We consider binary classification where we want to detect whether there was an objectifying element in the textual transcription of the speech of the characters. All the annotated segments are associated with the corresponding span of subtitles, and the positive class is made of the segments with the speech concept annotated. The text of the subtitles was downcased and HTML tags were removed.

\paragraph{Model} To extract textual representations, subtitle text is tokenized using a bert-base-uncased tokenizer with padding to 512 tokens and truncation. In the absence of subtitle text, a placeholder sequence of 512 [MASK] tokens is used to ensure consistent input length. The tokenized inputs, along with the attention masks, are processed through a frozen BERT model to obtain contextual token embeddings (last hidden state) and the corresponding attention mask. These embeddings are then processed by a transformer-based architecture, which serves as the backbone of our model and follows the design detailed in Subsec. ~\ref{subsubsec:transformer}.This architecture matches the vision baseline in both structure and parameters (approximately 800k). 

\paragraph{Training} We introduce an initial warmup phase stabilise the learning process. Specifically, we increased incrementally the learning rate from zero to the base value (0.00001) over 10\% of the total training steps. During warmup, the learning rate increases linearly with each step, controlled by a custom scheduler. After the warmup, a ReduceLROnPlateau scheduler adjusts the learning rate based on validation loss performance, reducing it when improvement plateaus with a patience of 5 epochs. Early stopping with a patience of 10 epochs is used to halt training.  Maximum number of epochs is set to 100 and the batch size to 32.

\subsubsection{Audio model}\label{audiomodel}

\paragraph{Data preparation}
To encode the audio track to capture aspects of voice and soundtrack, we select the speech audio encoder Audio Spectrogram Transformer (AST) \footnote{\url{https://huggingface.co/docs/transformers/model_doc/audio-spectrogram-transformer}.}
We process each clip by segmenting it into 2-second audio chunks. We extract feature representation for each chunk using AST kept frozen.

\paragraph{Model} AST features are first adjusted to a fixed sequence length of 15 tokens, corresponding to 30 seconds, using either zero-padding (for shorter sequences) or uniform down-sampling (for longer ones). A binary mask is generated to indicate valid token positions within each sequence. The baseline architecture follows the same Transformer-based design used for the vision and text modality, with a comparable number of trainable parameters (approx. 770k).

\paragraph{Training} The training configuration for this Transf. architecture is the same as BERT+Transf, described in \ref{subsubsec:bert-embeddings-transf}. Specifically, the dropout rate is set to 0.1. The batch size is 32 and the warmup phase similar to that for BERT+Transf in Sec. \ref{subsubsec:bert-embeddings-transf} above.

\subsubsection{Multimodal model}\label{app:multimodalmodel}
The multimodal model described below is generalizable to any combination of modalities.

\paragraph{Model}
For a given combination of modalities, we use the corresponding features described above. Formally, for a given observation: a video feature is a 160-token long sequence of 512-dimensional vectors; a text feature is a 768-token long sequence of 512-dimensional vectors, and an audio feature is a 160-token long sequence of 768-dimensional vectors. First, all modalities are projected to a common feature dimension (equal to the smaller feature dimension from all modalities); then they are concatenated along the temporal dimension, i.e., their tokens are concatenated.

This new dimensionality aside, the architecture of the multimodal model is identical to the Transformers model described in \ref{subsubsec:transformer}. For a text-vision multimodal model, this represents approximately 1,500k parameters.

\paragraph{Training}
For a given objectification task, we only keep as positive the examples that present at least one concept from the chosen modalities; others are treated as negative. For instance, for training a text-vision model, a Hard Negative example whose only annotated concept is Soundtrack is treated as a negative.

The new dimensionality and relabeling aside, the training of the multimodal model is identical to the training of the Transformers model described in \ref{subsubsec:transformer}.

\subsubsection{\revised{Details on MLLMs}}\label{app:MLLM}
\revised{The following hyperparameters are common to all our models.}

\revised{When using MLLMs, video inputs are actually provided in the form of sampled frames: from a given video clip, we uniformly sample 8 frames that are compressed so they do not exceed 50,176 pixels. This definition is sufficient to preserve semantic information of the frames.}

\revised{MLLMs perform inference by sampling, based on their predicted output distribution for each output token. This sampling is fully characterized by the parameters \texttt{temperature} and \texttt{top\_p}. We use standard values $\texttt{temperature}=0.1$ and $\texttt{top\_p}=0.9$, while enforcing reproducibility thanks to a random seed.}

\revised{For our classification task, we require the answer to be enclosed in \texttt{<answer></answer>} tags. If this formatting is not respected and recovery strategies (such as searching at the end of the output text) are unsuccessful, we consider the default prediction to be the negative class.
In practice, the arbitrary choice of the default class has little influence on the final performance, given that only $1\%$ of samples require the default prediction.}

\revised{The following prompt is used for the task EN vs HNS, that is detecting the presence of objectifying concepts.}
\begin{lstlisting}
Your task is to analyze whether the given video presents any objectifying concept.
Objectification can be due to the presence of various visual elements, including:
voyeuristic gaze, suggested or partial nudity, non-reciprocal gaze,
gesture or posture connoting submission,
discrepancy between appearance of woman and context (e.g., perfect makeup on waking up),
stereotypical activities, etc.

Think before answering, and generate:
1. A structured reasoning process enclosed in <think></think> tags
2. A final answer enclosed in <answer></answer> tags

For objectifying videos, the reasoning should be based on a structured 2-step process:
<think> must include the following steps enclosed in corresponding tags:
<step1>: Scene Description — Provide an objective overview of the environment and behaviors in the scene.
<step2>: Objectifying Element Description — Describe the objectifying elements and which characters are involved.
<answer> must be a single answer labeling the video as \"Objectifying\"

For normal videos, the reasoning should be based on a structured 2-step process:
<think> must include the following two steps:
<step1>: Scene and Object Description — Provide a concise and objective overview of the environment and typical behaviors.
<step2>: Normal Element Explanation — Explain why the video is not objectifying.
<answer> must be a single answer labeling the video as \"Normal\"
%\end{Verbatim}
\end{lstlisting}

\revised{Prompt for few-shot inference:}
\begin{lstlisting}
Your task is to analyze whether the given video is objectifying or normal.
Objectification occurs when a person or character is represented as an object of desire or service rather than a subject of action.
Objectification can be due to the presence of various visual elements, including:
voyeuristic gaze, suggested or partial nudity, non-reciprocal gaze,
gesture or posture connoting submission,
discrepancy between appearance of woman and context (e.g., perfect makeup on waking up),
stereotypical activities, etc.

Think before answering, and generate:
1. A structured reasoning process enclosed in <think></think> tags
2. A final answer enclosed in <answer></answer> tags

For objectifying videos, the reasoning should be based on a structured 2-step process:
<think> must include the following steps enclosed in corresponding tags:
<step1>: Scene Description - Provide an objective overview of the environment and behaviors in the scene.
<step2>: Objectifying Element Description - Describe the objectifying elements and which characters are involved.
<answer> must be a single answer labeling the video as \"Objectifying\"

For normal videos, the reasoning should be based on a structured 2-step process:
<think> must include the following two steps:
<step1>: Scene and Object Description - Provide a concise and objective overview of the environment and typical behaviors.
<step2>: Normal Element Explanation - Explain why the video is not objectifying.
<answer> must be a single answer labeling the video as \"Normal\"

<|video_token|>

Below, you will find example videos and corresponding desired answers.
<|video_token|>The video above is <answer>Normal</answer>. It contains no objectifying concept.
<|video_token|>The video above is <answer>Objectifying</answer>. More precisely, it contains the following concepts : Type of shot, Look, Exp of emotion.
<|video_token|>The video above is <answer>Normal</answer>. It contains no objectifying concept.
<|video_token|>The video above is <answer>Objectifying</answer>. More precisely, it contains the following concepts : Type of shot, Body, Clothing.
\end{lstlisting}

\subsection{The task of localizing objectification}\label{sec:localization}

We consider Actionformer, a reference model by \cite{zhang2022actionformer} for action detection, that we adapt to the new tasks of classifying (named TClassif) and localizing (named TLoc) objectification. We provide all the implementation details below.
Table \ref{table:T2_T1_Actionformer} shows results of Actionformer-Obj on both TLoc (involving localization) and TClassif. Results on TLoc are comparable to those of original Actionformer on EpicKitchen \cite[Table 2]{zhang2022actionformer}, showing again the accessibility of the task with the visual modality.

We re-use the code provided by \cite{zhang2022actionformer} in their Github repository \footnote{\url{https://github.com/happyharrycn/actionformer_release}}.
For task TLoc of temporal objectification localization, we use both original regression and classification branches of Actionformer.
For task TClassif of classification only, we replace the regression branch by the true segment boundaries and only predict the segment class.

\paragraph{Localization settings}
The dataset needs to be adapted to task TLoc of temporal objectification localization. To do so, each film is cut into 5-minute clips. All clips not overlapping a positive segment are discarded, to reproduce the data filtering in \cite{zhang2022actionformer}. We consider this 5 minute duration to hit a trade-off:  we do not consider a smaller value to limit the number of clips we discard because they do not overlap any positive segment, and we do not consider a higher value to limit the difficulty to scale attention. The resulting clips overlap in average two positive segments (with presence of objectification).

\paragraph{Hyperparameters}
A certain number of hyperparameters need to be adapted to our own dataset.
The hyperparameters we adapt are \textit{sequence length} (maximum length of a video in terms of number of features), \textit{window size} (for the attention mechanism) and \textit{regression ranges}. The latter are connected to the possible duration of action detected at each layer. Owing to various constraints including divisibility, we considered (450,15) and (512,17) for sequence length-window size pairs. We selected the latter from performance on a validation set. We also compared the validation performance obtained with the original regression ranges and regression ranges we set from the distribution of objectifying segment durations in our data. The latter gave best results in validation. The ranges are: [[0, 11], [11, 22], [22, 36], [36, 47], [47, 10000]].

\begin{table*}[]
\small
\centering
\caption{Performance on Actionformer-Obj on TLoc and TClassif on the visual modality. Baselines are on TClassif. Class configuration is EN vs S.}
\label{table:T2_T1_Actionformer} 
{\scriptsize
\begin{tabular}{llllllllll}
\toprule
 &  & \multicolumn{3}{c}{MAP} & Average & \multicolumn{3}{c}{Recall@1} & Average\\
Task & Model $\quad\quad\quad\quad\quad$ tIoU & 0.3 & 0.4 & 0.5 & MAP & 0.3 & 0.4 & 0.5 &  Recall@1\\
\cmidrule(rl){1-10}

TLoc & Actionformer-Obj & 0.252 & 0.169 & 0.093 & 0.171 & 0.385 & 0.296 & 0.198 & 0.293\\
\cmidrule(rl){1-10}
TClassif & \makecell{Actionformer-Obj\\(w/o reg.)} & \multicolumn{3}{c}{N/A} & 0.587 & \multicolumn{3}{c}{N/A} & 0.673\\
 & random & \multicolumn{3}{c}{N/A} & 0.281 & \multicolumn{3}{c}{N/A} &  0.447\\
        & allpos & \multicolumn{3}{c}{N/A} & 0.195  & \multicolumn{3}{c}{N/A} & 0.262\\
        & allneg & \multicolumn{3}{c}{N/A} & 0.364 & \multicolumn{3}{c}{N/A} & 0.384\\        
\bottomrule
\end{tabular}
}
\end{table*}

\end{document}